# Local and Global Inference for High Dimensional Nonparanormal Graphical Models

Quanquan Gu* Yuan Cao† Yang Ning‡ Han Liu§

**Abstract**

This paper proposes a unified framework to quantify local and global inferential uncertainty for high dimensional nonparanormal graphical models. In particular, we consider the problems of testing the presence of a single edge and constructing a uniform confidence subgraph. Due to the presence of unknown marginal transformations, we propose a pseudo likelihood based inferential approach. In sharp contrast to the existing high dimensional score test method, our method is free of tuning parameters given an initial estimator, and extends the scope of the existing likelihood based inferential framework. Furthermore, we propose a U-statistic multiplier bootstrap method to construct the confidence subgraph. We show that the constructed subgraph is contained in the true graph with probability greater than a given nominal level. Compared with existing methods for constructing confidence subgraphs, our method does not rely on Gaussian or sub-Gaussian assumptions. The theoretical properties of the proposed inferential methods are verified by thorough numerical experiments and real data analysis.

## 1 Introduction

Graphical models (Lauritzen, 1996) have been widely used to explore the dependence structure of multivariate distributions. In the Gaussian graphical model, a $d$-dimensional random vector

*Department of Systems and Information Engineering, University of Virginia, Charlottesville, VA 22904, USA; e-mail: qg5w@virginia.edu; Department of Operations Research and Financial Engineering, Princeton University, Princeton, NJ 08544, USA; e-mail: qgu@princeton.edu. This work was carried out while the first author was a postdoc at the SMiLe lab at the Princeton University.

†Program in Applied and Computational Mathematics, Princeton University, Princeton, NJ 08544, USA; e-mail: yuanc@princeton.edu

‡Department of Operations Research and Financial Engineering, Princeton University, Princeton, NJ 08544, USA; e-mail: yning@princeton.edu

§Department of Operations Research and Financial Engineering, Princeton University, Princeton, NJ 08544, USA; e-mail: hanliu@princeton.edu

$\boldsymbol{X} = (X_1, \ldots, X_d)^\top \in \mathbb{R}^d$ follows a multivariate normal distribution $N(\mathbf{0}, \boldsymbol{\Sigma})$. It corresponds to an undirected graph $G = (V, E)$, where $V$ contains nodes corresponding to the $d$ variables in $\boldsymbol{X}$, and the edge set $E$ describes the conditional independence relationships among $X_1, \ldots, X_d$. It is well-known that the graph $G$ is encoded by the sparsity pattern of the precision matrix $\boldsymbol{\Theta} = \boldsymbol{\Sigma}^{-1}$. More specifically, no edge connects $X_j$ and $X_k$ if and only if $\Theta_{jk} = 0$. Therefore, the graph estimation problem can be reduced to the estimation of the precision matrix $\boldsymbol{\Theta}$, based on $n$ independent observations $\{\boldsymbol{X}_i\}_{i=1}^n$ sampled from $N(\mathbf{0}, \boldsymbol{\Sigma})$. Such a problem is also known as the covariance selection (Dempster, 1972). A large body of literature (Meinshausen and Bühlmann, 2006; Yuan and Lin, 2007; Banerjee et al., 2008; Friedman et al., 2008; Rothman et al., 2008; Lam and Fan, 2009; Peng et al., 2009; Yuan, 2010; Cai et al., 2011; Ravikumar et al., 2011; Shen et al., 2012; Jalali et al., 2012; Sun and Zhang, 2012b; Zhu et al., 2013, 2014; Yang et al., 2015) has studied the estimation problem for the precision matrix $\boldsymbol{\Theta}$ under different assumptions in the high dimensional setting, where the number of parameters is much larger than the sample size, i.e., $d \gg n$.

Although the Gaussian graphical model has many appealing theoretical properties, the normality assumption is restrictive. The inferred graph can be misleading if the data distribution is not Gaussian. To relax the Gaussian distribution assumption, Liu et al. (2009, 2012); Xue and Zou (2012) extended the Gaussian graphical model to the more flexible nonparanormal graphical model, which is also known as Gaussian copula models (Klaassen and Wellner, 1997). A random vector $\boldsymbol{X}$ is said to belong to a nonparanormal family if there exists a set of univariate monotonic functions $\{f_j\}_{j=1}^d$ such that $f(\boldsymbol{X}) = [f_1(X_1), \ldots, f_d(X_d)]^\top \sim N(\mathbf{0}, \boldsymbol{\Sigma})$. While the parameter estimation and graph recovery are studied by Liu et al. (2012); Xue and Zou (2012), how to quantify the uncertainty of the estimation remains largely unknown.

This paper proposes a unified framework for local and global inference on high dimensional nonparanormal graphical models. In particular, we consider two types of inferential problems: (1) testing the presence of a single edge, i.e., $H_0 : \Theta_{jk} = 0$, and (2) constructing a confidence subgraph $\widehat{G}$ satisfying $\mathbb{P}(\widehat{G} \subseteq G^*) \geq 1 - \alpha$ asymptotically, where $G^*$ is the true graph, and $\alpha \in (0, 1)$ is a given significance level. The first inferential problem arises when testing the independence between two variables $X_j$ and $X_k$ is of interest. In addition, the confidence subgraph $\widehat{G}$ in the second inferential problem characterizes the dependence structure among all variables with a given confidence level. Compared to the recent work on high-dimensional inference (Belloni et al., 2012; Bühlmann, 2013; Zhang and Zhang, 2014; Javanmard and Montanari, 2013; van de Geer et al., 2013; Lockhart et al., 2014; Taylor et al., 2014; Ning and Liu, 2014), our work has the following novel contributions.

First, to eliminate the unknown nuisance functions $\{f_j\}_{j=1}^d$, we propose a pseudo likelihood approach. However, in order to construct a score test, existing high dimensional score test method (Ning and Liu, 2014) cannot be applied, because it needs to solve a Dantzig selector, which turns out to be a problem of size $d^2 \times d^2$ for nonparanormal graphical models and is computationally prohibitive. In contrast, our proposed pseudo likelihood approach does not require such a procedure. Thus, given an initial estimator, our method is free of the tuning parameters. In particular, we establish the asymptotic guarantees on the type I error as well as the local power of the proposed pseudo score test. Our pseudo likelihood method and theory addresses the unique challenge in the inference for high dimensional nonparanormal graphical models. It extends the scope of the existing likelihood based inferential framework and is of independent interest.

Second, we propose a novel U-statistic multiplier bootstrap method to construct the confidence subgraph. The existing method and theory of multiplier bootstrap are mainly developed for approximating the distribution of sum of independent random variables (Chernozhukov et al., 2013). However, due to the use of the pseudo likelihood, we show that the proposed test statistic is approximated by the maximum of high dimensional U-statistics. To address the challenges caused by the sum of nonindependent random variables, we develop a general method for bootstrapping U-statistics. In particular, we propose to apply the multiplier bootstrap to the leading terms of their Hoeffding decompositions. Based on the U-statistic bootstrap method, we can construct a confidence subgraph $\widehat{G}$ and we prove that it satisfies $\mathbb{P}(\widehat{G} \subseteq G^*) \geq 1 - \alpha$, as $n \to \infty$. The proof of this result requires more refined analysis of the Hoeffding decomposition for high dimensional nonlinear transformations of U-statistics. This technique has its own theoretical interest. Compared with existing work targeting at constructing confidence subgraph for Gaussian graphical models (Drton and Perlman, 2007; Drton et al., 2008; Wasserman et al., 2013, 2014), our method does not rely on Gaussian or sub-Gaussian assumptions.

## 1.1 Further Comparison with Related Work

There are several recent works for asymptotic inference in the context of Gaussian graphical models. For example, Jankova and van de Geer (2013) extended the de-sparsified method (van de Geer et al., 2013) to the Gaussian graphical model. They required the irrepresentable condition (Ravikumar et al., 2011) and the assumption $s^3(\log d)/n = o(1)$, where $s = \max_{1\leq j\leq d}\sum_{k=1}^{d}\mathbb{1}(\Theta_{jk} \neq 0)$. In contrast, we do not need the irrepresentable assumption and we improve their results in the sense that a weaker assumption $s^2 \log d/n = o(1)$ is sufficient for our local inference. In addition, Ren et al. (2013); Chen et al. (2015) proposed a scaled Lasso (Sun and Zhang, 2012a) based approach to test the presence of an edge in the Gaussian graphical model and covariate-adjusted Gaussian graphical model, and Liu (2013) further developed a new procedure to control the false discovery rate. Nevertheless, their inference method cannot be extended to nonparanormal graphical models, because they directly manipulate the residuals in the node-wise regression, which is not available in nonparanormal graphical models. We note that, while a larger family of graphical models is considered, our assumption $s^2 \log d/n = o(1)$ achieves the best possible scaling for the Gaussian graphical model (Ren et al., 2013; Liu, 2013).

## 1.2 Organization of the Paper

The remainder of this paper is organized as follows. We briefly review the nonparanormal graphical model in Section 2. In Section 3, we propose a new pseudo score test for local graph inference, and a global graph inferential procedure for constructing subgraph confidence intervals. In Section 4, we establish the asymptotic properties of the hypothesis testing procedures proposed in Section 3. Section 5 shows the numerical results on both synthetic and real-world datasets. Section 6 concludes this work with discussions.

### 1.3 Notation

We summarize the notation to be used throughout the paper. Let $[d] = \{1, 2, \ldots, d\}$. Let $\mathbf{A} = [A_{ij}] \in \mathbb{R}^{d\times d}$ be a $d \times d$ matrix and $\mathbf{x} = [x_1, \ldots, x_d]^\top \in \mathbb{R}^d$ be a $d$-dimensional vector. $\mathrm{vec}(\mathbf{A})$ is the vectorization of $\mathbf{A}$. For $0 < q < \infty$, we define the $\ell_0$, $\ell_q$ and $\ell_\infty$ vector norms as

$$\|\mathbf{x}\|_0 = \sum_{i=1}^{d} \mathbb{1}(x_i \neq 0), \quad \|\mathbf{x}\|_q = \left(\sum_{i=1}^{d} |x_i|^q\right)^{\frac{1}{q}}, \quad \text{and} \quad \|\mathbf{x}\|_\infty = \max_{1\le i\le d} |x_i|,$$

where $\mathbb{1}(\cdot)$ represents the indicator function. We use the following notation for the matrix $\ell_q$, $\ell_{\max}$ and $\ell_F$ norms:

$$\|\mathbf{A}\|_q = \max_{\|\mathbf{x}\|=1} \|\mathbf{A}\mathbf{x}\|_q, \ \|\mathbf{A}\|_{\max} = \max_{1\le i,j\le d} |A_{ij}|, \ \|\mathbf{A}\|_{1,1} = \sum_{i,j=1}^{d} |A_{ij}|, \ \|\mathbf{A}\|_F = \left(\sum_{ij} |A_{ij}|^2\right)^{\frac{1}{2}}.$$

For two matrices $\mathbf{A}$ and $\mathbf{B}$, we use $\mathbf{A} \otimes \mathbf{B}$ to denote the Kronecker product, and $\mathbf{A} \odot \mathbf{B}$ to denote the Hadamard (elementwise) product. For a matrix $\boldsymbol{\Theta}$ and an index set $S \subseteq [d] \times [d]$, $\boldsymbol{\Theta}_S$ denotes the set of numbers $[\Theta_{(jk)}]_{(jk)\in S}$. Let $\Phi(\cdot)$ denote the cumulative distribution function (CDF) of standard normal distribution, and $\Phi^{-1}(\cdot)$ denote its inverse function. For a sequence of random variables $X_n$, we write $X_n \xrightarrow{p} X$ if $X_n$ converges in probability to $X$, and $X_n \rightsquigarrow X$ if $X_n$ converges in distribution to $X$.

## 2 Nonparanormal Graphical Models

In this section, we briefly review nonparanormal graphical models (Liu et al., 2009), which is a semiparametric extension of the Gaussian graphical model. In particular, a random vector $\boldsymbol{X} = (X_1, ..., X_d)^\top$ follows nonparanormal distribution, $\boldsymbol{X} \sim NPN(\mathbf{f}, \boldsymbol{\Sigma})$, if and only if there exists a set of monotonic transformations $\mathbf{f} = (f_j)_{j=1}^d$, such that $\mathbf{f}(\boldsymbol{X}) = [f_1(X_1), ..., f_d(X_d)]^\top \sim N(\mathbf{0}, \boldsymbol{\Sigma})$ with $\mathrm{diag}(\boldsymbol{\Sigma}) = \mathbf{I}$. Given $n$ independent observations $\boldsymbol{X}_1, ..., \boldsymbol{X}_n$, where $\boldsymbol{X}_i = (X_{i1}, ..., X_{id})^\top \sim NPN(\mathbf{f}, \boldsymbol{\Sigma})$, Liu et al. (2012); Xue and Zou (2012) proposed a rank-based estimator, such as Spearman's rho or Kendall's tau, to estimate $\boldsymbol{\Sigma}$, due to their invariance under monotonic marginal transformations. For example, the Kendall's tau estimator is defined as

$$\widehat{\tau}_{jk} = \frac{2}{n(n-1)} \sum_{1\le i<i'\le n} \mathrm{sign}\left[\left(X_{ij} - X_{i'j}\right)\left(X_{ik} - X_{i'k}\right)\right]. \tag{2.1}$$

Liu et al. (2012) showed that $\widehat{\tau}_{jk}$ is an unbiased estimator of $\tau_{jk} = 2/\pi \arcsin(\Sigma_{jk})$, and the correlation matrix $\mathbf{\Sigma}$ can be estimated by $\widehat{\mathbf{\Sigma}} = (\widehat{\Sigma}_{jk})$, where

$$\widehat{\Sigma}_{jk} = \begin{cases} \sin\left(\frac{\pi}{2}\widehat{\tau}_{jk}\right), & j \neq k, \\ 1, & j = k. \end{cases} \tag{2.2}$$

Once an estimate of $\mathbf{\Sigma}$ is obtained, the existing procedures for the Gaussian graphical model can be used to estimate $\mathbf{\Theta} = \mathbf{\Sigma}^{-1}$, which encodes the conditional independence structure among $\boldsymbol{X}$.

# 3 Inference for Nonparanormal Graphical Models

Under the nonparanormal graphical models, $X_j$ and $X_k$ are independent given the remaining variables if and only if $\Theta^*_{jk} = 0$. In this section, we first propose a pseudo score test for $\Theta^*_{jk} = 0$, and then consider how to construct the confidence subgraph.

## 3.1 Local Graph Inference: A Pseudo Score Approach

Without loss of generality, assume that $\mathbf{\Theta}$ can be partitioned as $\mathbf{\Theta} = (\Theta_{jk}, \mathbf{\Theta}_{(jk)^c})$, where $\Theta_{jk} \in \mathbb{R}$ is the parameter of interest and $\mathbf{\Theta}_{(jk)^c} \in \mathbb{R}^{d^2-1}$ is the nuisance parameter. We are interested in testing $H_0 : \Theta^*_{jk} = 0$ versus $H_1 : \Theta^*_{jk} \neq 0$. In what follows, we consider a pseudo score test approach.

Liu et al. (2009) showed that the log-likelihood function of the nonparanormal graphical model contains unknown nuisance functions $\{f_1, \ldots, f_d\}$, and the plug-in estimation for $\{f_1, \ldots, f_d\}$ results in a sub-optimal estimator of $\mathbf{\Theta}$. This makes the likelihood based inference in Ning and Liu (2014) infeasible. To overcome this problem, we define a pseudo log-likelihood function as follows

$$\ell_n(\mathbf{\Theta}) = -\operatorname{tr}\left(\widehat{\mathbf{\Sigma}}\mathbf{\Theta}\right) + \log\det(\mathbf{\Theta}), \tag{3.1}$$

where $\widehat{\mathbf{\Sigma}}$ is the estimated correlation matrix via Kendall's tau correlation coefficients as in (2.1). The pseudo log-likelihood function in (3.1) is motivated by the log-likelihood function of Gaussian graphical model and is invariant to the marginal monotonic transformation functions $\{f_1, \ldots, f_d\}$.

Then we can easily calculate the derivatives as well as Hessians of the pseudo log-likelihood function in (3.1) with respect to $\Theta_{jk}$ and $\Theta_{(jk)^c}$ as follows

$$\nabla_{(jk)}\ell_n(\mathbf{\Theta}) = -\widehat{\Sigma}_{jk} + [\mathbf{\Theta}^{-1}]_{jk}, \quad \nabla_{(jk)^c}\ell_n(\mathbf{\Theta}) = \operatorname{vec}\big(-\widehat{\mathbf{\Sigma}}_{(jk)^c} + [\mathbf{\Theta}^{-1}]_{(jk)^c}\big)$$
$$\nabla^2_{(jk),(jk)^c}\ell_n(\mathbf{\Theta}) = \left[\mathbf{\Theta}^{-1} \otimes \mathbf{\Theta}^{-1}\right]_{(jk),(jk)^c}, \quad \nabla^2_{(jk)^c,(jk)^c}\ell_n(\mathbf{\Theta}) = \left[\mathbf{\Theta}^{-1} \otimes \mathbf{\Theta}^{-1}\right]_{(jk)^c,(jk)^c},$$

The above quantities are pivotal for the design and analysis of the hypotesis test. In addition, the U-statistic structure of $\widehat{\Sigma}_{jk}$ makes our analysis more challenging than the Gaussian graphical model analyzed by Jankova and van de Geer (2013).

Given some initial estimator of $\mathbf{\Theta}$, we define the pseudo score function for $\Theta_{jk}$ as $\nabla_{(jk)}\ell_n\big(\Theta_{jk}, \widehat{\mathbf{\Theta}}_{(jk)^c}\big)$, where $\widehat{\mathbf{\Theta}}_{(jk)^c}$ is the estimated nuisance parameter. The standard M-estimator theory relies on the asymptotic normality of the score type function, say $\nabla_{(jk)}\ell_n\big(\Theta_{jk}, \widehat{\mathbf{\Theta}}_{(jk)^c}\big)$. However, this property breaks down in the high dimensional setting, due to the regularization on the nuisance parameter estimator $\widehat{\mathbf{\Theta}}_{(jk)^c}$ (Ning and Liu, 2014). In order to address the problem, we propose a decorrelated pseudo score function by removing the effect of the estimated high dimensional nuisance parameters $\widehat{\mathbf{\Theta}}_{(jk)^c}$. More specifically, the decorrelated pseudo score function $S_n^{(jk)}(\mathbf{\Theta})$ is constructed as a linear combination of the pseudo score functions $\nabla\ell_n(\mathbf{\Theta})$, such that it achieves the robustness property with respect to the nuisance parameter, i.e., $\mathbb{E}(\partial S_n^{(jk)}(\mathbf{\Theta}^*)/\partial\mathbf{\Theta}_{(jk)^c}) = 0$. Simple algebra shows that the function satisfies these requirements is given by

$$S_n^{(jk)}(\mathbf{\Theta}) = -\widehat{\Sigma}_{jk} + [\mathbf{\Theta}^{-1}]_{jk} - \big\langle \mathbf{w}^*_{(jk)}, \mathrm{vec}\big(-\widehat{\mathbf{\Sigma}}_{(jk)^c} + [\mathbf{\Theta}^{-1}]_{(jk)^c}\big)\big\rangle, \tag{3.2}$$

where $\mathbf{w}^*_{(jk)} = [\mathbf{H}_{(jk)^c,(jk)^c}]^{-1}\mathbf{H}_{(jk)^c,(jk)}$ and $\mathbf{H} = \mathbf{\Theta}^{*-1} \otimes \mathbf{\Theta}^{*-1} = \mathbf{\Sigma}^* \otimes \mathbf{\Sigma}^*$. Note that $[\mathbf{H}_{(jk)^c,(jk)^c}]^{-1}$ is the inverse of $\mathbf{H}_{(jk)^c,(jk)^c}$, and it is a $(d^2-1)\times(d^2-1)$ matrix. The superscript $(jk)$ of $S_n^{(jk)}(\mathbf{\Theta})$ in (3.2) indicates that it is the decorrelated score function for the component $\Theta_{jk}$, and the subscript $(jk)$ of $\mathbf{w}^*_{(jk)}$ indicates that it is the decorrelation vector corresponding to the score function $S_n^{(jk)}(\mathbf{\Theta})$. Note that the pseudo score function $S_n^{(jk)}(\mathbf{\Theta})$ is a function of second-order U-statistics (Van der Vaart, 1998). To see this, we apply mean value theorem to expand the pseudo score function of $\Theta_{jk}$ as follows

$$\nabla_{(jk)}\ell_n(\mathbf{\Theta}) = T_{jk}(\mathbf{\Theta}) \cdot \frac{2}{n(n-1)} \sum_{1\le i<i'\le n} h_{jk}^{ii'}(\mathbf{\Theta}), \tag{3.3}$$

where $T_{jk}(\mathbf{\Theta})$ is defined as

$$T_{jk}(\mathbf{\Theta}) = \cos\left\{ t \arcsin\big([\mathbf{\Theta}^{-1}]_{jk}\big) + \frac{(1-t)\pi}{n(n-1)} \sum_{1\le i<i'\le n} \mathrm{sign}\big[(X_{ij}-X_{i'j})(X_{ik}-X_{i'k})\big] \right\}$$

for some $t \in [0,1]$, and $h_{jk}^{ii'}(\mathbf{\Theta})$ is a kernel function given by

$$h_{jk}^{ii'}(\mathbf{\Theta}) = -\frac{\pi}{2}\,\mathrm{sign}\big[(X_{ij}-X_{i'j})(X_{ik}-X_{i'k})\big] + \arcsin\big([\mathbf{\Theta}^{-1}]_{jk}\big). \tag{3.4}$$

Here, the superscript of $h_{jk}^{ii'}(\cdot)$ indicates that it is constructed using $\boldsymbol{X}_i$ and $\boldsymbol{X}_{i'}$, and the subscript of $h_{jk}^{ii'}$ indicates that it is associated with $\Theta_{jk}$. Denote $\mathbf{T} = [T_{jk}] \in \mathbb{R}^{d\times d}$ and $\mathbf{G} = [G_{jk}] \in \mathbb{R}^{d\times d}$, where $T_{jk} = T_{jk}(\mathbf{\Theta}^*)$ and $G_{jk} = 2/\big(n(n-1)\big)\sum_{1\le i<i'\le n} h_{jk}^{ii'}(\mathbf{\Theta})$. We have $\nabla\ell_n(\mathbf{\Theta}) = \mathbf{T} \odot \mathbf{G}$. Since $\nabla_{(jk)}\ell_n(\mathbf{\Theta})$ is the product of $\tau_{jk}(\mathbf{\Theta})$ and a second-order U-statistic, we can characterize its limiting distribution using the Hájeck projection (Van der Vaart, 1998).

In particular, we have

$$\frac{2}{n(n-1)} \sum_{1\le i<i'\le n} h_{jk}^{ii'}(\boldsymbol{\Theta}) = \frac{2}{n}\sum_{i=1}^{n} g_{jk}(\boldsymbol{X}_i, \boldsymbol{\Theta}) + \frac{2}{n(n-1)} \sum_{1\le i<i'\le n} [h_{jk}^{ii'}(\boldsymbol{\Theta}) - h_{jk}^{ii'|i}(\boldsymbol{\Theta}) - h_{jk}^{i'i|i'}(\boldsymbol{\Theta})], \tag{3.5}$$

where $g_{jk}(\boldsymbol{X}_i, \boldsymbol{\Theta}) = \frac{1}{n-1}\sum_{i'\neq i} h_{jk}^{ii'|i}(\boldsymbol{\Theta})$, and $h_{jk}^{ii'|i}(\boldsymbol{\Theta}) = \mathbb{E}\big[h_{jk}^{ii'}(\boldsymbol{\Theta})|\boldsymbol{X}_i\big]$ is the projection of $h_{jk}^{ii'}(\boldsymbol{\Theta})$ onto the sigma algebra generated by $\boldsymbol{X}_i$. Note that the first term on the right hand side of (3.5) is a sum of i.i.d. random variables, and the second term is asymptotically negligible. Therefore, $2/\big(n(n-1)\big)\sum_{1\le i<i'\le n} h_{jk}^{ii'}(\boldsymbol{\Theta})$ can be approximated by $2/n\sum_{i=1}^{n} g_{jk}(\boldsymbol{X}_i, \boldsymbol{\Theta})$.

Denote $\overline{\mathbf{G}} = [\overline{G}_{jk}] \in \mathbb{R}^{d\times d}$ and $\mathbf{F} = [F_{jk}] \in \mathbb{R}^{d\times d}$, where $\overline{G}_{jk} = 2/n\sum_{i=1}^{n} g_{jk}(\boldsymbol{X}_i, \boldsymbol{\Theta}^*)$ and $F_{jk} = \cos\big\{\arcsin\big([\boldsymbol{\Theta}^{*-1}]_{jk}\big)\big\} = \sqrt{1-[\boldsymbol{\Theta}^{*-1}]_{jk}^2} = \sqrt{1-(\Sigma_{jk}^*)^2}$. We can show that $F_{jk} = T_{jk} + o_{\mathbb{P}}(1)$ for all $(j,k) \in [d]\times[d]$. The pseudo score function of $\boldsymbol{\Theta}$, i.e., $\nabla \ell_n(\boldsymbol{\Theta}^*) = \mathbf{T}\odot\mathbf{G}$, can be approximated by $\mathbf{F}\odot\overline{\mathbf{G}}$. Moreover, since the pseudo score function $S_n^{(jk)}(\boldsymbol{\Theta}^*)$ is a linear combination of $\nabla\ell_n(\boldsymbol{\Theta}^*)$, i.e., $S_n^{(jk)}(\boldsymbol{\Theta}^*) = \mathbf{b}_{(jk)}^\top \mathrm{vec}(\mathbf{T}\odot\mathbf{G})$, where $\mathbf{b}_{(jk)} = [1, -\mathbf{w}_{(jk)}^{*\top}]^\top$. In Section 4, we show that the variance of $\nabla\ell_n(\boldsymbol{\Theta}^*)$ can be approximated by $\mathbf{R} = \mathbb{E}\big[\mathrm{vec}(\mathbf{F}\odot\overline{\mathbf{G}}^i)\mathrm{vec}(\mathbf{F}\odot\overline{\mathbf{G}}^i)^\top\big]$, where $\overline{\mathbf{G}}^i = [\overline{G}_{jk}^i]$ is a matrix such that $\overline{G}_{jk}^i = g_{jk}(\boldsymbol{X}_i, \boldsymbol{\Theta}^*)$. Furthermore, we prove that the limiting distribution of $\sqrt{n}S_n^{(jk)}(\boldsymbol{\Theta}^*)$ is $N(0, 4\sigma^2)$, where $\sigma^2 = \mathbf{R}_{(jk),(jk)} - 2\mathbf{R}_{(jk),(jk)^c}\mathbf{w}_{(jk)}^* + \mathbf{w}_{(jk)}^{*\top}\mathbf{R}_{(jk)^c,(jk)^c}\mathbf{w}_{(jk)}^*$ can be consistently estimated by some estimator $\widehat{\sigma}^2$.

In addition, in the decorrelated score function in (3.2), $\mathbf{w}_{(jk)}^*$ is unknown. To make it a practical test statistic, we also need to estimate it. By the definition of $\mathbf{w}_{(jk)}^*$, a natural estimator of $\mathbf{w}_{(jk)}^*$ is as follows:

$$\widehat{\mathbf{w}}_{(jk)}(\widehat{\boldsymbol{\Theta}}) = \big[\widehat{\mathbf{H}}_{(jk)^c,(jk)^c}\big]^{-1}\widehat{\mathbf{H}}_{(jk)^c,(jk)}, \tag{3.6}$$

where $\widehat{\mathbf{w}}$ is a function of $\widehat{\boldsymbol{\Theta}}$. Since $\widehat{\mathbf{H}}$ is a $d^2\times d^2$ matrix, its inversion is computationally prohibitive. Fortunately, by the property of Kronecker product, we have $\mathbf{H}^{-1} = \boldsymbol{\Theta}^*\otimes\boldsymbol{\Theta}^*$ and $\widehat{\mathbf{H}}^{-1} = \widehat{\boldsymbol{\Theta}}\otimes\widehat{\boldsymbol{\Theta}}$. In what follows, for notational simplicity, we drop out the super and sub script $(jk)$ in $S_n^{(jk)}(\boldsymbol{\Theta})$, $\mathbf{w}_{(jk)}^*$, $\mathbf{b}_{(jk)}$, $\widehat{S}_n^{(jk)}(\boldsymbol{\Theta})$, $\widehat{\mathbf{w}}_{(jk)}^*$ and $\widehat{\mathbf{b}}_{(jk)}$, when they can be inferred from context.

The following lemma shows that $\mathbf{w}^*$ and $\widehat{\mathbf{w}}(\widehat{\boldsymbol{\Theta}})$ can be efficiently calculated without performing matrix inversion.

**Lemma 3.1.** Let $\boldsymbol{\Omega} = \mathbf{H}^{-1} = \boldsymbol{\Theta}^*\otimes\boldsymbol{\Theta}^*$ and $\widehat{\boldsymbol{\Omega}} = \widehat{\mathbf{H}}^{-1} = \widehat{\boldsymbol{\Theta}}\otimes\widehat{\boldsymbol{\Theta}}$. We have

$$\mathbf{w}^* = -\frac{\boldsymbol{\Omega}_{(jk)^c,(jk)}}{\boldsymbol{\Omega}_{(jk),(jk)}} = -\frac{(\boldsymbol{\Theta}^*\otimes\boldsymbol{\Theta}^*)_{(jk)^c,(jk)}}{(\boldsymbol{\Theta}^*\otimes\boldsymbol{\Theta}^*)_{(jk),(jk)}}, \quad \widehat{\mathbf{w}}(\widehat{\boldsymbol{\Theta}}) = -\frac{\widehat{\boldsymbol{\Omega}}_{(jk)^c,(jk)}}{\widehat{\boldsymbol{\Omega}}_{(jk),(jk)}} = -\frac{(\widehat{\boldsymbol{\Theta}}\otimes\widehat{\boldsymbol{\Theta}})_{(jk)^c,(jk)}}{(\widehat{\boldsymbol{\Theta}}\otimes\widehat{\boldsymbol{\Theta}})_{(jk),(jk)}}. \tag{3.7}$$

From Lemma 3.1, we can see that the estimator $\widehat{\mathbf{w}}(\widehat{\boldsymbol{\Theta}})$ in (3.6) can be computed very efficiently from $\widehat{\mathbf{\Omega}}$. This avoids the node-wise Lasso procedure for estimating a high dimensional matrix in van de Geer et al. (2013) or the Dantzig estimator for estimating a high dimensional vector in Ning and Liu (2014). Replacing $\mathbf{w}^*_{(jk)}$ in (3.2) with $\widehat{\mathbf{w}}(\widehat{\mathbf{\Theta}})$, we can show that the decorrelated pseudo score function reduces to

$$\widehat{S}_n\big(0, \widehat{\mathbf{\Theta}}_{(jk)^c}\big) = \frac{\mathbf{e}_j^\top \check{\mathbf{\Theta}}^\top \widehat{\mathbf{\Sigma}} \check{\mathbf{\Theta}} \mathbf{e}_k}{\widehat{\Theta}_{jj}\widehat{\Theta}_{kk}}, \tag{3.8}$$

where $\check{\mathbf{\Theta}} = \big(0, \widehat{\mathbf{\Theta}}_{(jk)^c}\big)$. In Section 4, we prove the asymptotic normality of $\sqrt{n} \cdot \widehat{S}_n\big(0, \widehat{\mathbf{\Theta}}_{(jk)^c}\big)$ under $H_0 : \Theta^*_{jk} = 0$. In particular, we show that its limiting distribution is the same as the limiting distribution of $\sqrt{n} \cdot S_n\big(0, \mathbf{\Theta}^*_{(jk)^c}\big)$, i.e., $N(0, 4\sigma^2)$. Hence, the score test statistic for $H_0 : \Theta^*_{jk} = 0$ is defined as

$$\mathrm{ST}_n = \sqrt{n}\widehat{S}_n\big(0, \widehat{\mathbf{\Theta}}_{(jk)^c}\big)/(2\widehat{\sigma}), \tag{3.9}$$

where $\widehat{\sigma}^2$ is the estimator of $\sigma^2$ (which will be explained in details in Section 4). The score test with the significance level $\alpha$ is given by

$$\Psi_S(\alpha) = \mathbb{1}(\mathrm{ST}_n^2 > \chi^2_{1\alpha}), \tag{3.10}$$

where $\chi^2_{1\alpha}$ is the $(1-\alpha)$-th quantile of a $\chi^2_1$ random variable. The null hypothesis is rejected if and only if $\Phi_S(\alpha) = 1$. The associated $p$-value of this test is $P_S = 2\big(1 - \Phi(|\mathrm{ST}_n|)\big)$.

### 3.2 Local Graph Inference: Confidence Interval

Although the score test is convenient for hypothesis testing, it does not provide a confidence interval for the parameter of interest $\Theta_{jk}$. In this subsection, we present a Wald test for nonparanormal graphical models, which can be equivalently used to construct the confidence interval for $\Theta_{jk}$.

The Wald test is also based on the decorrelated pseudo score function. More specifically, the first-order Taylor approximation of the pseudo score function $\widehat{S}_n(\Theta_{jk}, \widehat{\mathbf{\Theta}}_{(jk)^c})$ around $\widehat{\Theta}_{jk}$ gives rise to $\widehat{S}_n\big(\Theta_{jk}, \widehat{\mathbf{\Theta}}_{(jk)^c}\big) \approx \widehat{S}_n\big(\widehat{\mathbf{\Theta}}\big) + \nabla_{(jk)}\widehat{S}_n\big(\widehat{\mathbf{\Theta}}\big)\big(\Theta_{jk} - \widehat{\Theta}_{jk}\big)$. This Taylor expansion yields an approximate root of the (approximately) unbiased estimating equation $\widehat{S}_n(\Theta_{jk}, \widehat{\mathbf{\Theta}}_{(jk)^c}) = 0$. We call it as the approximate pseudo-likelihood estimator. Since we have $\nabla_{(jk)}\widehat{S}_n\big(\widehat{\mathbf{\Theta}}\big) = \big[\mathbf{e}_j\widehat{\mathbf{\Theta}}\widehat{\mathbf{\Sigma}}\mathbf{e}_k + \mathbf{e}_j\widehat{\mathbf{\Sigma}}\widehat{\mathbf{\Theta}}\mathbf{e}_k - 1\big]/\big(\widehat{\Theta}_{jj}\widehat{\Theta}_{kk}\big)$ and $\widehat{S}_n\big(\widehat{\mathbf{\Theta}}\big) = \big[\mathbf{e}_j^\top\widehat{\mathbf{\Theta}}^\top\widehat{\mathbf{\Sigma}}\widehat{\mathbf{\Theta}}\mathbf{e}_k - \widehat{\Theta}_{jk}\big]/\big(\widehat{\Theta}_{jj}\widehat{\Theta}_{kk}\big)$, after some algebra, we show that the approximate pseudo-likelihood estimator is given by

$$\widehat{\Theta}^W_{jk} = \frac{\widehat{\Theta}_{jk}\mathbf{e}_j\widehat{\mathbf{\Theta}}\widehat{\mathbf{\Sigma}}\mathbf{e}_k + \widehat{\Theta}_{jk}\mathbf{e}_j\widehat{\mathbf{\Sigma}}\widehat{\mathbf{\Theta}}\mathbf{e}_k - \mathbf{e}_j^\top\widehat{\mathbf{\Theta}}^\top\widehat{\mathbf{\Sigma}}\widehat{\mathbf{\Theta}}\mathbf{e}_k}{\mathbf{e}_j\widehat{\mathbf{\Theta}}\widehat{\mathbf{\Sigma}}\mathbf{e}_k + \mathbf{e}_j\widehat{\mathbf{\Sigma}}\widehat{\mathbf{\Theta}}\mathbf{e}_k - 1}. \tag{3.11}$$

In Section 4, we prove that the limiting distribution of $\sqrt{n}\big(\widehat{\Theta}^W_{jk} - \Theta^*_{jk}\big)$ is $N(0, 4\sigma^2 \cdot H^{-2}_{(jk)|(jk)^c})$, where $H_{(jk)|(jk)^c} = H_{(jk),(jk)} - \mathbf{H}_{(jk),(jk)^c}[\mathbf{H}_{(jk)^c,(jk)^c}]^{-1}\mathbf{H}_{(jk)^c,(jk)}$ is the partial Fisher information. In order to construct the confidence interval, we need to estimate $\sigma$ and $H_{(jk)|(jk)^c}$.

For $\sigma$, we use the same estimator used in the score test. For $H_{(jk)|(jk)^c}$, by the definition of $\mathbf{w}^*$, we use the following estimator $\widehat{H}_{(jk)|(jk)^c} = \nabla^2_{(jk),(jk)}\ell_n(\widehat{\boldsymbol{\Theta}}) - \nabla^2_{(jk),(jk)^c}\ell_n(\widehat{\boldsymbol{\Theta}})\widehat{\mathbf{w}}(\widehat{\boldsymbol{\Theta}})$, which can be further simplified as $\widehat{H}_{(jk)|(jk)^c} = 1/(\widehat{\Theta}_{jj}\widehat{\Theta}_{kk})$. Thus, the two-sided confidence interval for $\Theta^*_{jk}$ with $1-\alpha$ coverage probability is given by

$$[\widehat{\Theta}^W_{jk} - n^{-1/2}\Phi^{-1}(1-\alpha/2)\cdot L_{jk}, \widehat{\Theta}^W_{jk} + n^{-1/2}\Phi^{-1}(1-\alpha/2)\cdot L_{jk}], \tag{3.12}$$

where $L_{jk} = 2\widehat{\sigma}\widehat{\Theta}_{jj}\widehat{\Theta}_{kk}$. From the hypothesis test perspective, we can consider the following Wald test statistic for $H_0 : \Theta^*_{jk} = 0$:

$$W_n = (2\widehat{\sigma})^{-1}\widehat{H}_{(jk)|(jk)^c}\sqrt{n}\widehat{\Theta}^W_{jk}. \tag{3.13}$$

Then the Wald test with significance level $\alpha$ is given by

$$\Psi_W(\alpha) = \mathbb{1}(W_n^2 > \chi^2_{1\alpha}). \tag{3.14}$$

The null hypothesis is rejected if and only if $\Phi_W(\alpha) = 1$. The associated $p$-value of this test is $P_W = 2\big(1 - \Phi(|W_n|)\big)$.

### 3.3 Global Graph Inference: Confidence Subgraph

In the previous subsections, we have discussed testing and confidence intervals for individual component of the precision matrix $\boldsymbol{\Theta}^*$, which is referred to as the local graph inference. Here, we consider how to construct a confidence subgraph with any specified confidence level, which is referred to as the global graph inference.

More specifically, we aim at providing a confidence band for $\boldsymbol{\Theta}^*$ with confidence level $\alpha$, which can be used to construct a confidence subgraph. The main idea is that, based on the approximate pseudo-likelihood estimator $\widehat{\Theta}^W_{jk}$ in (3.11), if we are able to obtain the $(1-\alpha)$-quantile of $\max_{(j,k)\in[d]\times[d]}\sqrt{n}\big|(\widehat{\Theta}^W_{jk} - \Theta^*_{jk})/L_{jk}\big|$, then we have

$$\mathbb{P}\bigg[\max_{(j,k)}\sqrt{n}\bigg|\frac{\widehat{\Theta}^W_{jk} - \Theta^*_{jk}}{L_{jk}}\bigg| \le c_\alpha\bigg] = \mathbb{P}\bigg[\widehat{\Theta}^W_{jk} - \frac{c_\alpha L_{jk}}{\sqrt{n}} \le \Theta^*_{jk} \le \widehat{\Theta}^W_{jk} + \frac{c_\alpha L_{jk}}{\sqrt{n}}, \forall (j,k)\in[d]\times[d]\bigg],$$

where $L_{jk} = 2\widehat{\sigma}\widehat{\Theta}_{jj}\widehat{\Theta}_{kk}$. Based on the above inequality, we can construct a confidence subgraph. In details, starting from a complete graph, for each edge $(j,k)$, if $0 \in \big[\widehat{\Theta}^W_{jk} - c_\alpha L_{jk}/\sqrt{n}, \widehat{\Theta}^W_{jk} + c_\alpha L_{jk}/\sqrt{n}\big]$, then this edge is removed. We denote the resulting graph as $\widehat{G}$, which is a $(1-\alpha)$-confidence subgraph.

As we have seen, the key to construct the confidence subgraph is to characterize the limiting distribution of $\max_{(j,k)\in[d]\times[d]}\sqrt{n}\big|(\widehat{\Theta}^W_{jk} - \Theta^*_{jk})/L_{jk}\big|$. More specifically, we define the following pivotal quantity

$$T = \max_{(j,k)\in[d]\times[d]}\sqrt{n}\Big(\widehat{\Theta}_{jk} - \Theta^*_{jk}\Big)\cdot\widehat{H}_{(jk)|(jk)^c}/(2\widehat{\sigma}). \tag{3.15}$$

Define its $(1-\alpha)$-th quantile as $c_T(1-\alpha) = \inf\big\{t \in \mathbb{R} : \mathbb{P}(T \le t) \ge \alpha\big\}$. Since we can show that

$$\max_{(j,k)\in[d]\times[d]} \sqrt{n}\big|\big(\widehat{\Theta}^W_{jk} - \Theta^*_{jk}\big) \cdot \widehat{H}_{(jk)|(jk)^c}/(2\widehat{\sigma}) - \mathbf{b}^\top_{(jk)}\mathrm{vec}(\mathbf{F} \odot \overline{\mathbf{G}})/(2\sigma)\big| = o_{\mathbb{P}}(1), \tag{3.16}$$

where $\mathbf{b}^\top_{(jk)}\mathrm{vec}(\mathbf{F} \odot \overline{\mathbf{G}})$ is the Hájek projection of $S_n^{(jk)}(\mathbf{\Theta}^*)$, this motivates us to construct a Gaussian multiplier bootstrap estimator of $T$ as follows

$$W = \max_{(j,k)\in[d]\times[d]} \frac{1}{\sqrt{n}} \sum_{i=1}^n z_{ijk} \cdot e_i, \tag{3.17}$$

where $z_{ijk} = -1/(2\widehat{\sigma}) \cdot \widehat{\mathbf{b}}^\top_{(jk)}\mathrm{vec}(\widehat{\mathbf{F}} \odot \widehat{\mathbf{G}}^i)$ with $\widehat{\mathbf{b}}_{(jk)} = [1, -\widehat{\mathbf{w}}^\top_{(jk)}]^\top$, $\widehat{\mathbf{G}}^i = [\widehat{G}^i_{jk}] \in \mathbb{R}^{d\times d}$ such that $\widehat{G}^i_{jk} = 1/(n-1)\sum_{i'\neq i} h^{ii'}_{jk}(\widehat{\mathbf{\Theta}})$, and $\widehat{\mathbf{F}} = [\widehat{F}_{jk}] \in \mathbb{R}^{d\times d}$ such that $\widehat{F}_{jk} = \sqrt{1-\widehat{\Sigma}^2_{jk}}$, and $\{e_i\}_{i=1}^n$ is a sequence of i.i.d. standard normal random variables that are independent of $\{\boldsymbol{X}_i\}_{i=1}^n$. Note that although we cannot derive the quantile of $W$ analytically, we can compute it using Monte-Carlo method as in standard bootstrap methods. We prove in Section 4 that the quantile of $W$ converges to the quantile of $T$ uniformly. To justify the theory of the resulting Gaussian multiplier bootstrap, we need more careful analysis of the uniform approximation error in (3.16) due to the Hájek projection. Given the validity of the resulting Gaussian multiplier bootstrap, we can obtain a uniform confidence interval of each $\Theta^*_{jk}$ as

$$\big[\widehat{\Theta}^W_{jk} - n^{-1/2} \cdot c_W(1-\alpha/2) \cdot L_{jk}, \widehat{\Theta}^W_{jk} + n^{-1/2} \cdot c_W(1-\alpha/2) \cdot L_{jk}\big],$$

for all $(j,k) \in [d]\times[d]$, where $c_W(1-\alpha/2)$ is the $(1-\alpha/2)$-quantile of $W$, $L_{jk} = 2\widehat{\sigma}\widehat{\Theta}_{jj}\widehat{\Theta}_{kk}$, from which, we can immediately construct the confidence subgraph.

# 4 Theory of Local and Global Inference

In this section, we establish the main theoretical results for the local and global inferential procedures proposed in Section 3.

## 4.1 Asymptotic Distribution of Score Test Under the Null Hypothesis

We consider the following precision matrix class with some $M > 0$ and $s > 0$,

$$\mathcal{U}(s, M) = \Big\{\mathbf{\Theta} : \mathbf{\Theta} \succ \mathbf{0}, \|\mathbf{\Theta}\|_1 \le M, \max_{1\le j\le d} \sum_{k=1}^d \mathbb{1}(\Theta_{jk} \neq 0) \le s\Big\}, \tag{4.1}$$

where $\mathbf{\Theta} \succ \mathbf{0}$ indicates that $\mathbf{\Theta}$ is a symmetric and positive definite matrix. Note that $\max_{1\le j\le d}\sum_{k=1}^d \mathbb{1}(\Theta_{jk} \neq 0) \le s$ constrains the maximum degree of the graph associated with the sparse precision matrix to be no more than $s$.

As shown in Section 3, the test statistic $\mathrm{ST}_n$ depends on the pseudo empirical score function $\widehat{S}_n(\boldsymbol{\Theta})$ defined in (3.8). To characterize the limiting distribution of $\widehat{S}_n(\widehat{\boldsymbol{\Theta}})$, we first derive the limiting distribution of $S_n(\boldsymbol{\Theta}^*)$ and then show that $\widehat{S}_n(\widehat{\boldsymbol{\Theta}})$ converges to $S_n(\boldsymbol{\Theta}^*)$ uniformly.

Since $S_n(\boldsymbol{\Theta}^*)$ is a linear combination of $\nabla \ell_n(\boldsymbol{\Theta}^*)$, it is pivotal to analyze the limiting behavior of $\nabla \ell_n(\boldsymbol{\Theta}^*)$. For nonparanormal graphical models, one challenge is that its pseudo score function $\nabla \ell_n(\boldsymbol{\Theta}^*)$ is a function of U-statistic. To establish its asymptotic normality, as we show in Section 3, we use the Hájek projection technique (Van der Vaart, 1998) to approximate this U-statistic by a sum of i.i.d. random variables, i.e., $\nabla \ell_n(\boldsymbol{\Theta}^*) = \mathbf{T} \odot \mathbf{G} \approx \mathbf{F} \odot \overline{\mathbf{G}}$. Recall that $\mathbf{T} = [T_{jk}] \in \mathbb{R}^{d\times d}$ with $T_{jk} = \tau_{jk}(\boldsymbol{\Theta}^*)$, $\mathbf{F} = [F_{jk}] \in \mathbb{R}^{d\times d}$ such that $F_{jk} = \sqrt{1 - [\boldsymbol{\Theta}^{*-1}]_{jk}^2}$, $\mathbf{G} = [G_{jk}] \in \mathbb{R}^{d\times d}$ such that $G_{jk} = 2/\big(n(n-1)\big) \sum_{1\le i<i'\le n} h_{jk}^{ii'}(\boldsymbol{\Theta})$, and $\overline{\mathbf{G}} = [\overline{G}_{jk}] \in \mathbb{R}^{d\times d}$ such that $\overline{G}_{jk} = 2/n \sum_{i=1}^n g_{jk}(\boldsymbol{X}_i, \boldsymbol{\Theta}^*)$. Note that $\overline{\mathbf{G}}$ is composed of a sum of independent random variables, and $F_{jk} \cdot 2/n \cdot g_{jk}(\boldsymbol{X}_i, \boldsymbol{\Theta}^*)$ is the projection of $\nabla \ell_n(\boldsymbol{\Theta}^*)$ onto the $\sigma$-filed generated by $\boldsymbol{X}_i$. We show that $\mathbf{F} \odot \overline{\mathbf{G}}$ is a good approximation of $\nabla \ell_n(\boldsymbol{\Theta}^*)$.

Furthermore, it is easy to show that $\mathbb{E}\big[g_{jk}(\boldsymbol{X}_i, \boldsymbol{\Theta}^*)\big] = 0$. Thus, we have $\mathbb{E}[\mathbf{F} \odot \overline{\mathbf{G}}] = \mathbf{0}$. Let $\overline{\mathbf{G}}^i = [\overline{G}_{jk}^i]$ be a matrix such that $\overline{G}_{jk}^i = g_{jk}(\boldsymbol{X}_i, \boldsymbol{\Theta}^*)$. Then the variance of $\nabla \ell_n(\boldsymbol{\Theta}^*)$ can be approximated by

$$\mathbf{R} = \mathbb{E}\big[\mathrm{vec}(\mathbf{F} \odot \overline{\mathbf{G}}^i)\mathrm{vec}(\mathbf{F} \odot \overline{\mathbf{G}}^i)^\top\big], \tag{4.2}$$

because $\mathbf{R}$ is the second moment of the Hájek projection of $\nabla \ell_n(\boldsymbol{\Theta}^*)$, which is a proxy to the variance of $\nabla \ell_n(\boldsymbol{\Theta}^*)$. Before presenting the main theoretical results, we first make several mild regularity assumptions, which are essential to establish our results.

**Assumption 4.1.** There exists a constant $\zeta > 0$ such that $\lambda_{\min}(\mathbf{R}) \ge \zeta$.

Assumption 4.1 says that the minimum eigenvalue of $\mathbf{R}$ is lower bounded away from 0. This assumption is a common condition to guarantee non-degeneracy of U-statistics (Van der Vaart, 1998). It guarantees the validity of using $\mathbf{F} \odot \overline{\mathbf{G}}$ to approximate $\nabla \ell_n(\boldsymbol{\Theta}^*)$. In order to derive the asymptotic normality of $S_n(\boldsymbol{\Theta}^*)$, we require the following assumption.

**Assumption 4.2.** There exists a constant $\nu > 0$ such that $1/\nu \le \lambda_{\min}(\boldsymbol{\Sigma}^*) < \lambda_{\max}(\boldsymbol{\Sigma}^*) \le \nu$. In addition, there exists a constant $K_{\boldsymbol{\Sigma}^*} > 0$, such that $\|\boldsymbol{\Sigma}^*\|_\infty \le K_{\boldsymbol{\Sigma}^*}$.

The first part of Assumption 4.2 requires that the smallest eigenvalue of the correlation $\boldsymbol{\Sigma}^*$ is bounded below from zero, and its largest eigenvalue is finite. This implies that $1/\nu \le \lambda_{\min}(\boldsymbol{\Theta}^*) < \lambda_{\max}(\boldsymbol{\Theta}^*) \le \nu$. This assumptions is commonly imposed in the literature for the analysis of Gaussian graphical models (Ravikumar et al., 2011; Yuan, 2010; Cai et al., 2011) and nonparanormal models (Liu et al., 2009, 2012; Xue and Zou, 2012). The second part of Assumption 4.2 has been made in Ravikumar et al. (2011); Jankova and van de Geer (2013). For example, this assumption is satisfied in the Töplitz covariance matrix, i.e., $\Sigma_{jk}^* = \gamma^{|j-k|}$ where $|\gamma| < 1$.

At the core of our proof, we show that $\sqrt{n}\big(\widehat{S}_n(0, \widehat{\boldsymbol{\Theta}}_{(jk)^c}) - S_n(0, \boldsymbol{\Theta}^*_{(jk)^c})\big) = o_{\mathbb{P}}(1)$. To show this, we also need the following assumption on the estimator error for the precision matrix estimator $\widehat{\boldsymbol{\Theta}}$.

**Assumption 4.3.** The estimator $\widehat{\mathbf{\Theta}}$ satisfies

$$\|\widehat{\mathbf{\Theta}} - \mathbf{\Theta}^*\|_{\max} = O_{\mathbb{P}}\big(\sqrt{\log d/n}\big), \ \|\widehat{\mathbf{\Theta}} - \mathbf{\Theta}^*\|_1 = O_{\mathbb{P}}\big(s\sqrt{\log d/n}\big).$$

Assumption 4.3 essentially requires $\widehat{\mathbf{\Theta}}$ has sufficiently fast convergence rates in terms of both elementwise infinity norm and matrix $\ell_1$ norm, which is the key to show that the remainder term diminishes asymptotically. It can be shown that for any precision matrix $\mathbf{\Theta}^*$ belonging to the class of matrices $\mathcal{U}(s, M)$ for some constant $M$, CLIME (Cai et al., 2011) estimator enjoys all these theoretical guarantees.

**Remark 4.4.** We can also use graph Dantzig selector and neighborhood selection estimators to estimate the precision matrix of nonparanormal graphical models. However, they do not enjoy optimal estimation rates in terms of matrix elementwise max norm. Although this norm can be rudely bounded by spectral norm, it results in suboptimal scaling for hypothesis test. Similarly, the graphical Lasso estimator does not have estimation guarantees in terms of the matrix $\ell_1$ norm by only making assumptions on the smallest eigenvalue of $\mathbf{\Sigma}^*$. To attain matrix $\ell_1$ norm based estimation error bounds, we need to make substantially stronger assumption such as inrrepresentable condition (Ravikumar et al., 2011). Thus, we use the CLIME estimator.

**Assumption 4.5.** It holds that $\lim_{n\to\infty} s\log d/n^{3/2} = 0$.

**Remark 4.6.** Assumption 4.5 is a mild assumption, which says $\log d = o(n^{1/2}/s)$. Recall that for parameter estimation, we require $\log d = o(n/s^2)$. This implies that in order to make the pseudo score test a valid procedure, we require that $d$ goes to infinity at a relatively slower rate. Moreover, we comment that our condition $\log d = o(n^{1/2}/s)$ is weaker than that in Jankova and van de Geer (2013) and already matches the best possible scaling under the Gaussian graphical model (Ren et al., 2013; Liu, 2013).

Equipped with Assumptions 4.1-4.5, we are ready to present the main result for score test, which establishes the asymptotic normality of the estimated decorrelated score function in (3.8).

**Theorem 4.7.** Under Assumptions 4.1, 4.2, 4.3 and 4.5, as $n, d \to \infty$, we have

$$\sqrt{n}\widehat{S}_n\big(0, \widehat{\mathbf{\Theta}}_{(jk)^c}\big) = \sqrt{n}S_n\big(0, \mathbf{\Theta}^*_{(jk)^c}\big) + o_{\mathbb{P}}(1) \rightsquigarrow N\big(0, 4\sigma^2\big),$$

where $\sigma^2 = \mathbf{R}_{(jk),(jk)} - 2\mathbf{R}_{(jk),(jk)^c}\mathbf{w}^* + \mathbf{w}^{*\top}\mathbf{R}_{(jk)^c,(jk)^c}\mathbf{w}^*$.

By Theorem 4.7, to make $\sqrt{n}\widehat{S}_n\big(0, \widehat{\mathbf{\Theta}}_{(jk)^c}\big)$ a valid test statistic, we need to estimate its asymptotic variance $\sigma^2$, which depends on the unknown matrix $\mathbf{R}$ and $\mathbf{w}^*$. Based on the definition of $\sigma^2$ in Theorem 4.7, we can estimate $\sigma^2$ by plugging in $\widehat{\mathbf{R}}$ and $\widehat{\mathbf{w}}$. This leads to the following estimator:

$$\widehat{\sigma}^2 = \widehat{\mathbf{R}}_{(jk),(jk)} - 2\widehat{\mathbf{R}}_{(jk),(jk)^c}\widehat{\mathbf{w}} + \widehat{\mathbf{w}}^\top\widehat{\mathbf{R}}_{(jk)^c,(jk)^c}\widehat{\mathbf{w}}, \tag{4.3}$$

where $\widehat{\mathbf{R}}$ is defined as follows

$$\widehat{\mathbf{R}} = \frac{1}{n}\sum_{i=1}^{n} \text{vec}\big(\widehat{\mathbf{F}} \odot \widehat{\mathbf{G}}^i\big)\text{vec}\big(\widehat{\mathbf{F}} \odot \widehat{\mathbf{G}}^i\big). \tag{4.4}$$

Recall that $\widehat{\mathbf{G}}^i = [\widehat{G}^i_{jk}] \in \mathbb{R}^{d\times d}$ is defined as $\widehat{G}^i_{jk} = 1/(n-1)\sum_{i'\neq i} h^{ii'}_{jk}(\widehat{\boldsymbol{\Theta}})$ and $\widehat{\mathbf{F}} = [\widehat{F}_{ik}] \in \mathbb{R}^{d\times d}$ is defined as $\widehat{F}_{jk} = \sqrt{1-\widehat{\Sigma}^2_{jk}}$. It is easy to show that $\widehat{\sigma}^2$ in (4.3) is a consistent estimator of $\sigma^2$. This, together with Theorem 4.7, establishes the validity of the score test statistic in (3.9) under the null hypothesis.

**Corollary 4.8.** Under the same assumptions of Theorem 4.7, we have

$$\lim_{n\to\infty} \mathbb{P}(\Psi_S(\alpha) = 1|H_0) = \alpha \text{ and } P_S \rightsquigarrow U[0,1],$$

where $U[0,1]$ is a random variable with a uniform distribution on $[0,1]$.

## 4.2 Asymptotic Power Under Local Alternative Hypotheses

In this subsection, we analyze the power of the score test for detecting the alternative hypothesis. More specifically, we are interested in the limiting behavior of $\text{ST}_n$ under a sequence of alternative hypotheses $H_{1n} : \Theta_{jk} = K \cdot n^{-\eta}$, where $K$ is a constant, and $\eta$ is a positive constant. We consider the following parameter space $\mathcal{U}^{(jk)}_1 = \mathcal{U}^{(jk)}_1(K, \eta, s^*, M^*)$:

$$\mathcal{U}^{(jk)}_1 = \Big\{\boldsymbol{\Theta} : \boldsymbol{\Theta} \succ \mathbf{0}, \Theta_{jk} = K\cdot n^{-\eta}, \|\boldsymbol{\Theta}\|_1 \le M^*, \max_{1\le j\le d}\sum_{k=1}^{d} \mathbb{1}(\Theta_{jk} \neq 0) \le s^*\Big\},$$

where $s^* = \max_{1\le j\le d}\sum_{k=1}^{d} \mathbb{1}(\Theta^*_{jk} \neq 0)$ and $\|\boldsymbol{\Theta}^*\|_1 = M^*$. The parameter space $\mathcal{U}^{(jk)}_1(K, \eta, s^*, M^*)$ characterizes the local alternative hypotheses around the null hypothesis $\Theta_{jk} = 0$, in the sense that $\Theta_{jk} = K \cdot n^{-\eta}$ goes to 0 as $n \to \infty$. To ensure that the parameters are estimable under the alternatives, we focus on the sparse local alternative hypotheses. In the rest of this paper, for notational simplicity, we drop out the superscript $(jk)$ in $\mathcal{U}^{(jk)}_1$ when it can be inferred from context.

In the following assumption, we assume that the estimator $\widehat{\boldsymbol{\Theta}}$ has desirable estimation error rates for any $\boldsymbol{\Theta}$ in $\mathcal{U}_1(K, \eta, s^*, U^*)$ uniformly.

**Assumption 4.9.** The estimator $\widehat{\boldsymbol{\Theta}}$ satisfies

$$\lim_{n\to\infty} \inf_{\boldsymbol{\Theta}\in\mathcal{U}_1(K,\eta,s^*,M^*)} \mathbb{P}_{\boldsymbol{\Theta}}\Big(\big\|\widehat{\boldsymbol{\Theta}} - \boldsymbol{\Theta}\big\|_{\max} \le C_0\sqrt{\log d/n}\Big) = 1,$$
$$\lim_{n\to\infty} \inf_{\boldsymbol{\Theta}\in\mathcal{U}_1(K,\eta,s^*,M^*)} \mathbb{P}_{\boldsymbol{\Theta}}\Big(\big\|\widehat{\boldsymbol{\Theta}} - \boldsymbol{\Theta}\big\|_1 \le C_1 s\sqrt{\log d/n}\Big) = 1,$$

where $C_0, C_1$ are positive constants.

Assumption 4.9 is analogous to Assumption 4.3, which holds for the CLIME estimator; see Cai et al. (2011). Now given Assumptions 4.1,4.2 and 4.9, we are able to prove the following theorem, which characterizes the uniform limiting distributions of the score test statistic $\mathrm{ST}_n$ in (3.9) for $\mathbf{\Theta}$ in $\mathcal{U}_1(K, \eta, s^*, U^*)$.

**Theorem 4.10.** Under Assumptions 4.1, 4.2, 4.9, and suppose that we have $\sqrt{n}\Big[s \log d/n + K \cdot n^{-\eta}\sqrt{\log d/n}\Big] = o(1)$, if $n$ is sufficiently large, it holds that

$$\lim_{n\to\infty} \sup_{\mathbf{\Theta}\in\mathcal{U}_1(K,\eta,s^*,M^*)} \sup_{t\in\mathbb{R}} \big|\mathbb{P}_{\mathbf{\Theta}}(\mathrm{ST}_n \le t) - \Phi(t)\big| = 0, \quad \textbf{if} \ \ \eta > 1/2, \tag{4.5}$$

$$\lim_{n\to\infty} \sup_{\mathbf{\Theta}\in\mathcal{U}_1(K,\eta,s^*,M^*)} \sup_{t\in\mathbb{R}} \left|\mathbb{P}_{\mathbf{\Theta}}(\mathrm{ST}_n \le t) - \Phi\left(t + K\frac{H_{(jk)|(jk)^c}}{2\sigma}\right)\right| = 0, \quad \textbf{if} \ \ \eta = 1/2, \tag{4.6}$$

$$\lim_{n\to\infty} \sup_{\mathbf{\Theta}\in\mathcal{U}_1(K,\eta,s^*,M^*)} \sup_{t\in\mathbb{R}} \big|\mathbb{P}_{\mathbf{\Theta}}(\mathrm{ST}_n \le t)\big| = 0, \quad \textbf{if} \ \ \eta < 1/2. \tag{4.7}$$

In fact, this theorem also implies the uniform convergence of $\mathrm{ST}_n$ under the null hypothesis. This is seen by taking $K = 0$ in (4.6). Recall that the power of the score test $\Psi_S(\alpha)$ is defined to be the probability of $\Psi_S(\alpha) = 1$ when $\mathbf{\Theta} \in \mathcal{U}_1(K, \eta, s^*, M^*)$. Recall the fact that the type I error of $\Psi_S(\alpha)$ can be controlled at level $\alpha$ asymptotically as shown in Corollary 4.8. Theorem 4.10 immediately characterizes the uniform asymptotic power of $\Psi_S(\alpha)$ under the alternative hypothesis $H_{1n} : \Theta_{jk} = Kn^{-\eta}$. In particular, Theorem 4.10 implies that

$$\lim_{n\to\infty} \sup_{\mathbf{\Theta}\in\mathcal{U}_1(K,\eta,s^*,M^*)} \sup_{\alpha\in(0,1)} \big|\mathbb{P}_{\mathbf{\Theta}}\big(\Psi_S(\alpha) = 1\big) - \alpha\big| = 0, \quad \textbf{if} \ \ \eta > 1/2, \tag{4.8}$$

$$\lim_{n\to\infty} \sup_{\mathbf{\Theta}\in\mathcal{U}_1(K,\eta,s^*,M^*)} \sup_{\alpha\in(0,1)} \big|\mathbb{P}_{\mathbf{\Theta}}\big(\Psi_S(\alpha) = 1\big) - \psi_\alpha\big| = 0, \quad \textbf{if} \ \ \eta = 1/2, \tag{4.9}$$

where $\psi_\alpha = 1 - \Phi\big(\Phi^{-1}(1-\alpha/2) + KH_{(jk)|(jk)^c}/(2\sigma)\big) + \Phi\big(-\Phi^{-1}(1-\alpha/2) + KH_{(jk)|(jk)^c}/(2\sigma)\big)$, and for $\alpha \in [\delta, 1)$ with $\delta > 0$, we have

$$\lim_{n\to\infty} \inf_{\mathbf{\Theta}\in\mathcal{U}_1(K,\eta,s^*,M^*)} \big|\mathbb{P}_{\mathbf{\Theta}}\big(\Psi_S(\alpha) = 1\big)\big| = 1 \quad , \quad \textbf{if} \ \ \eta < 1/2. \tag{4.10}$$

More specifically, (4.8) implies that the score test $\Phi_S(\alpha)$ has no power beyond the type I error to distinguish $H_0$ from $H_{1n}$ if $\eta > 1/2$. In addition, when $\eta = 1/2$, since $\psi_\alpha > \alpha$ for any $K \neq 0$, (4.9) indicates that $\Psi_S(\alpha)$ has asymptotic power larger than the type I error for detecting the alternative hypothesis $H_{1n} : \Theta_{jk} = Kn^{-1/2}$. We plot $\psi_\alpha$ in Figure 1. We can see that the larger $K$ is, the larger power the pseudo score test can achieve. In other words, our proposed pseudo score test attains larger power when the magnitude of $\Theta^*_{jk}$ is large. Lastly, (4.10) implies that, if $\eta < 1/2$, the minimal power of $\Psi_S(\alpha)$ goes to 1 as $n \to \infty$.

## 4.3 Theoretical Property of Confidence Interval

In this subsection, we present the main theoretical result of the confidence interval and the related Wald test.

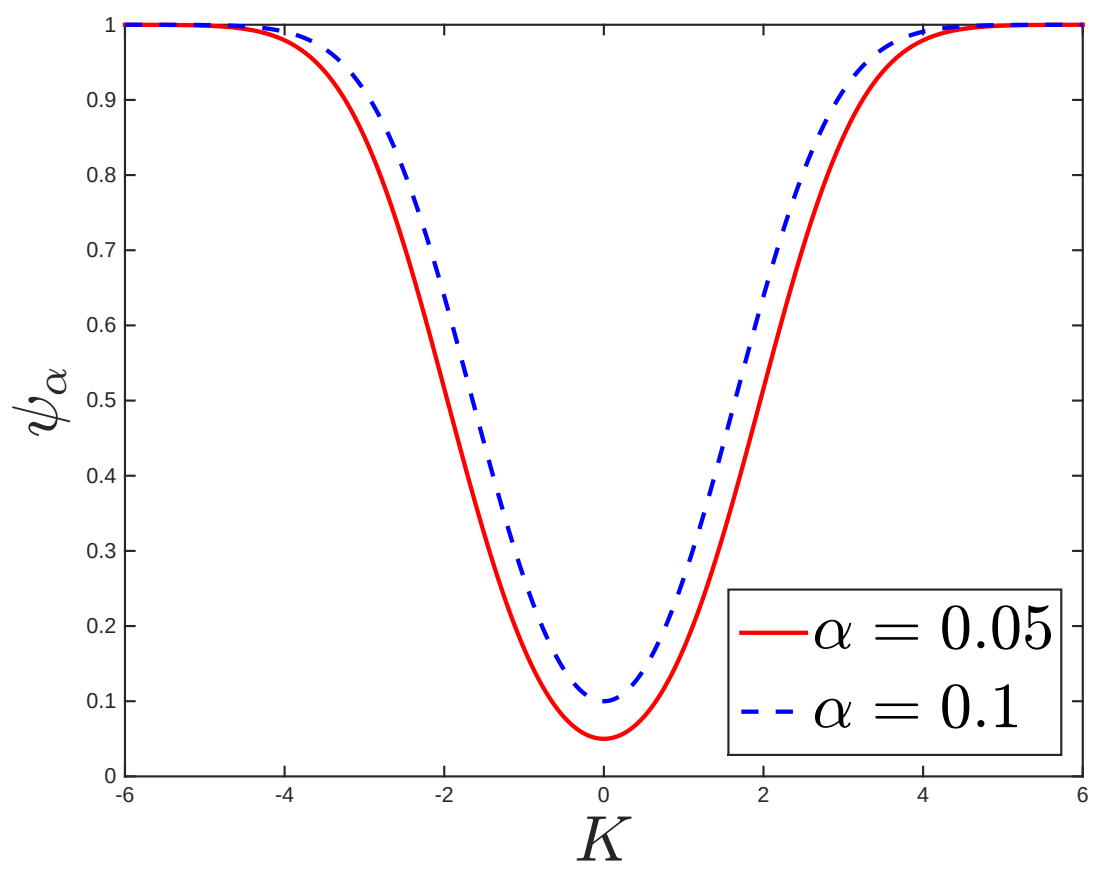

Figure 1: The plot of $\psi_\alpha = 1 - \Phi\big(\Phi^{-1}(1-\alpha/2) + K\big) + \Phi\big(-\Phi^{-1}(1-\alpha/2) + K\big)$, when $\alpha = 0.05$ and $\alpha = 0.1$. The x-axis is the value of $K$, and the y-axis is the value of $\psi_\alpha$.

**Theorem 4.11.** Under the same assumptions of Theorem 4.7, we have

$$\sqrt{n}(\widehat{\Theta}^W_{jk} - \Theta^*_{jk}) = -\sqrt{n} S_n(\boldsymbol{\Theta}^*) H^{-1}_{(jk)|(jk)^c} + o_{\mathbb{P}}(1) \rightsquigarrow N(0, 4\sigma^2 \cdot H^{-2}_{(jk)|(jk)^c}),$$

where $H_{(jk)|(jk)^c} = H_{(jk),(jk)} - \mathbf{H}_{(jk),(jk)^c}[\mathbf{H}_{(jk)^c,(jk)^c}]^{-1}\mathbf{H}_{(jk)^c,(jk)}$ is the partial information matrix for $\boldsymbol{\Theta}^*_{jk}$.

The asymptotic variance of the pseudo-likelihood estimator is $4\sigma^2 \cdot H^{-2}_{(jk)|(jk)^c}$, where both $\sigma^2$ and $H^{-2}_{(jk)|(jk)^c}$ are unknown. To apply Theorem 4.11, we need consistent estimators for $\sigma^2$ and $H_{(jk)|(jk)^c}$. For $\sigma^2$, we have already provided a consistent estimator $\widehat{\sigma}^2$ as in (4.3). For $H_{(jk)|(jk)^c}$, we use $\widehat{H}_{(jk)|(jk)^c} = 1/(\widehat{\Theta}_{jj}\widehat{\Theta}_{kk})$. To summarize, the following corollary indicates that the asymptotic variance of $\sqrt{n}(\widehat{\Theta}^W_{jk} - \Theta^*_{jk})$ can be consistently estimated.

**Corollary 4.12.** Under the same assumptions of Theorem 4.11, we have

$$(2\widehat{\sigma})^{-1}\widehat{H}_{(jk)|(jk)^c}\sqrt{n}(\widehat{\Theta}^W_{jk} - \Theta^*_{jk}) \rightsquigarrow N(0,1),$$

where $\widehat{H}_{(jk)|(jk)^c} = 1/(\widehat{\Theta}_{jj}\widehat{\Theta}_{kk})$.

Corollary 4.12 verifies the validity of the confidence interval (3.12). Furthermore, the following corollary shows that under $H_0$, type I error of the Wald test $\Psi_W(\alpha)$ defined in (3.14) converges to the significance level and the $p$-value is asymptotically uniformly distributed.

**Corollary 4.13.** Under the same assumptions of Theorem 4.11, we have

$$\lim_{n\to\infty} \mathbb{P}(\Psi_W(\alpha) = 1 | H_0) = \alpha \text{ and } P_W \rightsquigarrow U[0,1],$$

where $P_W = 2\big(1 - \Phi(|W_n|)\big)$ is the $p$-value of the Wald test.

### 4.4 Theoretical Property of Confidence Subgraph

We present the theoretical property of our confidence subgraph. Recall that the global test statistic in (3.15) is constructed based on Wald test statistic, so the global inference of confidence subgraph is closely related to the Wald test. As described in Section 3, we use the $\alpha$-quantile of $W$ conditioned on $\{\boldsymbol{X}_i\}_{i=1}^n$ as an estimator of the $\alpha$-quantile of $T$, i.e., $c_W(\alpha) = \inf\big\{t \in \mathbb{R} : \mathbb{P}_e(W \leq t) \geq \alpha\big\}$. We now lay out the main theorem for the confidence subgraph $\widehat{G}$.

**Theorem 4.14.** Under the same assumptions of Theorem 4.11, we have

$$\lim_{n\to\infty} \sup_{\alpha\in(0,1)} \big|\mathbb{P}(T \leq c_W(\alpha)) - \alpha\big| = 0,$$

and the confidence subgraph $\widehat{G}$ satisfies $\lim_{n\to\infty} \mathbb{P}(\widehat{G} \subseteq G^*) \geq 1 - \alpha$.

## 5 Numerical Experiments

We verify our theoretical result using both synthetic and real-world datasets empirically in this section. In all the experiments, we use the CLIME estimator implemented in the `R` package `clime` (ver. 0.4.1). For synthetic datasets, we use the `R` package `huge` (ver. 1.2.6) to generate samples with different graph structures.

### 5.1 Synthetic Datasets

In our numerical simulations, we consider 2 settings: (i) $n = 100, d = 100$; (ii) $n = 100, d = 400$, and generate data from nonparanormal distribution $\boldsymbol{X} \sim NPN(\mathbf{f}, \boldsymbol{\Sigma})$. For the monotonic transformations $\mathbf{f}$, we consider 2 settings: (i) extended square root function $f^{-1}(x) = (\mathrm{sign}(x)|x|^{1/2})/\sqrt{\int |t|\phi(t)dt}$; (ii) cubic function $f^{-1}(x) = (x^3)/\sqrt{\int t^6\phi(t)dt}$, where $\phi(t)$ is the marginal density function. For the choice of $\boldsymbol{\Sigma}$, we construct 3 different graph structures: scale-free graph, hub graph and band3 graph.

The detailed procedures for generating the 3 kinds of graphical models are as follows.

**Scale-free graph**. The degree distribution of the sacle-free graph follows a power law. The graph is generated by the preferential attachment mechanism. The graph begins with an initial small chain graph of 2 nodes. New nodes are added to the graph one at a time. Each new node is connected to one existing node with a probability that is proportional to the number of degrees that the existing node already has. Formally, the probability $p_i$ that the new node is connected to an existing node $i$ is, $p_i = k_i/(\sum_j k_j)$, where $k_i$ is the degree of the node $i$. The resulting graph has $d$ edges ($d = 100$ or $d = 400$). Once the graph is obtained, we generate an adjacency matrix $\mathbf{A}$ by setting the nonzero off-diagonal elements to be 0.3 and the diagonal elements to be 0. We calculate its smallest eigenvalue $\lambda_{\min}(\mathbf{A})$. The precision matrix is constructed as $\boldsymbol{\Theta} = \mathbf{A} + \big(|\lambda_{\min}(\mathbf{A})| + 0.2\big) \cdot \mathbf{I}_d$. The covariance matrix $\boldsymbol{\Sigma} := \boldsymbol{\Theta}^{-1}$ is then computed to generate the multivariate normal data: $\boldsymbol{x}_1, \ldots, \boldsymbol{x}_n \sim N_d(\mathbf{0}, \boldsymbol{\Sigma})$. The data are obtained by applying the marginal transformations to $\boldsymbol{x}_1, \ldots, \boldsymbol{x}_n$.

**Hub graph**. The $d$ nodes are evenly partitioned into $d/20$ disjoint groups with each group containing 20 nodes. Within each group, one node is selected as the hub and we add edges between the hub and the other 19 nodes in that group. The resulting graph has 190 edges when $d = 200$ and 380 edges when $d = 400$.

**Band3 graph**. Each node is associated with a coordinate $j$ with $j = 1, ..., d$. Two nodes are connected by an edge whenever the corresponding coordinates are at distance less than or equal to 3. The resulting graph has approximately 294 edges when $d = 100$ and $1, 194$ edges when $d = 400$.

We apply the proposed Score and Wald tests on the simulated nonparanormal data. In particular, We use CLIME for parameter estimation and use 5-fold cross validation to select the regularization parameter $\lambda$. We compute the type I errors at the 0.05 and 0.10 significance level, and compare them with the results given by the desparsity method (Jankova and van de Geer, 2013) which assumes that the data come from a Gaussian graphical model. The methods by Ren et al. (2013) and Liu (2013) yield similar results to Jankova and van de Geer (2013) and we only report the results by Jankova and van de Geer (2013). We repeat the tests for 500 times. Table 1 and Table 2 show the type I errors of Score, Wald tests and the desparsity method.

Table 1: Type I errors of Score, Wald and desparsity methods, with extended square root transformation function.

| | | scale-free | | hub | | band3 | |
|---|---|---|---|---|---|---|---|
| | significance level | $d = 100$ | $d = 400$ | $d = 100$ | $d = 400$ | $d = 100$ | $d = 400$ |
| Score | 0.05 | 0.055 | 0.060 | 0.050 | 0.045 | 0.050 | 0.055 |
| Wald | 0.05 | 0.045 | 0.045 | 0.050 | 0.055 | 0.050 | 0.045 |
| desparsity | 0.05 | 0.120 | 0.115 | 0.075 | 0.075 | 0.105 | 0.090 |
| Score | 0.10 | 0.090 | 0.115 | 0.105 | 0.095 | 0.100 | 0.095 |
| Wald | 0.10 | 0.095 | 0.110 | 0.100 | 0.105 | 0.090 | 0.100 |
| desparsity | 0.10 | 0.170 | 0.205 | 0.155 | 0.175 | 0.160 | 0.155 |

We now compare the power of the hypothesis tests at 0.05 significance level. We use the scale-free graph for a case study. Similar results can be observed in hub and band3 graphs as well. In particular, we generate the precision and covariance matrices by the same procedure as in scale-free model. To show the power curve, we randomly select one edge in the graph, and change its weight from 0 to 0.8 incrementally. Each time we generate $n = 200$ data samples from the nonparanormal graphical model with the extended square root transformation. Then we apply different test methods on the generated data. We repeat each experiment for 500 times and record the averaged power. The power curves are shown in Figure 2.

From these simulation results, we see that our methods achieve accurate type I errors and are more powerful than the desparsity method. In comparison, the desparsity method does not control the type I error well, which is not surprising because it is mainly designed for Gaussian graphical models. In summary, the proposed hypothesis testing methods yield accurate testing results, and outperform the existing methods.

Table 2: Type I errors of Score, Wald and desparsity methods, with cubic transformation function.

| | | scale-free | | hub | | band3 | |
|---|---|---|---|---|---|---|---|
| | significance level | $d = 100$ | $d = 400$ | $d = 100$ | $d = 400$ | $d = 100$ | $d = 400$ |
| Score | 0.05 | 0.055 | 0.055 | 0.045 | 0.055 | 0.050 | 0.055 |
| Wald | 0.05 | 0.040 | 0.045 | 0.050 | 0.066 | 0.050 | 0.060 |
| desparsity | 0.05 | 0.150 | 0.145 | 0.075 | 0.075 | 0.105 | 0.110 |
| Score | 0.10 | 0.095 | 0.115 | 0.105 | 0.095 | 0.100 | 0.095 |
| Wald | 0.10 | 0.095 | 0.110 | 0.110 | 0.100 | 0.095 | 0.090 |
| desparsity | 0.10 | 0.210 | 0.235 | 0.160 | 0.160 | 0.215 | 0.190 |

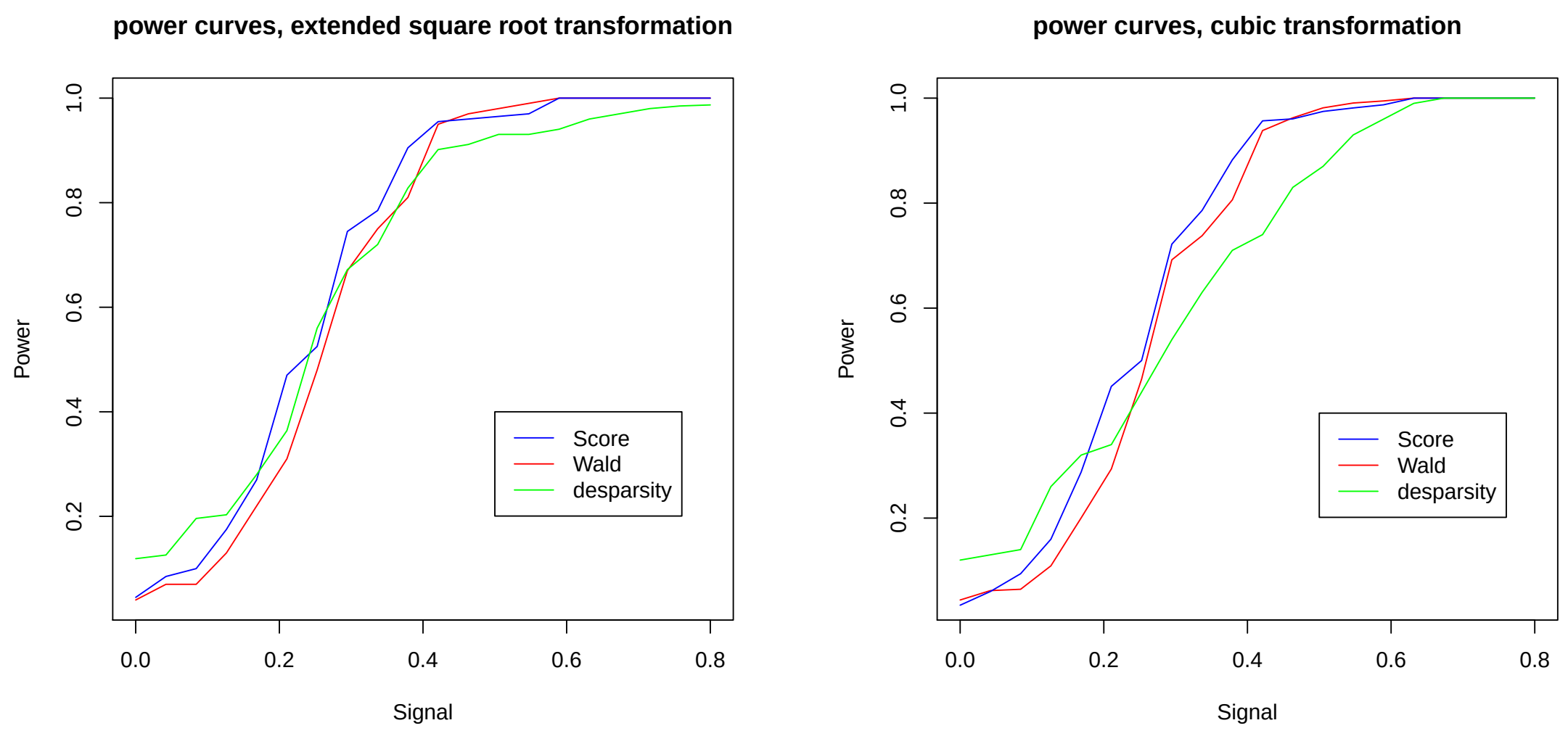

(a)

(b)

Figure 2: Power curves for testing $H_0 : \Theta_{jk} = 0$ at 0.05 significance level. (a) is for $d = 100$ and (b) is for $d = 200$.

### 5.2 Real-world Datasets

In this section, we apply our method to a human gene expression dataset. The dataset can be found in the `R` package `BDgraph`. It contains 60 unrelated individuals of Northern and Western European ancestry from Utah (CEU), whose genotypes are available from the Sanger

Institute website (ftp://ftp.sanger.ac.uk/pub/genevar). In this project, Illuminas Sentrix Human-6 Expression BeadChips are used to measure gene expression in Blymphocyte cells for all the individuals (Stranger et al., 2007). The genotypes for rare homozygous, heterozygous and homozygous common alleles are coded by 0, 1, and 2, respectively. The raw data are first background corrected and then quantile normalized across four replicates of a single individual. Finally, they are median normalized across all individuals. We choose the 100 most variable probes among the 47,293 probes corresponding to different Illumina TargetID. Each selected probe corresponds to a different transcript (Mohammadi and Wit, 2012). The more detailed discussion on this dataset can be found in Bhadra and Mallick (2013); Mohammadi and Wit (2012); Chen et al. (2008); Stranger et al. (2007). Our goal is to infer the significant interactions between gene transcripts.

We model the human gene expression dataset using the nonparanormal graphical model, with sample size $n = 60$ and dimension $d = 100$. We use the CLIME estimator with the regularization parameter selected by 5-fold cross validation for precision matrix estimation. For each gene pair, the null hypothesis is that these two variables are conditionally independent given the rest of variables, and we apply the score test with 0.05 significance level. We present the resulting graph in Figure 3. The $p$-values are shown on the edges. Most of the obtained edges meet the results in existing works (Bhadra and Mallick, 2013). For example, the connected component {GI_41190507S, Hs.512137S, Hs.512124S, Hs.449605S, GI_37546969S} reveals an interaction structure that is similar to the structure discovered by Bhadra and Mallick (2013). In addition, our method also finds some new gene-gene interactions, such as gene pairs hmm9615S and GI_21614524S, GI_21614524S and GI_16554578S.

We also construct the confidence subgraph for the gene expression dataset. The estimation procedure for $\mathbf{\Theta}$ is the same. We use the Monte-Carlo method to compute the quantile of the bootstrap estimator $W$ defined in (3.17). The random sampling is repeated for 1000 times. Figure 4 shows the confidence subgraph. In comparison with the local test result in Figure 3, the confidence subgraph has less edges. This is not surprising because the confidence interval of each $\Theta^*_{jk}$ given by the global inference is wider than the confidence interval given by the local inference due to the adjustment for the multiplicity of tests. Thus, more conservative results are given in Figure 4.

In conclusion, our method recovers many well-established human gene interactions and it also identifies some additional gene interactions potentially for future scientific investigations.

# 6 Conclusions

In this paper, we study both local and global asymptotic inference for the nonparanormal graphical model. For local inference, we aim at testing the presence of a single edge. In particular, we propose the score test and Wald test. We provide the theoretical guarantees on the type I error as well as the local power of the score test. We also construct the confidence interval for each edge based on the Wald test statistic. For global inference, we construct a confidence subgraph. The asymptotic property of the confidence subgraph is also established.

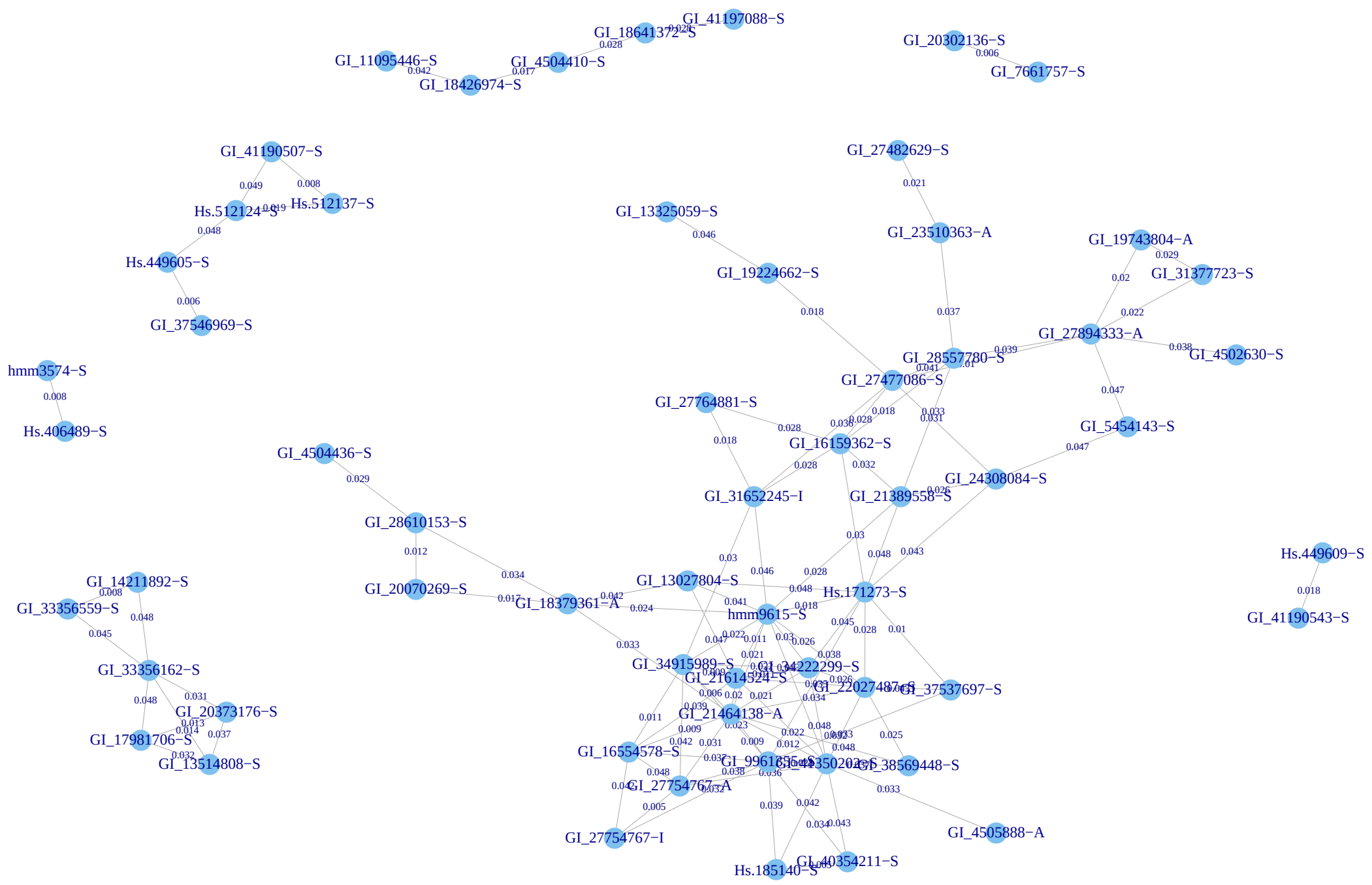

Figure 3: The inferred graph for the human gene expression data set. For each gene pair, we apply the score test with 0.05 significance level. The $p$-values are shown on the edges.

The numerical results show that our proposed methods outperform the existing ones.

## Acknowledgement

This research is partially supported by the grants NSF IIS1408910, NSF IIS1332109, NIH R01MH102339, NIH R01GM083084, and NIH R01HG06841.

## A Proof of Lemma 3.1

*Proof.* By permutation, we can have

$$\mathbf{H} = \begin{bmatrix} H_{(jk),(jk)} & \mathbf{H}_{(jk),(jk)^c} \\ \mathbf{H}_{(jk)^c,(jk)} & \mathbf{H}_{(jk)^c,(jk)^c} \end{bmatrix}.$$

In order to compute $\boldsymbol{\Omega} = \mathbf{H}^{-1}$, we can use the following blockwise matrix inversion identity (Golub and Loan, 1996):

$$\begin{bmatrix} \mathbf{A} & \mathbf{B} \\ \mathbf{C} & \mathbf{D} \end{bmatrix}^{-1} = \begin{bmatrix} (\mathbf{A}-\mathbf{B}\mathbf{D}^{-1}\mathbf{C})^{-1} & -(\mathbf{A}-\mathbf{B}\mathbf{D}^{-1}\mathbf{C})^{-1}\mathbf{B}\mathbf{D}^{-1} \\ -\mathbf{D}^{-1}\mathbf{C}(\mathbf{A}-\mathbf{B}\mathbf{D}^{-1}\mathbf{C})^{-1} & \mathbf{D}^{-1}+\mathbf{D}^{-1}(\mathbf{A}-\mathbf{B}\mathbf{D}^{-1}\mathbf{C})^{-1}\mathbf{B}\mathbf{D}^{-1} \end{bmatrix}. \quad \text{(A.1)}$$

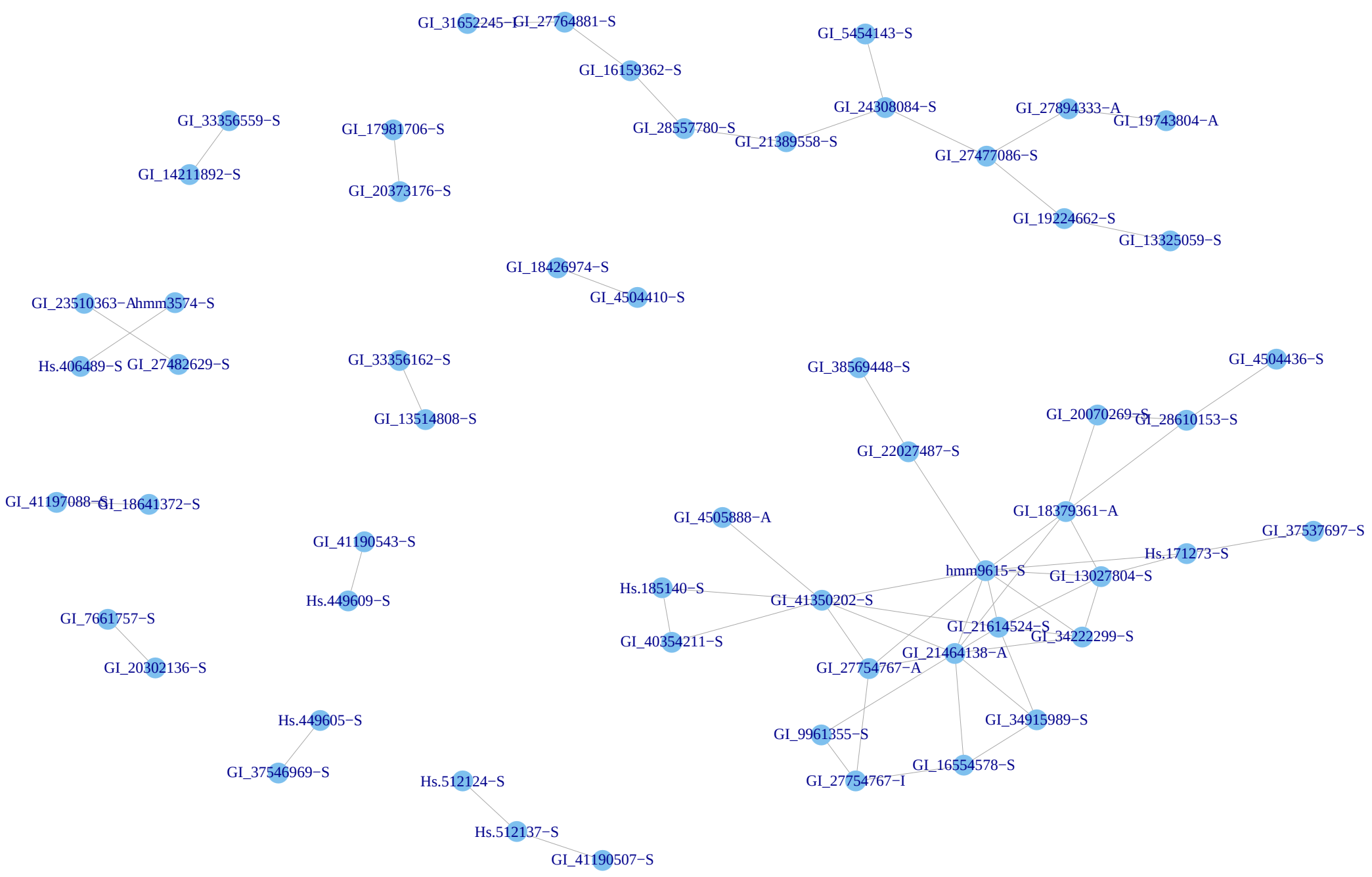

Figure 4: The graph for the human gene expression data set given by global graph inference with 0.05 significance level.

Let $\mathbf{A} = H_{(jk),(jk)}$, $\mathbf{B} = \mathbf{H}_{(jk),(jk)^c}$, $\mathbf{C} = \mathbf{H}_{(jk)^c,(jk)}$ and $\mathbf{D} = \mathbf{H}_{(jk)^c,(jk)^c}$. It can be seen from (A.1) that

$$\mathbf{\Omega}_{(jk),(jk)} = (H_{(jk),(jk)} - \mathbf{H}_{(jk),(jk)^c}[\mathbf{H}_{(jk)^c,(jk)^c}]^{-1}\mathbf{H}_{(jk)^c,(jk)})^{-1} \tag{A.2}$$

$$\mathbf{\Omega}_{(jk)^c,(jk)} = -[\mathbf{H}_{(jk)^c,(jk)^c}]^{-1}\mathbf{H}_{(jk)^c,(jk)}(H_{(jk),(jk)} - \mathbf{H}_{(jk),(jk)^c}[\mathbf{H}_{(jk)^c,(jk)^c}]^{-1}\mathbf{H}_{(jk)^c,(jk)})^{-1}. \tag{A.3}$$

Therefore, dividing (A.3) by (A.2), we obtain

$$\frac{\mathbf{\Omega}_{(jk)^c,(jk)}}{\mathbf{\Omega}_{(jk),(jk)}} = -[\mathbf{H}_{(jk)^c,(jk)^c}]^{-1}\mathbf{H}_{(jk)^c,(jk)} = -\mathbf{w}^*,$$

which immediately yields $\mathbf{w}^* = -\mathbf{\Omega}_{(jk)^c,(jk)}/\mathbf{\Omega}_{(jk),(jk)}$. Similarly, we can show that $\widehat{\mathbf{w}} = -\widehat{\mathbf{\Omega}}_{(jk)^c,(jk)}/\widehat{\mathbf{\Omega}}_{(jk),(jk)}$. □

# B Proof of Results in Section 4

## B.1 Proof of Theorem 4.7

In order to prove Theorem 4.7, we first lay out several key technical lemmas. The first lemma establishes the limiting distribution of the decorrelated score function $\sqrt{n}S_n(0, \mathbf{\Theta}^*_{(jk)^c})$ in the high-dimensional setting, which is asymptotically normal. It is the key for establishing the asymptotic normality of $\sqrt{n}\widehat{S}_n(0, \widehat{\mathbf{\Theta}}_{(jk)^c})$.

**Lemma B.1.** Under Assumptions 4.1 and 4.2, we have

$$\sqrt{n}S_n(0, \mathbf{\Theta}^*_{(jk)^c})/(2\sigma) \rightsquigarrow N(0,1),$$

where $\sigma^2 = \mathbf{R}_{(jk),(jk)} - 2\mathbf{R}_{(jk),(jk)^c}\mathbf{w}^* + \mathbf{w}^{*\top}\mathbf{R}_{(jk)^c,(jk)^c}\mathbf{w}^*$.

*Proof.* See Appendix C.1 for a detailed proof. □

Lemma B.1 not only shows the asymptotic distribution of $\sqrt{n}S_n(0, \mathbf{\Theta}^*_{(jk)^c})$ is normal, but also gives rise to the variance of the limiting distribution. In particular, the variance of the limiting normal distribution is determined by $\mathbf{R}$ and $\mathbf{w}^*$. The following lemma shows that the Hessian of the negative log-pseudo likelihood at $\widehat{\mathbf{\Theta}}$ is well-concentrated at the Hessian of the negative log-pseudo likelihood at $\mathbf{\Theta}^*$, in terms of matrix elementwise sup-norm.

**Lemma B.2.** Suppose $\mathbf{\Theta}^* \in \mathcal{U}(s, M)$. Under Assumptions 4.2 and 4.3, if $n$ is sufficiently large, we have

$$\big\|\widehat{\mathbf{\Theta}}^{-1} - \mathbf{\Theta}^{*-1}\big\|_{\max} = O_{\mathbb{P}}\Bigg(\sqrt{\frac{\log d}{n}}\Bigg) \quad \text{and} \quad \big\|\nabla^2\ell_n(\widehat{\mathbf{\Theta}}) - \mathbf{H}\big\|_{\max} = O_{\mathbb{P}}\Bigg(\sqrt{\frac{\log d}{n}}\Bigg).$$

*Proof.* See Appendix C.2 for a detailed proof. □

The following lemma provides an $\ell_1$ norm based estimation error for the estimator $\widehat{\mathbf{w}}(\widehat{\mathbf{\Theta}})$ in (3.6).

**Lemma B.3.** Suppose $\mathbf{\Theta}^* \in \mathcal{U}(s, M)$. Under Assumptions 4.2 and 4.3, if $n$ is sufficiently large, we have

$$\big\|\widehat{\mathbf{w}}(\widehat{\mathbf{\Theta}}) - \mathbf{w}^*\big\|_1 = O_{\mathbb{P}}\Bigg(s^2\frac{\log d}{n} + s\sqrt{\frac{\log d}{n}}\Bigg).$$

*Proof.* See Appendix C.3 for a detailed proof. □

In order to show the asymptotic normality of $\widehat{S}_n(\widehat{\mathbf{\Theta}})$, we also need Assumption 4.5, which together with Lemma B.2 and B.3 guarantee that $\sqrt{n}\big(\widehat{S}_n(0, \widehat{\mathbf{\Theta}}_{(jk)^c}) - S_n(0, \mathbf{\Theta}^*_{(jk)^c})\big) = o_{\mathbb{P}}(1)$.

*Proof of Theorem 4.7.* In this proof, we use $\widehat{\mathbf{w}}$ as the shorthand for $\widehat{\mathbf{w}}_{(jk)}(\widehat{\mathbf{\Theta}})$. We have

$$\begin{aligned}
\widehat{S}_n\big(0, \widehat{\mathbf{\Theta}}_{(jk)^c}\big) &= -\widehat{\Sigma}_{jk} + [\widehat{\mathbf{\Theta}}^{-1}]_{jk} - \widehat{\mathbf{w}}^\top \mathrm{vec}\big(-\widehat{\mathbf{\Sigma}}_{(jk)^c} + [\widehat{\mathbf{\Theta}}^{-1}]_{(jk)^c}\big) \\
&= -\widehat{\Sigma}_{jk} + [\mathbf{\Theta}^{*-1}]_{jk} - (\mathbf{w}^*)^\top \mathrm{vec}\big(-\widehat{\mathbf{\Sigma}}_{(jk)^c} + [\mathbf{\Theta}^{*-1}]_{(jk)^c}\big) - [\mathbf{\Theta}^{*-1}]_{jk} + [\widehat{\mathbf{\Theta}}^{-1}]_{jk} \\
&\quad + (\mathbf{w}^* - \widehat{\mathbf{w}})^\top \mathrm{vec}\big(-\widehat{\mathbf{\Sigma}}_{(jk)^c} + [\widehat{\mathbf{\Theta}}^{-1}]_{(jk)^c}\big) - (\mathbf{w}^*)^\top \mathrm{vec}\big([\widehat{\mathbf{\Theta}}^{-1}]_{(jk)^c} - [\mathbf{\Theta}^{*-1}]_{(jk)^c}\big).
\end{aligned}$$

The right hand side of the above equation can be reorganized as follows

$$\widehat{S}_n\big(0,\widehat{\boldsymbol{\Theta}}_{(jk)^c}\big)=S_n\big(0,\boldsymbol{\Theta}^*_{(jk)^c}\big)+\underbrace{(\mathbf{w}^*-\widehat{\mathbf{w}})^\top\mathrm{vec}\big(-\widehat{\boldsymbol{\Sigma}}_{(jk)^c}+[\widehat{\boldsymbol{\Theta}}^{-1}]_{(jk)^c}\big)}_{\text{(i)}}$$
$$\underbrace{-[\boldsymbol{\Theta}^{*-1}]_{jk}+[\widehat{\boldsymbol{\Theta}}^{-1}]_{jk}-(\mathbf{w}^*)^\top\mathrm{vec}\big([\widehat{\boldsymbol{\Theta}}^{-1}]_{(jk)^c}-[\boldsymbol{\Theta}^{*-1}]_{(jk)^c}\big)}_{\text{(ii)}}. \tag{B.1}$$

In the following, we are going to bound terms (i) and (ii) respectively.
**Bounding term** (i)**:** We have that

$$\text{(i)}\le\|\widehat{\mathbf{w}}-\mathbf{w}^*\|_1\cdot\big\|\mathrm{vec}(-\widehat{\boldsymbol{\Sigma}}_{(jk)^c}+[\widehat{\boldsymbol{\Theta}}^{-1}]_{(jk)^c})\big\|_\infty\le\|\widehat{\mathbf{w}}-\mathbf{w}^*\|_1\cdot\big\|-\widehat{\boldsymbol{\Sigma}}+\widehat{\boldsymbol{\Theta}}^{-1}\big\|_{\max}, \tag{B.2}$$

where the first inequality follows from Hölder's inequality, and the second inequality follows from $\|\mathrm{vec}(\mathbf{A})\|_\infty=\|\mathbf{A}\|_{\max}$. For the right hand side of (B.2), by Lemma B.3, we have

$$\|\widehat{\mathbf{w}}-\mathbf{w}^*\|_1=O_{\mathbb{P}}\bigg(s^2\frac{\log d}{n}+s\sqrt{\frac{\log d}{n}}\bigg). \tag{B.3}$$

Furthermore, we have $\big\|-\widehat{\boldsymbol{\Sigma}}+\widehat{\boldsymbol{\Theta}}^{-1}\big\|_{\max}\le\big\|-\widehat{\boldsymbol{\Sigma}}+\boldsymbol{\Sigma}^*\big\|_{\max}+\big\|-\boldsymbol{\Theta}^{*-1}+\widehat{\boldsymbol{\Theta}}^{-1}\big\|_{\max}$. Since $\big\|-\widehat{\boldsymbol{\Sigma}}+\boldsymbol{\Sigma}^*\big\|_{\max}=O_{\mathbb{P}}\big(\sqrt{\log d/n}\big)$, together with Lemma B.2, we can obtain that

$$\big\|-\widehat{\boldsymbol{\Sigma}}+\widehat{\boldsymbol{\Theta}}^{-1}\big\|_{\max}=O_{\mathbb{P}}\big(\sqrt{\log d/n}\big). \tag{B.4}$$

Furthermore, plugging (B.3) and (B.4) into (B.2), we get

$$\text{(i)}=O_{\mathbb{P}}\bigg(s^2\Big(\frac{\log d}{n}\Big)^{3/2}+s\frac{\log d}{n}\bigg)=o_{\mathbb{P}}(n^{-1/2}), \tag{B.5}$$

according to Assumption 4.5.
**Bounding term** (ii)**:** Consider the Taylor expansion of $\widehat{\boldsymbol{\Theta}}$ at $\boldsymbol{\Theta}^*$, we have

$$\begin{aligned}\widehat{\boldsymbol{\Theta}}^{-1}&=\boldsymbol{\Theta}^{*-1}-\boldsymbol{\Theta}^{*-1}\widehat{\boldsymbol{\Delta}}\boldsymbol{\Theta}^{*-1}-(\boldsymbol{\Theta}^{*-1}\widehat{\boldsymbol{\Delta}})^2\sum_{k=0}^{\infty}(-1)^k(\boldsymbol{\Theta}^{*-1}\widehat{\boldsymbol{\Delta}})^k\boldsymbol{\Theta}^{*-1}\\&=\boldsymbol{\Theta}^{*-1}-\boldsymbol{\Theta}^{*-1}\widehat{\boldsymbol{\Delta}}\boldsymbol{\Theta}^{*-1}-\mathbf{R}_1(\widehat{\boldsymbol{\Delta}}),\end{aligned} \tag{B.6}$$

where $\mathbf{R}_1(\widehat{\boldsymbol{\Delta}})=(\boldsymbol{\Theta}^{*-1}\widehat{\boldsymbol{\Delta}})^2\mathbf{J}\boldsymbol{\Theta}^{*-1}=(\boldsymbol{\Theta}^{*-1}\widehat{\boldsymbol{\Delta}})\mathbf{R}(\widehat{\boldsymbol{\Delta}})$, $\mathbf{R}(\widehat{\boldsymbol{\Delta}})=\boldsymbol{\Theta}^{*-1}\widehat{\boldsymbol{\Delta}}\mathbf{J}\boldsymbol{\Theta}^{*-1}=\widehat{\boldsymbol{\Theta}}^{-1}-\boldsymbol{\Theta}^{*-1}$ and $\mathbf{J}=\sum_{k=0}^{\infty}(-1)^k(\boldsymbol{\Theta}^{*-1}\widehat{\boldsymbol{\Delta}})^k$. By (B.6), we have

$$\begin{aligned}\big[\widehat{\boldsymbol{\Theta}}^{-1}-\boldsymbol{\Theta}^{*-1}\big]_{jk}&=-[\boldsymbol{\Theta}^{*-1}]_{j*}\widehat{\boldsymbol{\Delta}}[\boldsymbol{\Theta}^{*-1}]_{*k}-\big[\mathbf{R}_1(\widehat{\boldsymbol{\Delta}})\big]_{jk}\\&=-\big([\boldsymbol{\Theta}^{*-1}]_{k*}\otimes[\boldsymbol{\Theta}^{*-1}]_{j*}\big)\mathrm{vec}(\widehat{\boldsymbol{\Delta}})-\big[\mathbf{R}_1(\widehat{\boldsymbol{\Delta}})\big]_{jk}\\&=-(\boldsymbol{\Theta}^{*-1}\otimes\boldsymbol{\Theta}^{*-1})_{(jk),*}\mathrm{vec}(\widehat{\boldsymbol{\Delta}})-\big[\mathbf{R}_1(\widehat{\boldsymbol{\Delta}})\big]_{jk},\end{aligned} \tag{B.7}$$

where the second equality follows from $\mathbf{a}^\top\mathbf{B}\mathbf{c} = (\mathbf{c}^\top \otimes \mathbf{a}^\top)\text{vec}(\mathbf{B})$. Since $\widehat{\boldsymbol{\Delta}}_{(jk)} = 0$, the right hand side of (B.7) can be further rewritten as

$$[\widehat{\boldsymbol{\Theta}}^{-1} - \boldsymbol{\Theta}^{*-1}]_{jk} = -(\boldsymbol{\Theta}^{*-1} \otimes \boldsymbol{\Theta}^{*-1})_{(jk),(jk)^c}\text{vec}(\widehat{\boldsymbol{\Delta}}_{(jk)^c}) - \big[\mathbf{R}_1(\widehat{\boldsymbol{\Delta}})\big]_{jk}. \tag{B.8}$$

By (B.6), we can also get

$$\begin{aligned}\text{vec}\big([\widehat{\boldsymbol{\Theta}}^{-1} - \boldsymbol{\Theta}^{*-1}]_{(jk)^c}\big) &= -\text{vec}\big([\boldsymbol{\Theta}^{*-1}\widehat{\boldsymbol{\Delta}}\boldsymbol{\Theta}^{*-1}]_{(jk)^c}\big) - \text{vec}\big([\mathbf{R}_1(\widehat{\boldsymbol{\Delta}})]_{(jk)^c}\big)\\ &= -(\boldsymbol{\Theta}^{*-1} \otimes \boldsymbol{\Theta}^{*-1})_{(jk)^c,*}\text{vec}(\widehat{\boldsymbol{\Delta}}) - \text{vec}\big([\mathbf{R}_1(\widehat{\boldsymbol{\Delta}})]_{(jk)^c}\big), \end{aligned} \tag{B.9}$$

where the second equality follows from $\text{vec}(\mathbf{ABC}) = (\mathbf{C}^\top \otimes \mathbf{A})\text{vec}(\mathbf{B})$. Again, since $\widehat{\boldsymbol{\Delta}}_{(jk)} = 0$, the right hand side of (B.9) can be further rewritten as

$$\text{vec}\big([\widehat{\boldsymbol{\Theta}}^{-1} - \boldsymbol{\Theta}^{*-1}]_{(jk)^c}\big) = -(\boldsymbol{\Theta}^{*-1} \otimes \boldsymbol{\Theta}^{*-1})_{(jk)^c,(jk)^c}\text{vec}(\widehat{\boldsymbol{\Delta}}_{(jk)^c}) - \text{vec}\big([\mathbf{R}_1(\widehat{\boldsymbol{\Delta}})]_{(jk)^c}\big). \tag{B.10}$$

Combining (B.8) and (B.10), we obtain

$$\begin{aligned}\text{(ii)} &= -(\boldsymbol{\Theta}^{*-1} \otimes \boldsymbol{\Theta}^{*-1})_{(jk),(jk)^c}\text{vec}(\widehat{\boldsymbol{\Delta}}_{(jk)^c}) - \big[\mathbf{R}_1(\widehat{\boldsymbol{\Delta}})\big]_{jk}\\ &\quad - \big\langle \mathbf{w}^*, -(\boldsymbol{\Theta}^{*-1} \otimes \boldsymbol{\Theta}^{*-1})_{(jk)^c,(jk)^c}\text{vec}(\widehat{\boldsymbol{\Delta}}_{(jk)^c}) - \text{vec}\big([\mathbf{R}_1(\widehat{\boldsymbol{\Delta}})]_{(jk)^c}\big)\big\rangle\\ &= \underbrace{-\mathbf{H}_{(jk),(jk)^c}\text{vec}(\widehat{\boldsymbol{\Delta}}_{(jk)^c}) - \big\langle \mathbf{w}^*, -\mathbf{H}_{(jk)^c,(jk)^c}\text{vec}(\widehat{\boldsymbol{\Delta}}_{(jk)^c})\big\rangle}_{\text{(ii).a}} - \underbrace{[\mathbf{R}_1(\widehat{\boldsymbol{\Delta}})]_{jk}}_{\text{(ii).b}}\\ &\quad + \underbrace{\big\langle \mathbf{w}^*, \text{vec}\big([\mathbf{R}_1(\widehat{\boldsymbol{\Delta}})]_{(jk)^c}\big)\big\rangle}_{\text{(ii).c}}. \end{aligned}$$

By the definition of $\mathbf{w}^* = [\mathbf{H}_{(jk)^c,(jk)^c}]^{-1}\mathbf{H}_{(jk),(jk)^c}$, we have (ii).a $= 0$. It remains to bound terms (ii).b and (ii).c. Recall that $\mathbf{R}_1(\widehat{\boldsymbol{\Delta}}) = (\boldsymbol{\Theta}^{*-1}\widehat{\boldsymbol{\Delta}})^2\mathbf{J}\boldsymbol{\Theta}^{*-1} = (\boldsymbol{\Theta}^{*-1}\widehat{\boldsymbol{\Delta}})\mathbf{R}(\widehat{\boldsymbol{\Delta}})$. In the proof of Lemma B.2, we have shown that

$$\big\|\mathbf{R}(\widehat{\boldsymbol{\Delta}})\big\|_{\max} \le K_{\boldsymbol{\Sigma}^*}^2/2\big\|\widehat{\boldsymbol{\Delta}}\big\|_{\max}.$$

Therefore, We have

$$\begin{aligned}\text{(ii).b} &\le \|\mathbf{R}_1(\widehat{\boldsymbol{\Delta}})\|_{\max} = \big\|(\boldsymbol{\Theta}^{*-1}\widehat{\boldsymbol{\Delta}})\mathbf{R}(\widehat{\boldsymbol{\Delta}})\big\|_{\max} \le \|\boldsymbol{\Theta}^{*-1}\widehat{\boldsymbol{\Delta}}\|_\infty \cdot \|\mathbf{R}(\widehat{\boldsymbol{\Delta}})\|_{\max}\\ &\le \|\boldsymbol{\Theta}^{*-1}\|_\infty \cdot \|\widehat{\boldsymbol{\Delta}}\|_\infty \cdot \|\mathbf{R}(\widehat{\boldsymbol{\Delta}})\|_{\max} \le K_{\boldsymbol{\Sigma}^*}^3/2\|\widehat{\boldsymbol{\Delta}}\|_\infty \cdot \|\widehat{\boldsymbol{\Delta}}\|_{\max}, \end{aligned} \tag{B.11}$$

where the second inequality follows from $\|\mathbf{AB}\|_{\max} \le \|\mathbf{A}\|_\infty \cdot \|\mathbf{B}\|_{\max}$, the third inequality is due to $\|\mathbf{AB}\|_\infty \le \|\mathbf{A}\|_\infty \cdot \|\mathbf{B}\|_\infty$. and the forth inequality follows from Assumption 4.2. Since $\|\widehat{\boldsymbol{\Delta}}\|_\infty = \|\widehat{\boldsymbol{\Delta}}\|_1 = O_{\mathbb{P}}(s\sqrt{\log d/n})$ and $\|\widehat{\boldsymbol{\Delta}}\|_{\max} = O_{\mathbb{P}}(\sqrt{\log d/n})$, we have (ii).b $= O_{\mathbb{P}}\big(s\log d/n\big)$. Similarly, we can show that

$$\text{(ii).c} \le \|\mathbf{w}^*\|_1 \cdot \big\|[\mathbf{R}_1(\widehat{\boldsymbol{\Delta}})]_{(jk)^c}\big\|_\infty \le \|\mathbf{w}^*\|_1 \cdot \|\mathbf{R}_1(\widehat{\boldsymbol{\Delta}})\|_{\max} = O_{\mathbb{P}}(s\log d/n),$$

since

$$\|\mathbf{w}^*\|_1 = \left\| \frac{\boldsymbol{\Omega}_{(jk)^c,(jk)}}{\boldsymbol{\Omega}_{(jk),(jk)}} \right\|_1 = \frac{\big\|(\boldsymbol{\Theta}^* \otimes \boldsymbol{\Theta}^*)_{(jk)^c,(jk)}\big\|_1}{\big|(\boldsymbol{\Theta}^* \otimes \boldsymbol{\Theta}^*)_{(jk),(jk)}\big|} \leq \frac{\|\boldsymbol{\Theta}^*_{*j} \otimes \boldsymbol{\Theta}^*_{*k}\|_1}{|\Theta^*_{jj}\Theta^*_{kk}|} \leq \frac{\|\boldsymbol{\Theta}^*\|_1^2}{\lambda_{\min}(\boldsymbol{\Theta}^*)^2} \leq \nu^2 M^2,$$

where last inequality follows from $\boldsymbol{\Theta}^* \in \mathcal{U}(s, M)$ and Assumption 4.2. Combining terms (ii).a, (ii).b and (ii).c, we obtain

$$\text{(ii)} = O_{\mathbb{P}}\big(s \log d/n\big) = o_{\mathbb{P}}\big(n^{-1/2}\big), \tag{B.12}$$

according to Assumption 4.5. Therefore, substituting (B.5) and (B.12) into (B.1), we get $\sqrt{n}\widehat{S}_n\big(0, \widehat{\boldsymbol{\Theta}}_{(jk)^c}\big) = \sqrt{n}S_n\big(0, \boldsymbol{\Theta}^*_{(jk)^c}\big) + o_{\mathbb{P}}(1)$. Combining with Lemma B.1, we complete the proof. □

## B.2 Consistency of $\widehat{\sigma}^2$

In this subsection, we show that $\widehat{\sigma}^2$ in (4.3) is a consistent estimator of $\sigma^2$. This, together with Theorem 4.7, establishes the validity of the score test statistic in (3.9).

**Theorem B.4.** For $\sigma^2$ and $\widehat{\sigma}^2$ as defined in Theorem 4.7 and (4.3), under Assumptions 4.1, 4.2 and 4.3, we have

$$|\widehat{\sigma}^2 - \sigma^2| = O_{\mathbb{P}}\left(s^2\left(\frac{\log d}{n}\right)^{3/2} + s^2\frac{\log d}{n} + s\sqrt{\frac{\log d}{n}}\right).$$

Before we prove Theorem B.4, we need the following auxiliary lemma.

**Lemma B.5.** Under Assumptions 4.1, 4.2 and 4.3, we have

$$\big\|\widehat{\mathbf{R}} - \mathbf{R}\big\|_{\max} = O_{\mathbb{P}}\left(\sqrt{\frac{\log d}{n}}\right). \tag{B.13}$$

*Proof.* See Appendix C.4 for a detailed proof. □

*Proof of Theorem B.4.* We have

$$|\widehat{\sigma}^2 - \sigma^2| \leq \underbrace{|\widehat{\mathbf{R}}_{(jk),(jk)} - \mathbf{R}_{(jk),(jk)}|}_{\text{(i)}} + 2\underbrace{|\widehat{\mathbf{R}}_{(jk),(jk)^c}\widehat{\mathbf{w}} - \mathbf{R}_{(jk),(jk)^c}\mathbf{w}^*|}_{\text{(ii)}}$$
$$+ \underbrace{|\widehat{\mathbf{w}}^{\top}\widehat{\mathbf{R}}_{(jk)^c,(jk)^c}\widehat{\mathbf{w}} - \mathbf{w}^{*\top}\mathbf{R}_{(jk)^c,(jk)^c}\mathbf{w}^*|}_{\text{(iii)}}.$$

It remains to bound terms (i), (ii) and (iii) respectively.
**Bounding term** (i)**:** By Lemma B.5, we have (i) $= O_{\mathbb{P}}(\sqrt{\log d/n})$.

**Bounding term** (ii)**:** By the triangle inequality, term (ii) can be upper bounded by

$$\underbrace{\big|(\widehat{\mathbf{R}}_{(jk),(jk)^c} - \mathbf{R}_{(jk),(jk)^c})(\widehat{\mathbf{w}} - \mathbf{w}^*)\big|}_{\text{(ii).a}} + \underbrace{\big|\mathbf{R}_{(jk),(jk)^c}(\widehat{\mathbf{w}} - \mathbf{w}^*)\big|}_{\text{(ii).b}} + \underbrace{\big|(\widehat{\mathbf{R}}_{(jk),(jk)^c} - \mathbf{R}_{(jk),(jk)^c})\mathbf{w}^*\big|}_{\text{(ii).c}}.$$

We are going to bound terms (ii).a, (ii).b and (ii).c respectively. Term (ii).a is bounded by

$$\|\widehat{\mathbf{R}}_{(jk),(jk)^c} - \mathbf{R}_{(jk),(jk)^c}\|_\infty \cdot \|\widehat{\mathbf{w}} - \mathbf{w}^*\|_1 = O_{\mathbb{P}}\bigg(s^2\Big(\frac{\log d}{n}\Big)^{3/2} + s\frac{\log d}{n}\bigg),$$

where the equality follows from Lemma B.5 and Lemma B.3. To bound term (ii).b, we have

$$\text{(ii).b} \le \|\mathbf{R}_{(jk),(jk)^c}\|_\infty \cdot \|\widehat{\mathbf{w}} - \mathbf{w}^*\|_1 = O_{\mathbb{P}}\bigg(s^2\frac{\log d}{n} + s\sqrt{\frac{\log d}{n}}\bigg),$$

since $\|\mathbf{R}_{(jk),(jk)^c}\|_\infty = O_{\mathbb{P}}(1)$. To see this, we note that each element of $\mathbf{R}$ is a continuous function of the elements of $\boldsymbol{\Sigma}$, and each element of $\boldsymbol{\Sigma}$ is bounded. To bound term (ii).c, we have

$$\text{(ii).c} \le \|\mathbf{w}^*\|_1 \cdot \|\widehat{\mathbf{R}}_{(jk),(jk)^c} - \mathbf{R}_{(jk),(jk)^c}\|_\infty = O_{\mathbb{P}}\big(\sqrt{\log d/n}\big),$$

because $\|\mathbf{w}^*\|_1 \le \nu^2 M^2$. It follows immediately that (ii) $= O_{\mathbb{P}}\big(s^2\big(\log d/n\big)^{3/2} + s\log d/n\big)$.

**Bounding term** (iii)**:** We have

$$\text{(iii)} \le \underbrace{\big|\widehat{\mathbf{w}}^\top\big(\widehat{\mathbf{R}}_{(jk)^c,(jk)^c} - \mathbf{R}_{(jk)^c,(jk)^c}\big)\widehat{\mathbf{w}}\big|}_{\text{(iii).a}} + \underbrace{\big|\widehat{\mathbf{w}}^\top\mathbf{R}_{(jk)^c,(jk)^c}\widehat{\mathbf{w}} - \mathbf{w}^{*\top}\mathbf{R}_{(jk)^c,(jk)^c}\mathbf{w}^*\big|}_{\text{(iii).b}}.$$

For term (iii).a, we have

$$\text{(iii).a} \le \|\widehat{\mathbf{w}}\|_1^2 \cdot \|\widehat{\mathbf{R}}_{(jk)^c,(jk)^c} - \mathbf{R}_{(jk)^c,(jk)^c}\|_{\max} = O_{\mathbb{P}}\bigg(\sqrt{\frac{\log d}{n}}\bigg),$$

because $\|\mathbf{w}^*\|_1 \le \nu^2 M^2$. For term (iii).b, using the fact that

$$|\widehat{\mathbf{v}}^\top\mathbf{W}\widehat{\mathbf{v}} - \mathbf{v}^\top\mathbf{W}\mathbf{v}| \le \|\mathbf{W}\|_{\max} \cdot \|\widehat{\mathbf{v}} - \mathbf{v}\|_1^2 + 2\|\mathbf{W}\|_{\max} \cdot \|\mathbf{v}\|_1 \cdot \|\widehat{\mathbf{v}} - \mathbf{v}\|_1$$

for any symmetric matrix $\mathbf{W} \in \mathbb{R}^{d\times d}$ and $\widehat{\mathbf{v}}, \mathbf{v} \in \mathbb{R}^d$, we obtain

$$\begin{aligned}\text{(iii).b} &\le \|\mathbf{R}_{(jk)^c,(jk)^c}\|_{\max} \cdot \|\widehat{\mathbf{w}} - \mathbf{w}^*\|_1^2 + 2\|\mathbf{R}_{(jk),(jk)^c}\|_\infty \cdot \|\mathbf{w}^*\|_1 \cdot \|\widehat{\mathbf{w}} - \mathbf{w}^*\|_1 \\ &= O_{\mathbb{P}}\bigg(s^2\frac{\log d}{n} + s\sqrt{\frac{\log d}{n}}\bigg),\end{aligned}$$

because $\|\mathbf{R}_{(jk)^c,(jk)^c}\|_{\max} = O_{\mathbb{P}}(1)$, $\|\mathbf{R}_{(jk),(jk)^c}\|_\infty = O_{\mathbb{P}}(1)$ and $\|\mathbf{w}^*\|_1 \le \nu^2 M^2$. It follows

that

$$\text{(iii)} = O_{\mathbb{P}}\bigg(s^2\frac{\log d}{n} + s\sqrt{\frac{\log d}{n}}\bigg). \tag{B.14}$$

Combining terms (i), (ii) and (iii) together, we complete the proof. □

## B.3 Proof of Corollary 4.8

*Proof.* By Theorem 4.7 and Theorem B.4, we have

$$\mathrm{ST}_n = \sqrt{n}\widehat{S}_n\big(0, \mathbf{\Theta}_{(jk)^c}\big)/\widehat{\sigma} \rightsquigarrow N(0,1).$$

Thus, we have$\mathbb{P}\big(\Psi_S(\alpha) = 1 | H_0\big) = \mathbb{P}\big(\mathrm{ST}_n^2 \geq \chi^2_{1\alpha}\big) = \alpha$. Similarly, for any $t \in (0,1)$, we have

$$\mathbb{P}(P_S < t) = \mathbb{P}\bigg(\Phi(\mathrm{ST}_n) > 1 - \frac{t}{2}\bigg) = \mathbb{P}\bigg(\mathrm{ST}_n > \Phi^{-1}\Big(1 - \frac{t}{2}\Big)\bigg) \to t.$$

This completes the proof. □

## B.4 Proof of Theorem 4.10

In order to prove Theorem 4.10, we first lay out several auxiliary lemmas as follows.

First, we have the following result, which essentially guarantees uniformly local asymptotic normality condition for $\Theta_{jk}$.

**Lemma B.6.** Under Assumption 4.2, we have

$$\lim_{n\to\infty} \inf_{\mathbf{\Theta}\in\mathcal{U}_1} \mathbb{P}_{\mathbf{\Theta}}\Bigg[\sqrt{n}\Big|S_n\big(\Theta_{jk}, \mathbf{\Theta}_{(jk)^c}\big) - S_n\big(0, \mathbf{\Theta}_{(jk)^c}\big) - \Theta_{jk}H_{(jk)|(jk)^c}\Big| \\ \leq C_0 K n^{1/2-\eta}\sqrt{\frac{\log d}{n}}\Bigg] = 1,$$

where $\mathcal{U}_1 = \mathcal{U}_1(K, \eta, s^*, M^*)$, $H_{(jk)|(jk)^c} = \big[\Theta_{jj}\Theta_{kk} + \Theta_{jk}^2\big]/\big(\Theta_{jj}\Theta_{kk}\big)^2$ and $C_0 > 0$ is a positive constant.

*Proof.* See Appendix C.5 for a detailed proof. □

**Lemma B.7.** Under Assumptions 4.1 and 4.2, we have

$$\lim_{n\to\infty} \inf_{\mathbf{\Theta}\in\mathcal{U}_1(K,\eta,s^*,M^*)} \mathbb{P}_{\mathbf{\Theta}}\Bigg(\|\widehat{\mathbf{R}} - \mathbf{R}\|_{\max} \leq C_8\sqrt{\frac{\log d}{n}}\Bigg) = 1, \tag{B.15}$$

where $C_8$ is a positive constant.

*Proof.* See Appendix C.6 for a detailed proof. □

In addition, we have the uniform central limit theorem holds in $\mathcal{U}_1(K, \eta, s^*, M^*)$.

**Lemma B.8.** Under Assumptions 4.1 and 4.2, we have

$$\lim_{n\to\infty} \sup_{\boldsymbol{\Theta}\in\mathcal{U}_1(K,\eta,s^*,M^*)} \sup_{t\in\mathbb{R}} \left| \mathbb{P}_{\boldsymbol{\Theta}}\left( \frac{\sqrt{n}}{2} \big(\mathbf{b}^\top \mathbf{R}\mathbf{b}\big)^{-\frac{1}{2}} \mathbf{b}^\top \mathrm{vec}\big(\nabla \ell_n(\boldsymbol{\Theta})\big) \right) - \Phi(t) \right| = 0,$$

where $\mathbf{b} = [1, -(\mathbf{w})^\top]^\top$.

*Proof.* See Appendix C.7 for a detailed proof. □

We have that the estimation error bound of $\widehat{\mathbf{w}}(\widehat{\boldsymbol{\Theta}})$ holds uniformly in for any $\boldsymbol{\Theta}$ in $\mathcal{U}_1(K,\eta,s^*,M^*)$.

**Lemma B.9.** Under Assumptions 4.2 and 4.9, if $n$ is sufficiently large, the estimator $\widehat{\mathbf{w}}$ satisfies

$$\lim_{n\to\infty} \inf_{\boldsymbol{\Theta}\in\mathcal{U}_1(K,\eta,s^*,M^*)} \mathbb{P}_{\boldsymbol{\Theta}}\left[ \|\widehat{\mathbf{w}}(\widehat{\boldsymbol{\Theta}}) - \mathbf{w}\|_1 \le C_3\left( s^2 \frac{\log d}{n} + s\sqrt{\frac{\log d}{n}} \right) \right] = 1,$$

where $\mathbf{w} = -(\boldsymbol{\Theta}\otimes\boldsymbol{\Theta})_{(jk)^c,(jk)}/(\boldsymbol{\Theta}\otimes\boldsymbol{\Theta})_{(jk),(jk)}$, and $C_3$ is a positive constant.

*Proof.* See Appendix C.8 for a detailed proof. □

The following lemma shows that the uniform convergence of the Hessian matrix can be attained under Assumptions 4.2 and 4.9.

**Lemma B.10.** Under Assumptions 4.2 and 4.9, if $n$ is sufficiently large, we have

$$\lim_{n\to\infty} \inf_{\boldsymbol{\Theta}\in\mathcal{U}_1(K,\eta,s^*,M^*)} \mathbb{P}_{\boldsymbol{\Theta}}\left( \|\widehat{\boldsymbol{\Theta}}^{-1} - \boldsymbol{\Theta}^{-1}\|_{\max} \le C_5\sqrt{\frac{\log d}{n}} \right) = 1, \tag{B.16}$$

$$\lim_{n\to\infty} \inf_{\boldsymbol{\Theta}\in\mathcal{U}_1(K,\eta,s^*,M^*)} \mathbb{P}_{\boldsymbol{\Theta}}\left( \big\| -\widehat{\boldsymbol{\Sigma}} + \widehat{\boldsymbol{\Theta}}^{-1} \big\|_{\max} \le C_6\sqrt{\frac{\log d}{n}} \right) = 1. \tag{B.17}$$

where $\mathbf{H} = \boldsymbol{\Theta}^{-1}\otimes\boldsymbol{\Theta}^{-1}$, $C_5$ and $C_6$ are constants.

*Proof.* See Appendix C.9 for a detailed proof. □

**Lemma B.11.** Under Assumptions 4.1, 4.2 and 4.9, we have

$$\lim_{n\to\infty} \inf_{\boldsymbol{\Theta}\in\mathcal{U}_1} \mathbb{P}_{\boldsymbol{\Theta}}\left[ \left| \widehat{S}_n(0, \widehat{\boldsymbol{\Theta}}_{(jk)^c}) - S_n(\boldsymbol{\Theta}) + Kn^{-\eta}\widetilde{\sigma}^2 \right| \le C_3 C_6 s^2 \left(\frac{\log d}{n}\right)^{3/2} + Cs\frac{\log d}{n} \right] = 1, \tag{B.18}$$

where $\mathcal{U}_1 = \mathcal{U}_1(K,\eta,s^*,M^*)$, $\widetilde{\sigma}^2 = H_{(jk)|(jk)^c} = \big[\Theta_{jj}\Theta_{kk} + \Theta_{jk}^2\big]/\big(\Theta_{jj}\Theta_{kk}\big)^2$, and $C_3$ and $C_6$ are the same as the constants in Lemmas B.9 and B.10, $C$ is a constant. If in addition

Assumption 4.5 holds, we have

$$\lim_{n\to\infty} \sup_{\boldsymbol{\Theta}\in\mathcal{U}_1(K,\eta,s^*,M^*)} \sup_{t\in\mathbb{R}} \left| \mathbb{P}_{\boldsymbol{\Theta}}\left(\sqrt{n}\widehat{S}_n(0,\widehat{\boldsymbol{\Theta}}_{(jk)^c})/(2\sigma) \le t\right) - \Phi(t)\right| = 0, \quad \text{if} \quad \eta > 1/2, \tag{B.19}$$

$$\lim_{n\to\infty} \sup_{\boldsymbol{\Theta}\in\mathcal{U}_1(K,\eta,s^*,M^*)} \sup_{t\in\mathbb{R}} \left| \mathbb{P}_{\boldsymbol{\Theta}}\left(\sqrt{n}\widehat{S}_n(0,\widehat{\boldsymbol{\Theta}}_{(jk)^c})/(2\sigma) \le t\right) - \Phi\left(t + K\frac{\widetilde{\sigma}^2}{2\sigma}\right)\right| = 0, \quad \text{if} \quad \eta = 1/2, \tag{B.20}$$

and for any fixed $t \in \mathbb{R}$ and $K \neq 0$, we have

$$\lim_{n\to\infty} \sup_{\boldsymbol{\Theta}\in\mathcal{U}_1(K,\eta,s^*,M^*)} \sup_{t\in\mathbb{R}} \mathbb{P}_{\boldsymbol{\Theta}}\left(\left|\sqrt{n}\widehat{S}_n(0,\widehat{\boldsymbol{\Theta}}_{(jk)^c})/(2\sigma)\right| \le t\right) = 0, \quad \text{if} \quad \eta < 1/2. \tag{B.21}$$

*Proof.* See Appendix C.10 for a detailed proof. □

*Proof of Theorem 4.10.* Define the following events:

$$\mathcal{E}_3 = \left\{ \left\|\widehat{\mathbf{w}}(\widehat{\boldsymbol{\Theta}}) - \mathbf{w}\right\|_1 \le C_3\left(s^2\frac{\log d}{n} + s\sqrt{\frac{\log d}{n}}\right)\right\},$$
$$\mathcal{E}_8 = \left\{ \|\widehat{\mathbf{R}} - \mathbf{R}\|_{\max} \le C_8\sqrt{\frac{\log d}{n}}\right\}.$$

By the similar proof of Theorem B.4, and invoking Lemma B.9, we have

$$\inf_{\boldsymbol{\Theta}\in\mathcal{U}_0(s^*,M^*)} \mathbb{P}_{\boldsymbol{\Theta}}\left[|\widehat{\sigma}^2 - \sigma^2| \le C\left(s^2\left(\frac{\log d}{n}\right)^{3/2} + s\frac{\log d}{n} + \sqrt{\frac{\log d}{n}}\right)\right]$$
$$\ge \inf_{\boldsymbol{\Theta}\in\mathcal{U}_1(K,\eta,s^*,M^*)} \mathbb{P}_{\boldsymbol{\Theta}}(\mathcal{E}_3\cap\mathcal{E}_8) \ge 1 - \sup_{\boldsymbol{\Theta}\in\mathcal{U}_1(K,\eta,s^*,M^*)} \mathbb{P}_{\boldsymbol{\Theta}}(\mathcal{E}_3) - \sup_{\boldsymbol{\Theta}\in\mathcal{U}_1(K,\eta,s^*,M^*)} \mathbb{P}_{\boldsymbol{\Theta}}(\mathcal{E}_8) \to 1.$$

Let $\widehat{U}_n = \mathrm{ST}_n = \sqrt{n}\widehat{S}_n(0,\widehat{\boldsymbol{\Theta}}_{(jk)^c})/(2\widehat{\sigma})$, and $U_n = \sqrt{n}\widehat{S}_n(0,\widehat{\boldsymbol{\Theta}}_{(jk)^c})/(2\sigma)$, For any $t$ and a positive sequence $\delta_n \to 0$, we have

$$\mathbb{P}_{\boldsymbol{\Theta}}(\mathrm{ST}_n \le t) - \Phi\left(t + K\frac{\widetilde{\sigma}^2}{2\sigma}\right) = \underbrace{\mathbb{P}_{\boldsymbol{\Theta}}(\widehat{U}_n \le t) - \mathbb{P}_{\boldsymbol{\Theta}}(U_n \le t + \delta_n)}_{\text{(i)}}$$
$$+ \underbrace{\mathbb{P}_{\boldsymbol{\Theta}}\left(U_n \le t + \delta_n\right) - \Phi\left(t + K\frac{\widetilde{\sigma}^2}{2\sigma} + \delta_n\right)}_{\text{(ii)}} + \underbrace{\Phi\left(t + K\frac{\widetilde{\sigma}^2}{2\sigma} + \delta_n\right) - \Phi\left(t + K\frac{\widetilde{\sigma}^2}{2\sigma}\right)}_{\text{(iii)}}.$$

**Bounding term** (i)**:** By triangle inequality, we have

$$\sup_t\left\{\mathbb{P}_{\boldsymbol{\Theta}}(\widehat{U}_n \le t) - \mathbb{P}_{\boldsymbol{\Theta}}(U_n \le t + \delta_n)\right\} \le \mathbb{P}\left(|\widehat{U}_n - U_n| \ge \delta_n\right) = \mathbb{P}\left(|U_n|\cdot|1 - \sigma/\widehat{\sigma}| \ge \delta_n\right)$$
$$\le \underbrace{\mathbb{P}\left(|U_n| \ge 1/\delta_n\right)}_{\text{(i).a}} + \underbrace{\mathbb{P}\left(|1 - \sigma/\widehat{\sigma}| \ge \delta_n^2\right)}_{\text{(i).b}}.$$

The term (i).a can be further bounded by

$$\mathbb{P}_{\mathbf{\Theta}}\big(|U_n| \geq 1/\delta_n\big) \leq \big|\mathbb{P}_{\mathbf{\Theta}}\big(|U_n| \geq \delta_n^{-1}\big) - \mathbb{P}\bigg(\bigg|Z - K\frac{\widetilde{\sigma}^2}{2\sigma}\bigg| \geq \delta_n^{-1}\bigg)\big| + \mathbb{P}\bigg(\bigg|Z - K\frac{\widetilde{\sigma}^2}{2\sigma}\bigg| \geq \delta_n^{-1}\bigg),$$

where $Z$ is a standard normal random variable. By Lemma B.11, we have

$$\limsup_{n\to\infty} \sup_{\mathbf{\Theta}\in\mathcal{U}_1(K,\eta,s^*,M^*)} \sup_{\delta_n} \bigg|\mathbb{P}_{\mathbf{\Theta}}\big(|U_n| \geq \delta_n^{-1}\big) - \mathbb{P}\bigg(\bigg|Z - K\frac{\widetilde{\sigma}^2}{2\sigma}\bigg| \geq \delta_n^{-1}\bigg)\bigg| = 0.$$

Since the tail bound for the standard normal distribution yields $\mathbb{P}\big(|Z - K\widetilde{\sigma}^2/(2\sigma)| \geq \delta_n^{-1}\big) \leq \mathbb{P}(|Z| \geq \delta_n^{-1} - K\widetilde{\sigma}^2/(2\sigma)) \leq 2/\sqrt{2\pi\big(\delta_n^{-1} - K\widetilde{\sigma}^2/(2\sigma)\big)}\exp\big[-\big(\delta_n^{-1} - K\widetilde{\sigma}^2/(2\sigma)\big)^2/2\big] \to 0$ as $\delta_n \to 0$, we have $\mathbb{P}\big(|U_n| \geq 1/\delta_n\big) \leq 0$. For term (i).b, we have

$$\mathbb{P}_{\mathbf{\Theta}}\big(\big|1 - \widehat{\sigma}/\sigma\big| \geq \delta_n^2\big) = \mathbb{P}_{\mathbf{\Theta}}\bigg(|\widehat{\sigma}^2 - \sigma^2|/\big((\widehat{\sigma} + \sigma)\widehat{\sigma}\big) \geq \delta_n^2\bigg).$$

Let $\eta_n = s^2M^4(\log d/n)^{3/2} + s^2M^6 \log d/n + sM^6\sqrt{\log d/n}$. Given events $\mathcal{E}_3$ and $\mathcal{E}_8$, and the assumption that $\sigma \geq \kappa$, there exists a constant $C$ such that $|\widehat{\sigma}^2 - \sigma^2|/\big((\widehat{\sigma} + \sigma)\widehat{\sigma}\big) \leq C\eta_n$, since $\widehat{\sigma} \geq \kappa - \eta_n \geq \kappa/2$ for sufficiently large $n$. Hence, by setting $\delta_n = C\eta_n^{1/2}$, for some sufficiently large constant $C$, we have $\mathbb{P}_{\mathbf{\Theta}}(|1 - \sigma/\widehat{\sigma}| \geq \delta_n^2) = 0$. Thus, we obtain

$$\limsup_{n\to\infty} \sup_{\mathbf{\Theta}\in\mathcal{U}_1(K,\eta,s^*,M^*)} \sup_{t\in\mathbb{R}} \mathbb{P}_{\mathbf{\Theta}}(\widehat{U}_n \leq t) - \mathbb{P}_{\mathbf{\Theta}}(U_n \leq t + \delta_n) \leq 0. \tag{B.22}$$

**Bounding Term** (ii)**:** By Lemma B.11, we have

$$\limsup_{n\to\infty} \sup_{\mathbf{\Theta}\in\mathcal{U}_1(K,\eta,s^*,M^*)} \sup_{t\in\mathbb{R}} \bigg\{\mathbb{P}_{\mathbf{\Theta}}(U_n \leq t + \delta_n) - \Phi\bigg(t + K\frac{\widetilde{\sigma}^2}{2\sigma} + \delta_n\bigg)\bigg\} \leq 0. \tag{B.23}$$

**Bounding Term** (iii)**:** We have $\Phi\big(t + K\widetilde{\sigma}^2/(2\sigma) + \delta_n\big) - \Phi\big(t + K\widetilde{\sigma}^2/(2\sigma)\big) \leq \delta_n/\sqrt{2\pi}$, which immediately implies that

$$\limsup_{n\to\infty} \sup_{\mathbf{\Theta}\in\mathcal{U}_1(K,\eta,s^*,M^*)} \sup_{t\in\mathbb{R}} \Big\{\Phi\big(t + K\widetilde{\sigma}^2/(2\sigma) + \delta_n\big) - \Phi\big(t + K\widetilde{\sigma}^2/(2\sigma)\big)\Big\} \leq 0. \tag{B.24}$$

Combining (B.22), (B.23) and (B.24), we obtain that

$$\limsup_{n\to\infty} \sup_{\mathbf{\Theta}\in\mathcal{U}_1(K,\eta,s^*,M^*)} \sup_{t\in\mathbb{R}} \Big\{\mathbb{P}_{\mathbf{\Theta}}(\mathrm{ST}_n \leq t) - \Phi(t + K\sigma)\Big\} \leq 0.$$

By similar arguments, we can obtain that

$$\limsup_{n\to\infty} \sup_{\mathbf{\Theta}\in\mathcal{U}_1(K,\eta,s^*,M^*)} \sup_{t\in\mathbb{R}} \Big\{\mathbb{P}_{\mathbf{\Theta}}(\mathrm{ST}_n \leq t) - \Phi(t + K\sigma)\Big\} \geq 0.$$

This completes the proof of (4.6). □

### B.5 Proof of Theorem 4.11

*Proof.* In this proof, we use $\widehat{\mathbf{w}}$ as the shorthand for $\widehat{\mathbf{w}}(\widehat{\boldsymbol{\Theta}})$. By the definition of $\widehat{\Theta}_{jk}^W$, we have

$$\sqrt{n}(\widehat{\Theta}_{jk}^W - \Theta_{jk}^*) = \sqrt{n}(\widehat{\Theta}_{jk} - \Theta_{jk}^*) - \sqrt{n}\big(\nabla_{(jk)}\widehat{S}_n(\widehat{\boldsymbol{\Theta}})\big)^{-1}\widehat{S}_n(\widehat{\boldsymbol{\Theta}}). \tag{B.25}$$

Consider the Taylor expansion of $\widehat{S}_n(\widehat{\boldsymbol{\Theta}})$ at $\Theta_{jk}^*$ as follows

$$\widehat{S}_n(\widehat{\boldsymbol{\Theta}}) = \widehat{S}_n(\Theta_{jk}^*, \widehat{\boldsymbol{\Theta}}_{(jk)^c}) + \nabla_{(jk)}\widehat{S}_n(\bar{\Theta}_{(jk)}, \widehat{\boldsymbol{\Theta}}_{(jk)^c})(\widehat{\Theta}_{jk} - \Theta_{jk}^*), \tag{B.26}$$

where $\bar{\Theta}_{(jk)}$ lies in between $\Theta_{jk}^*$ and $\widehat{\Theta}_{jk}$. Substituting (B.26) into (B.25), we obtain

$$\begin{aligned}
\sqrt{n}(\widehat{\Theta}_{jk}^W - \Theta_{jk}^*) &= \sqrt{n}(\widehat{\Theta}_{jk} - \Theta_{jk}^*) - \sqrt{n}\big(\nabla_{(jk)}\widehat{S}_n(\widehat{\boldsymbol{\Theta}})\big)^{-1}\big[\widehat{S}_n(\Theta_{jk}^*, \widehat{\boldsymbol{\Theta}}_{(jk)^c}) \\
&\quad + \nabla_{(jk)}\widehat{S}_n(\bar{\Theta}_{(jk)}, \widehat{\boldsymbol{\Theta}}_{(jk)^c})(\widehat{\Theta}_{jk} - \Theta_{jk}^*)\big] \\
&= \underbrace{-\sqrt{n}\big(\nabla_{(jk)}\widehat{S}_n(\widehat{\boldsymbol{\Theta}})\big)^{-1}\widehat{S}_n(\Theta_{jk}^*, \boldsymbol{\Theta}_{(jk)^c})}_{\text{(i)}} \\
&\quad + \underbrace{\sqrt{n}(\widehat{\Theta}_{jk} - \Theta_{jk}^*)\Big(1 - \big(\nabla_{(jk)}\widehat{S}_n(\widehat{\boldsymbol{\Theta}})\big)^{-1}\nabla_{(jk)}\widehat{S}_n\big(\bar{\Theta}_{(jk)}, \widehat{\boldsymbol{\Theta}}_{(jk)^c}\big)\Big)}_{\text{(ii)}}.
\end{aligned} \tag{B.27}$$

In the following, we are going to bound terms (i) and (ii) separately.
**Bounding Term** (i)**:** We first show that $\nabla_{(jk)}\widehat{S}_n(\widehat{\boldsymbol{\Theta}}) \xrightarrow{p} H_{(jk)|(jk)^c}$. By triangle inequality, we have

$$|\nabla_{(jk)}\widehat{S}_n(\widehat{\boldsymbol{\Theta}}) - H_{(jk)|(jk)^c}| \le \underbrace{|\nabla^2_{(jk),(jk)}\ell_n(\widehat{\boldsymbol{\Theta}}) - H_{(jk)(jk)}|}_{\text{(iii)}} + \underbrace{|\nabla^2_{(jk),(jk)^c}\ell_n(\widehat{\boldsymbol{\Theta}})\widehat{\mathbf{w}} - \mathbf{H}_{(jk)(jk)^c}\mathbf{w}^*|}_{\text{(iv)}}. \tag{B.28}$$

By Lemma B.2, we have (iii) $\le \|\nabla^2\ell_n(\widehat{\boldsymbol{\Theta}}) - \mathbf{H}\|_{\max} = O_{\mathbb{P}}\big(\sqrt{\log d/n}\big)$. By the triangle inequality, we have

$$\begin{aligned}
\text{(iv)} &\le \underbrace{|\big(\nabla^2_{(jk),(jk)^c}\ell_n(\widehat{\boldsymbol{\Theta}}) - \mathbf{H}_{(jk),(jk)^c}\big)(\widehat{\mathbf{w}} - \mathbf{w}^*)|}_{\text{(iv).a}} + \underbrace{|\mathbf{H}_{(jk),(jk)^c}(\widehat{\mathbf{w}} - \mathbf{w}^*)|}_{\text{(iv).b}} \\
&\quad + \underbrace{|\big(\nabla^2_{(jk),(jk)^c}\ell_n(\widehat{\boldsymbol{\Theta}}) - \mathbf{H}_{(jk),(jk)^c}\big)\mathbf{w}^*|}_{\text{(iv).c}}.
\end{aligned} \tag{B.29}$$

For term (iv).a, we obtain

$$\text{(iv).a} \le \big\|\widehat{\mathbf{w}} - \mathbf{w}^*\big\|_1 \cdot \|\nabla^2_{(jk),(jk)^c}\ell(\widehat{\boldsymbol{\Theta}}) - \mathbf{H}_{(jk),(jk)^c}\|_\infty = \left(s^2\left(\frac{\log d}{n}\right)^{3/2} + s\frac{\log d}{n}\right), \tag{B.30}$$

where the inequality follows from Hölder's inequality, and the equality follows from Lemma B.3

and Lemma B.2. For term (iv).b, we have

$$\text{(iv).b} \le \big\|\widehat{\mathbf{w}} - \mathbf{w}^*\big\|_1 \cdot \|\mathbf{H}_{(jk),(jk)^c}\|_{\max} = O_{\mathbb{P}}\bigg(s^2\frac{\log d}{n} + s\sqrt{\frac{\log d}{n}}\bigg), \tag{B.31}$$

where the inequality follows from Hölder's inequality, and the equality follows from Lemma B.3 and $\|\mathbf{H}\|_{\max} = O_{\mathbb{P}}(1)$. For term (iv).c, we have

$$\text{(iv).c} \le \|\mathbf{w}^*\|_1 \cdot \|\nabla^2_{(jk),(jk)^c}\ell(\widehat{\boldsymbol{\Theta}}) - \mathbf{H}_{(jk),(jk)^c}\|_\infty = O_{\mathbb{P}}\bigg(\sqrt{\frac{\log d}{n}}\bigg), \tag{B.32}$$

where the first inequality follows from Hölder's inequality, and the second inequality follows from Lemma B.2 and $\|\mathbf{w}^*\|_{\max} \le \nu^2 M^2$. Substituting (B.30), (B.31) and (B.32) into (B.29), we obtain

$$\text{(iv)} = \bigg(s^2\Big(\frac{\log d}{n}\Big)^{3/2} + s\frac{\log d}{n} + \sqrt{\frac{\log d}{n}}\bigg). \tag{B.33}$$

Hence, combining terms (iii) and (iv), and submitting it into (B.28), we have

$$|\nabla_{(jk)}\widehat{S}_n(\widehat{\boldsymbol{\Theta}}) - H_{(jk)|(jk)^c}| = \bigg(s^2\Big(\frac{\log d}{n}\Big)^{3/2} + s\frac{\log d}{n} + \sqrt{\frac{\log d}{n}}\bigg) = o_{\mathbb{P}}(1), \tag{B.34}$$

according to Assumption 4.5. Therefore, by (B.34), we have $\nabla_{(jk)}\widehat{S}_n(\widehat{\boldsymbol{\Theta}}) \xrightarrow{p} H_{(jk)|(jk)^c}$. By Theorem 4.7 and Slutsky's theorem, we have

$$\text{(i)} = -\sqrt{n}S(\Theta^*_{jk}, \boldsymbol{\Theta}^*_{(jk)^c})H^{-1}_{(jk)|(jk)^c} + o_{\mathbb{P}}(1) \rightsquigarrow N\Big(0, 4\sigma^2 H^{-2}_{(jk)|(jk)^c}\Big). \tag{B.35}$$

**Bounding Term** (ii)**:** We aim to show that (ii) $= o_{\mathbb{P}}(1)$. In detail, we have

$$\begin{aligned}\text{(ii)} &\le \big|\sqrt{n}(\widehat{\Theta}_{jk} - \Theta^*_{jk})\big| \cdot |1 - \big(\nabla_{(jk)}\widehat{S}_n(\widehat{\boldsymbol{\Theta}})\big)^{-1}\nabla_{(jk)}\widehat{S}_n(\bar{\Theta}_{(jk)}, \widehat{\boldsymbol{\Theta}}_{(jk)^c})| \\ &\le \sqrt{n}\,\underbrace{\big|\widehat{\Theta}_{jk} - \Theta^*_{jk}\big|}_{\text{(ii).a}} \cdot \underbrace{\big|\big(\nabla_{(jk)}\widehat{S}_n(\widehat{\boldsymbol{\Theta}})\big)^{-1}\big|}_{\text{(ii).b}} \cdot \underbrace{\big|\nabla_{(jk)}\widehat{S}_n(\widehat{\boldsymbol{\Theta}}) - \nabla_{(jk)}\widehat{S}_n(\bar{\Theta}_{(jk)}, \widehat{\boldsymbol{\Theta}}_{(jk)^c})\big|}_{\text{(ii).c}}.\end{aligned} \tag{B.36}$$

For term (ii).a, we have (ii).a $\le \|\widehat{\boldsymbol{\Theta}} - \boldsymbol{\Theta}^*\|_{\max} = O_{\mathbb{P}}\big(\sqrt{\log d/n}\big)$. For term (ii).b, we have shown in (B.34) that $\nabla_{(jk)}\widehat{S}_n(\widehat{\boldsymbol{\Theta}}) \xrightarrow{p} H_{(jk)|(jk)^c}$. Therefore, we have (ii).b $\xrightarrow{p} H^{-1}_{(jk)|(jk)^c} = O(1)$. For term (ii).c, by triangle inequality, we have

$$\text{(ii).c} \le \underbrace{\big|\nabla_{(jk)}\widehat{S}_n(\widehat{\boldsymbol{\Theta}}) - H_{(jk)|(jk)^c}\big|}_{I_1} + \underbrace{\big|H_{(jk)|(jk)^c} - \nabla_{(jk)}\widehat{S}_n(\bar{\Theta}_{(jk)}, \widehat{\boldsymbol{\Theta}}_{(jk)^c})\big|}_{I_2}. \tag{B.37}$$

By similar argument as in (B.28), we can show that

$$
\begin{aligned}
I_1 &= O_{\mathbb{P}}\bigg(s^2\bigg(\frac{\log d}{n}\bigg)^{3/2} + s\frac{\log d}{n} + \sqrt{\frac{\log d}{n}}\bigg),\\
I_2 &= O_{\mathbb{P}}\bigg(s^2\bigg(\frac{\log d}{n}\bigg)^{3/2} + s\frac{\log d}{n} + \sqrt{\frac{\log d}{n}}\bigg).
\end{aligned} \tag{B.38}
$$

Submitting (B.38) into (B.37), we can get

$$
\text{(ii).c} = O_{\mathbb{P}}\bigg(s^2\bigg(\frac{\log d}{n}\bigg)^{3/2} + s\frac{\log d}{n} + \sqrt{\frac{\log d}{n}}\bigg). \tag{B.39}
$$

Combining terms (i).a, (i).b and (i).c and substituting it back into (B.36), and invoking Assumption 4.5, we obtain

$$
\text{(ii)} = \sqrt{n}O_{\mathbb{P}}\bigg(s^2\bigg(\frac{\log d}{n}\bigg)^2 + s\bigg(\frac{\log d}{n}\bigg)^{3/2} + \frac{\log d}{n}\bigg) = o_{\mathbb{P}}(1). \tag{B.40}
$$

Substituting (B.35) and (B.40) into (B.27), by Slustky's theorem, we have

$$
\sqrt{n}(\widehat{\Theta}^W_{jk} - \Theta^*_{jk}) = -\sqrt{n}S_n(\boldsymbol{\Theta}^*)H^{-1}_{(jk)|(jk)^c} + o_{\mathbb{P}}(1) \rightsquigarrow N(0, 4\sigma^2 H^{-2}_{(jk)|(jk)^c}).
$$

This completes the proof. □

## B.6 Proof of Corollary 4.12

*Proof.* Note that by Theorem B.4, we have $|\widehat{\sigma}^2 - \sigma^2| = o_{\mathbb{P}}(1)$. It is sufficient to show that $\widehat{H}_{(jk)|(jk)^c} - H_{(jk)|(jk)^c} = o_{\mathbb{P}}(1)$. Note that

$$
\big|\widehat{H}_{(jk)|(jk)^c} - H_{(jk)|(jk)^c}\big| \le \underbrace{\big|\nabla^2_{(jk),(jk)}\ell_n(\widehat{\boldsymbol{\Theta}}) - H_{(jk)(jk)}\big|}_{\text{(i)}} + \underbrace{\big|\nabla^2_{(jk),(jk)^c}\ell_n(\widehat{\boldsymbol{\Theta}})\widehat{\mathbf{w}} - \mathbf{H}_{(jk),(jk)^c}\mathbf{w}^*\big|}_{\text{(ii)}}. \tag{B.41}
$$

For term (i), according to Lemma B.2, we have

$$
\text{(i)} \le \big\|\nabla^2\ell_n(\widehat{\boldsymbol{\Theta}}) - \mathbf{H}\big\|_{\max} = O_{\mathbb{P}}\bigg(\sqrt{\frac{\log d}{n}}\bigg) = o_{\mathbb{P}}(1). \tag{B.42}
$$

For term (ii), by similar argument as in (B.29), we can show that

$$
\text{(ii)} = \bigg(s^2\bigg(\frac{\log d}{n}\bigg)^{3/2} + s\frac{\log d}{n} + \sqrt{\frac{\log d}{n}}\bigg) = o_{\mathbb{P}}(1), \tag{B.43}
$$

according to Assumption 4.5. Combining terms (i) and (ii) in (B.42) and (B.43) and submitting it into (B.41), we obtain

$$\left|\widehat{H}_{(jk)|(jk)^c} - H_{(jk)|(jk)^c}\right| = o_{\mathbb{P}}(1).$$

Applying Slustky's theorem completes the proof. □

### B.7 Proof of Corollary 4.13

*Proof.* By Theorem 4.11 and Corollary 4.12, we have

$$\mathbb{P}(\Psi_W(\alpha) = 1|H_0) = \mathbb{P}(W_n^2 \geq \chi^2_{1\alpha}) = \alpha.$$

Similarly, for any $t \in (0,1)$, we have

$$\mathbb{P}(P_W < t) = \mathbb{P}\Big(\Phi(W_n) > 1 - \frac{t}{2}\Big) = \mathbb{P}\Big(W_n > \Phi^{-1}\Big(1 - \frac{t}{2}\Big)\Big) \to t.$$

This completes the proof. □

### B.8 Proof of Theorem 4.14

In order to analyze the theoretical property of confidence subgraph, we introduce intermediate test statistics $T_0$, $T_1$ and $T_2$ as follows,

$$T_0 = \max_{(j,k)\in[d]\times[d]} \frac{1}{\sqrt{n}} \sum_{i=1}^{n} Z_{ijk}, \tag{B.44}$$

$$T_1 = \max_{(j,k)\in[d]\times[d]} \sqrt{n}\Big(\widehat{\Theta}_{jk} - \Theta^*_{jk}\Big) \cdot H_{(jk)|(jk)^c}/(2\sigma), \tag{B.45}$$

$$T_2 = \max_{(j,k)\in[d]\times[d]} \sqrt{n} S_n(\mathbf{\Theta}^*)/(2\sigma), \tag{B.46}$$

where $Z_{ijk} = -1/(2\sigma) \cdot \mathbf{b}^\top \text{vec}(\mathbf{T} \odot \overline{\mathbf{G}}^i)$. Remind that $S_n(\mathbf{\Theta}^*) = \mathbf{b}^\top \text{vec}\big(\mathbf{T} \odot \mathbf{G}\big)$. Using $T_1$ and $T_2$ as intermediate test statistics, we can establish the connection between $T_0$ and $T$. Furthermore, we introduce another test statistic $W_0$ as follows

$$W_0 = \max_{(j,k)\in[d]\times[d]} \frac{1}{\sqrt{n}} \sum_{i=1}^{n} Z_{ijk} \cdot e_i, \tag{B.47}$$

where $\{e_i\}_{i=1}^n$ is a sequence of i.i.d. standard normal random variables that are independent of $\{\boldsymbol{X}_i\}_{i=1}^n$. Note that both $T_0$ and $W_0$ have the structure of summation of independent random variables. In what follows, we are going to build connections among the quantiles of $T_0$, $T$, $W_0$, and $W$. We define the multiplier bootstrap estimator of the $\alpha$-quantile of $T_0$ as the conditional $\alpha$-quantile of $W_0$ conditioned on $\{\boldsymbol{X}_i\}_{i=1}^n$, that is,

$$c_{W_0}(\alpha) = \inf\big\{t \in \mathbb{R} : \mathbb{P}_e(W_0 \le t) \ge \alpha\big\},$$

where $\mathbb{P}_e$ is the probability measure induced by the multiplier variables $\{e_i\}_{i=1}^n$, given fixed $\{\boldsymbol{X}_i\}_{i=1}^n$, i.e., $\mathbb{P}_e(W_0 \le t) = \mathbb{P}(W_0 \le t | \{\boldsymbol{X}_i\}_{i=1}^n)$.

In order to prove Theorem 4.14, we need a series of auxiliary lemmas. We first give a lemma, which is proved in Chernozhukov et al. (2012).

**Lemma B.12.** (Chernozhukov et al., 2012) Suppose that there exist constants $c_1, c_2, C_1, C_2 > 0$ and a sequence $B_n \ge 1$, $\lim_{n\to\infty} B_n \to \infty$ such that

$$c_1 \le \frac{1}{n}\sum_{i=1}^n \mathbb{E}[Z_{ijk}^2] \le C_1, \tag{B.48}$$

$$\max_{r=1,2} \frac{1}{n}\sum_{i=1}^n \mathbb{E}[|Z_{ijk}|^{2+r}] + \mathbb{E}[\exp(|Z_{ijk}|/B_n)] \le 4, \tag{B.49}$$

and $B_n^2(\log(dn))^7/n \le C_2 n^{-c_2}$ for $1 \le k \le n$ and $(j,k) \in E^c$. Then there exists constants $c > 0$ and $C > 0$ depending only on $c_1, C_1, c_2$ and $C_2$ such that

$$\sup_{\alpha\in(0,1)} |P\big(T_0 \le c_{U_0}(\alpha)\big) - \alpha| \le Cn^{-c}.$$

Lemma B.12 shows that $c_{U_0}(\alpha)$ is an estimator of the $\alpha$-quantile of $T$. Note that in our model $\mathbf{w}^*$ is sparse and $Z_{ijk}$ is sub-exponential for all $i, j, k$. Setting $B_n = C_1$, we get

$$\frac{1}{n}\sum_{i=1}^n \mathbb{E}[Z_{ijk}^2] \ge c_1 \text{ and } \mathbb{E}[\exp(|Z_{ijk}|/C_1)] \le 2.$$

This verifies the conditions (B.48) and (B.49).

**Lemma B.13.** (Chernozhukov et al., 2012) Let $V^{(1)}$ and $V^{(2)}$ be contered Gaussian random vectors in $\mathbb{R}^p$ with covariance matrices $\Sigma^{(1)}$ and $\Sigma^{(2)}$ respectively. Suppose that there are some constants $0 < c_1 < C_1$ such that $c_1 \le \Sigma_{jj}^{(2)} \le C_1$ for all $j = 1, \ldots, p$. Then there exists a constant $C > 0$ depending only on $c_1$ and $C_1$ such that

$$\sup_{t\in\mathbb{R}} \Big|\mathbb{P}(\max_{1\le j\le p} V_j^{(1)} \le t) - \mathbb{P}(\max_{1\le j\le p} V_j^{(2)} \le t)\Big| \le C\Delta_0^{1/3}(1 \vee \log(p/\Delta_0))^{2/3},$$

where $\Delta_0 = \max_{j,k} |\Sigma_{jk}^{(1)} - \Sigma_{jk}^{(2)}|$.

Let $U_0 = \max_{(j,k)\in[d]\times[d]} 1/\sqrt{n}\sum_{i=1}^n y_{ijk}$, where $\{y_{ijk}\}_{(j,k)\in[d]\times[d],i=1,\ldots,n}$ are Gaussian analogs of $\{Z_{ijk}\}_{(j,k)\in[d]\times[d]}, i = 1, \ldots, n$. Then we have the following result.

**Lemma B.14.** Suppose that there exist some constants $0 < c_1 < C_1$ such that $c_1 \leq \frac{1}{n}\sum_{k=1}^{n}\mathbb{E}[Z_{ijk}^2] \leq C_1$ for all $(j,k) \in [d] \times [d]$. Then for every $\alpha \in (0,1)$,

$$\mathbb{P}(c_{W_0}(\alpha) \leq c_{U_0}(\alpha + \pi(\vartheta))) \geq 1 - \mathbb{P}(\Delta \geq \vartheta),$$
$$\mathbb{P}(c_{U_0}(\alpha) \leq c_{W_0}(\alpha + \pi(\vartheta))) \geq 1 - \mathbb{P}(\Delta \geq \vartheta),$$

where $\Delta = \max_{(j,k),(j',k')\in[d]\times[d]}\left|\frac{1}{n}\sum_{i=1}^{n}\big(Z_{ijk}Z_{ij'k'} - \mathbb{E}[Z_{ijk}Z_{ij'k'}]\big)\right|$, $\pi(\vartheta) = C_2\vartheta^{1/3}(1 \vee \log(|E^c|/\vartheta))^{2/3}$ and $C_2$ is a constant that only depends on $c_1$ and $C_1$.

*Proof.* Now on the event $\{\Delta \leq \vartheta\}$, we have $|\mathbb{P}(U_0 \leq t) - \mathbb{P}_e(W_0 \leq t)| \leq \pi(\vartheta)$ for all $t \in \mathbb{R}$, so

$$\mathbb{P}\big(W_0 \leq c_{U_0}(\alpha + \pi(\vartheta))\big) \geq \mathbb{P}(U_0 \leq c_{U_0}(\alpha + \pi(\vartheta))) - \pi(\vartheta) \geq \alpha,$$

implying the first claim. The second claim follows similarly. □

We need the following auxiliary lemma, which shows the basic approximation properties for multiplier bootstrap method.

**Lemma B.15.** For $T, T_0, W, W_0$ defined in (B.44), (3.15), and (B.47), under the same assumptions of Theorem 4.11, we have

$$|T - T_0| \xrightarrow{p} 0, \ |W - W_0| \xrightarrow{p} 0.$$

*Proof.* See Appendix C.11 for a detailed proof. □

**Lemma B.16.** Under the same assumptions of Theorem 4.11, there exist $\zeta_1 \geq 0$ and $\zeta_2 \geq 0$, depending on $n$ and typically $\zeta_1 \to 0, \zeta_2 \to 0$ as $n \to \infty$ such that

$$\mathbb{P}\big(|T - T_0| > \zeta_1\big) \leq \zeta_2, \tag{B.50}$$
$$\mathbb{P}\big(\mathbb{P}_e(W - W_0) > \zeta_1) > \zeta_2\big) < \zeta_2. \tag{B.51}$$

*Proof.* See Appendix C.12 for a detailed proof. □

**Lemma B.17.** Suppose that condition (B.51) holds. Then for every $\alpha \in (0,1)$,

$$\mathbb{P}\big(c_{W_0} \leq c_{W_0}(\alpha + \zeta_2) + \zeta_1\big) \geq 1 - \zeta_2,$$
$$\mathbb{P}\big(c_{W_0} \leq c_W(\alpha + \zeta_2) + \zeta_1\big) \geq 1 - \zeta_2.$$

*Proof.* By (B.51), the probability of the event $\mathbb{P}_e(W - W_0) > \zeta_1) \leq \zeta_2$ is at least $1 - \zeta_2$. On this event,

$$\mathbb{P}_e\big(W_0 \leq c_{W_0}(\alpha + \zeta_2) + \zeta_1\big) \geq \mathbb{P}_e\big(U_0 \leq c_{U_0}(\alpha + \pi(\vartheta))\big) - \pi(\vartheta) \geq \alpha,$$

which implies the first claim. The second claim follows similarly. □

*Proof of Theorem 4.14.* Recall that

$$\pi(\vartheta) = C_2 \vartheta^{1/3}(1 \vee \log(|E^c|/\vartheta))^{2/3}.$$

In addition, let $\kappa_1(\vartheta) = c_{Z_0}(\alpha - \zeta_2 - \pi(\vartheta))$ and $\kappa_2(\vartheta) = c_{Z_0}(\alpha + \zeta_2 + \pi(\vartheta))$. Then by Lemmas B.12, B.14 and B.17, we have

$$\begin{aligned}
&\mathbb{P}\big(\{T \le c_W(\alpha), T_0 > c_{U_0}(\alpha)\} \cup \{T > c_W(\alpha), T_0 \le c_{U_0}(\alpha)\}\big) \\
&\le \mathbb{P}\big(\kappa_1(\vartheta) - 2\zeta_1 < T_0 \le \kappa_2(\vartheta) + 2\zeta_1\big) + 2\mathbb{P}(\Delta > \vartheta) + 3\zeta_2 \\
&\le \mathbb{P}\big(\kappa_1(\vartheta) - 2\zeta_1 < U_0 \le \kappa_2(\vartheta) + 2\zeta_1\big) + 2\mathbb{P}(\Delta > \vartheta) + 3\zeta_2 + 2Cn^{-c} \\
&\le 2\pi(\vartheta) + 2\mathbb{P}(\Delta > \vartheta) + 2Cn^{-c} + C_3\zeta_1\sqrt{1 \vee \log(|E^c|/\zeta_1)} + 5\zeta_2.
\end{aligned}$$

□

# C Proof of Auxiliary Lemmas in Section B

## C.1 Proof of Lemma B.1

**Lemma C.1.** Suppose $\boldsymbol{\Theta}^* \in \mathcal{U}(s, K)$. Under Assumptions 4.1 and 4.2, we have

$$\frac{\sqrt{n}}{2}\big(\mathbf{b}^\top \mathbf{R} \mathbf{b}\big)^{-\frac{1}{2}} \mathbf{b}^\top \mathrm{vec}\big(\nabla \ell_n(\boldsymbol{\Theta}^*)\big) \rightsquigarrow N(0,1).$$

where $\mathbf{b} = [1, -(\mathbf{w}^*)^\top]^\top$.

*Proof of Lemma B.1.* By Lemma C.1, replacing $\boldsymbol{\Theta}^*$ with $(0, \boldsymbol{\Theta}^*_{(jk)^c})$ completes the proof. □

## C.2 Proof of Lemma B.2

*Proof.* By definition, we have

$$\begin{aligned}
&\|\nabla^2 \ell_n(\widehat{\boldsymbol{\Theta}}) - \mathbf{H}\|_{\max} = \|\widehat{\boldsymbol{\Theta}}^{-1} \otimes \widehat{\boldsymbol{\Theta}}^{-1} - \boldsymbol{\Theta}^{*-1} \otimes \boldsymbol{\Theta}^{*-1}\|_{\max} \\
&\le \|\widehat{\boldsymbol{\Theta}}^{-1} \otimes (\widehat{\boldsymbol{\Theta}}^{-1} - \boldsymbol{\Theta}^{*-1})\|_{\max} + \|(\widehat{\boldsymbol{\Theta}}^{-1} - \boldsymbol{\Theta}^{*-1}) \otimes \boldsymbol{\Theta}^{*-1}\|_{\max} \\
&\le \underbrace{\|(\widehat{\boldsymbol{\Theta}}^{-1} - \boldsymbol{\Theta}^{*-1}) \otimes (\widehat{\boldsymbol{\Theta}}^{-1} - \boldsymbol{\Theta}^{*-1})\|_{\max}}_{\text{(i)}} + \underbrace{2\|(\widehat{\boldsymbol{\Theta}}^{-1} - \boldsymbol{\Theta}^{*-1}) \otimes \boldsymbol{\Theta}^{*-1}\|_{\max}}_{\text{(ii)}}.
\end{aligned} \tag{C.1}$$

For term (i), we have (i) $\le \big\|\widehat{\boldsymbol{\Theta}}^{-1} - \boldsymbol{\Theta}^{*-1}\big\|_{\max}^2$. For term (ii), we have

$$\text{(ii)} \le 2\|(\widehat{\boldsymbol{\Theta}}^{-1} - \boldsymbol{\Theta}^{*-1})\|_{\max} \cdot \|\boldsymbol{\Sigma}^*\|_2 \le 2v\|(\widehat{\boldsymbol{\Theta}}^{-1} - \boldsymbol{\Theta}^{*-1})\|_{\max},$$

where the second inequality follows from Assumption 4.2. Substituting terms (i) and (ii) into (C.1), we obtain

$$\big\|\nabla^2 \ell_n(\widehat{\boldsymbol{\Theta}}) - \mathbf{H}\big\|_{\max} \le \big\|(\widehat{\boldsymbol{\Theta}}^{-1} - \boldsymbol{\Theta}^{*-1})\big\|_{\max}^2 + 2v\big\|(\widehat{\boldsymbol{\Theta}}^{-1} - \boldsymbol{\Theta}^{*-1})\big\|_{\max}. \tag{C.2}$$

From (C.2), we can see that in order to bound $\big\|\nabla^2\ell_n(\widehat{\boldsymbol{\Theta}})-\mathbf{H}\big\|_{\max}$, it is sufficient to get a bound on $\big\|\widehat{\boldsymbol{\Theta}}^{-1}-\boldsymbol{\Theta}^{*-1}\big\|_{\max}$. Thus, in the following, we aim at bounding $\|\widehat{\boldsymbol{\Theta}}^{-1}-\boldsymbol{\Theta}^{*-1}\|_{\max}$. Consider the Taylor expansion of $\widehat{\boldsymbol{\Theta}}^{-1}$ at $\boldsymbol{\Theta}^*$ as follows:

$$\big(\boldsymbol{\Theta}^*+\widehat{\boldsymbol{\Delta}}\big)^{-1}=\big(\mathbf{I}+(\boldsymbol{\Theta}^*)^{-1}\widehat{\boldsymbol{\Delta}}\big)^{-1}\boldsymbol{\Theta}^{*-1}=\sum_{k=0}^{\infty}(-1)^k\big(\boldsymbol{\Theta}^{*-1}\widehat{\boldsymbol{\Delta}}\big)^k\boldsymbol{\Theta}^{*-1}. \tag{C.3}$$

The right hand side of (C.3) can be further rewritten as

$$\sum_{k=0}^{\infty}(-1)^k\big(\boldsymbol{\Theta}^{*-1}\widehat{\boldsymbol{\Delta}}\big)^k\boldsymbol{\Theta}^{*-1}=\boldsymbol{\Theta}^{*-1}-\boldsymbol{\Theta}^{*-1}\widehat{\boldsymbol{\Delta}}\sum_{k=0}^{\infty}(-1)^k\big(\boldsymbol{\Theta}^{*-1}\widehat{\boldsymbol{\Delta}}\big)^k\boldsymbol{\Theta}^{*-1}=\boldsymbol{\Theta}^{*-1}-\mathbf{R}(\widehat{\boldsymbol{\Delta}}), \tag{C.4}$$

where $\mathbf{R}(\widehat{\boldsymbol{\Delta}})=\boldsymbol{\Theta}^{*-1}\widehat{\boldsymbol{\Delta}}\mathbf{J}\boldsymbol{\Theta}^{*-1}$ and $\mathbf{J}=\sum_{k=0}^{\infty}(-1)^k\big(\boldsymbol{\Theta}^{*-1}\widehat{\boldsymbol{\Delta}}\big)^k$. Thus, combining (C.3) and (C.4), we obtain

$$\widehat{\boldsymbol{\Theta}}^{-1}-\boldsymbol{\Theta}^{*-1}=\mathbf{R}(\widehat{\boldsymbol{\Delta}}). \tag{C.5}$$

From (C.5), in order to bound $\big\|\widehat{\boldsymbol{\Theta}}^{-1}-\boldsymbol{\Theta}^{*-1}\big\|_{\max}$, it is equivalent to bound $\big\|\mathbf{R}(\widehat{\boldsymbol{\Delta}})\big\|_{\max}$. We have

$$\begin{aligned}\big\|\mathbf{R}(\widehat{\boldsymbol{\Delta}})\big\|_{\max}&=\max_{i,j}\mathbf{e}_i^\top\boldsymbol{\Theta}^{*-1}\widehat{\boldsymbol{\Delta}}\mathbf{J}\boldsymbol{\Theta}^{*-1}\mathbf{e}_j\le\big\|\boldsymbol{\Theta}^{*-1}\big\|_\infty\cdot\big\|\widehat{\boldsymbol{\Delta}}\mathbf{J}\big\|_{\max}\cdot\big\|\boldsymbol{\Theta}^{*-1}\big\|_\infty\\&\le\big\|\boldsymbol{\Theta}^{*-1}\big\|_\infty\cdot\big\|\widehat{\boldsymbol{\Delta}}\big\|_{\max}\cdot\big\|\mathbf{J}\big\|_\infty\cdot\big\|\boldsymbol{\Theta}^{*-1}\big\|_\infty\le K_{\boldsymbol{\Sigma}^*}^2\big\|\widehat{\boldsymbol{\Delta}}\big\|_{\max}\cdot\big\|\mathbf{J}\big\|_\infty,\end{aligned} \tag{C.6}$$

where the second inequality follows from $\|\mathbf{AB}\|_{\max}\le\|\mathbf{A}\|_{\max}\cdot\|\mathbf{B}\|_\infty$, and the third inequality follows from Assumption 4.2. Recall that $\mathbf{J}=\sum_{k=0}^{\infty}(-1)^k(\boldsymbol{\Theta}^{*-1}\widehat{\boldsymbol{\Delta}})^k$, we have

$$\|\mathbf{J}\|_\infty\le\sum_{k=0}^{\infty}\big\|(\boldsymbol{\Theta}^{*-1}\widehat{\boldsymbol{\Delta}})^k\big\|_\infty\le\sum_{k=0}^{\infty}\big\|\boldsymbol{\Theta}^{*-1}\widehat{\boldsymbol{\Delta}}\big\|_\infty^k\le\frac{1}{1-\big\|\boldsymbol{\Theta}^{*-1}\big\|_\infty\cdot\big\|\widehat{\boldsymbol{\Delta}}\big\|_\infty}\le\frac{1}{1-K_{\boldsymbol{\Sigma}^*}\big\|\widehat{\boldsymbol{\Delta}}\big\|_\infty},$$

where the first inequality follows from triangle inequality, the second inequality follows from submultiplicativity of matrix $\|\cdot\|_\infty$ norm, and the last inequality follows from Assumption 4.2. Since $\|\widehat{\boldsymbol{\Delta}}\|_\infty=\|\widehat{\boldsymbol{\Delta}}\|_1=O(s\sqrt{\log d/n})$, when $n$ is sufficiently large, we have $K_{\boldsymbol{\Sigma}^*}\|\widehat{\boldsymbol{\Delta}}\|_\infty\le 1/2$, and therefore

$$\|\mathbf{J}\|_\infty\le\frac{1}{1-1/2}\le 2. \tag{C.7}$$

Substituting (C.7) into (C.6) and (C.5), we obtain

$$\|\widehat{\boldsymbol{\Theta}}^{-1}-\boldsymbol{\Theta}^{*-1}\|_{\max}=\|R(\widehat{\boldsymbol{\Delta}})\|_{\max}\le 2K_{\boldsymbol{\Sigma}^*}^2\|\widehat{\boldsymbol{\Delta}}\|_{\max}=O_{\mathbb{P}}\bigg(\sqrt{\frac{\log d}{n}}\bigg). \tag{C.8}$$

where the last equality follows $\nu = O(1)$. Finally, substituting (C.8) into (C.2), we get

$$\|\nabla^2 \ell_n(\widehat{\mathbf{\Theta}}) - \mathbf{H}\|_{\max} = O_{\mathbb{P}}\bigg(\sqrt{\frac{\log d}{n}}\bigg).$$

This completes the proof. ☐

### C.3 Proof of Lemma B.3

*Proof.* In this proof, we use $\widehat{\mathbf{w}}$ as the shorthand for $\widehat{\mathbf{w}}(\widehat{\mathbf{\Theta}})$. By the definition of $\widehat{\mathbf{w}}$ and $\mathbf{w}^*$ in (3.7), we have

$$\begin{aligned}
\|\widehat{\mathbf{w}} - \mathbf{w}^*\|_1 &= \bigg\| -\frac{\widehat{\mathbf{\Omega}}_{(jk)^c,(jk)}}{\widehat{\mathbf{\Omega}}_{(jk),(jk)}} + \frac{\mathbf{\Omega}_{(jk)^c,(jk)}}{\mathbf{\Omega}_{(jk),(jk)}} \bigg\|_1 = \bigg\| \frac{(\widehat{\mathbf{\Theta}} \otimes \widehat{\mathbf{\Theta}})_{(jk)^c,(jk)}}{(\widehat{\mathbf{\Theta}} \otimes \widehat{\mathbf{\Theta}})_{(jk),(jk)}} - \frac{(\mathbf{\Theta}^* \otimes \mathbf{\Theta}^*)_{(jk)^c,(jk)}}{(\mathbf{\Theta}^* \otimes \mathbf{\Theta}^*)_{(jk),(jk)}} \bigg\|_1 \\
&\le \underbrace{\bigg\| \frac{(\widehat{\mathbf{\Theta}} \otimes \widehat{\mathbf{\Theta}})_{(jk)^c,(jk)}}{(\widehat{\mathbf{\Theta}} \otimes \widehat{\mathbf{\Theta}})_{(jk),(jk)}} - \frac{(\mathbf{\Theta}^* \otimes \mathbf{\Theta}^*)_{(jk)^c,(jk)}}{(\widehat{\mathbf{\Theta}} \otimes \widehat{\mathbf{\Theta}})_{(jk),(jk)}} \bigg\|_1}_{\text{(i)}} \\
&\quad + \underbrace{\bigg\| \frac{(\mathbf{\Theta}^* \otimes \mathbf{\Theta}^*)_{(jk)^c,(jk)}}{(\widehat{\mathbf{\Theta}} \otimes \widehat{\mathbf{\Theta}})_{(jk),(jk)}} - \frac{(\mathbf{\Theta}^* \otimes \mathbf{\Theta}^*)_{(jk)^c,(jk)}}{(\mathbf{\Theta}^* \otimes \mathbf{\Theta}^*)_{(jk),(jk)}} \bigg\|_1}_{\text{(ii)}}.
\end{aligned} \tag{C.9}$$

In the sequel, we bound terms (i) and (ii) respectively.
**Bounding term (i):** We have

$$\text{(i)} = \underbrace{\frac{1}{(\widehat{\mathbf{\Theta}} \otimes \widehat{\mathbf{\Theta}})_{(jk),(jk)}}}_{\text{(i).a}} \cdot \underbrace{\big\|(\widehat{\mathbf{\Theta}} \otimes \widehat{\mathbf{\Theta}})_{(jk)^c,(jk)} - (\mathbf{\Theta}^* \otimes \mathbf{\Theta}^*)_{(jk)^c,(jk)}\big\|_1}_{\text{(i).b}}. \tag{C.10}$$

To bound term (i).a, we have

$$\begin{aligned}
(\widehat{\mathbf{\Theta}} \otimes \widehat{\mathbf{\Theta}})_{(jk),(jk)} &\ge (\mathbf{\Theta}^* \otimes \mathbf{\Theta}^*)_{(jk),(jk)} - \big|(\widehat{\mathbf{\Theta}} \otimes \widehat{\mathbf{\Theta}})_{(jk),(jk)} - (\mathbf{\Theta}^* \otimes \mathbf{\Theta}^*)_{(jk),(jk)}\big| \\
&\ge \frac{1}{\nu^2} - \big|(\widehat{\mathbf{\Theta}} \otimes \widehat{\mathbf{\Theta}})_{(jk),(jk)} - (\mathbf{\Theta}^* \otimes \mathbf{\Theta}^*)_{(jk),(jk)}\big|,
\end{aligned} \tag{C.11}$$

where the second inequality uses the fact that $(\mathbf{\Theta}^* \otimes \mathbf{\Theta}^*)_{(jk),(jk)} \ge \lambda_{\min}(\mathbf{\Theta}^* \otimes \mathbf{\Theta}^*) \ge \lambda_{\min}(\mathbf{\Theta}^*)^2 \ge \frac{1}{\nu^2}$ under Assumption 4.2. It remains to a upper bound $\big|(\widehat{\mathbf{\Theta}} \otimes \widehat{\mathbf{\Theta}})_{(jk),(jk)} - (\mathbf{\Theta}^* \otimes \mathbf{\Theta}^*)_{(jk),(jk)}\big|$ in (C.11). We have

$$\begin{aligned}
\big|(\widehat{\mathbf{\Theta}} \otimes \widehat{\mathbf{\Theta}})_{(jk),(jk)} - (\mathbf{\Theta}^* \otimes \mathbf{\Theta}^*)_{(jk),(jk)}\big| &= \big|\widehat{\mathbf{\Theta}}_{jj}\widehat{\mathbf{\Theta}}_{kk} - \mathbf{\Theta}^*_{jj}\mathbf{\Theta}^*_{kk}\big| \\
&\le \big|\widehat{\mathbf{\Theta}}_{jj}\widehat{\mathbf{\Theta}}_{kk} - \mathbf{\Theta}^*_{jj}\widehat{\mathbf{\Theta}}_{kk}\big| + \big|\mathbf{\Theta}^*_{jj}\widehat{\mathbf{\Theta}}_{kk} - \mathbf{\Theta}^*_{jj}\mathbf{\Theta}^*_{kk}\big|.
\end{aligned} \tag{C.12}$$

The right hand side of (C.12) can be bounded by

$$
\begin{aligned}
\big|\widehat{\boldsymbol{\Theta}}_{jj}\widehat{\boldsymbol{\Theta}}_{kk} - \boldsymbol{\Theta}^*_{jj}\widehat{\boldsymbol{\Theta}}_{kk}\big| + \big|\boldsymbol{\Theta}^*_{jj}\widehat{\boldsymbol{\Theta}}_{kk} - \boldsymbol{\Theta}^*_{jj}\boldsymbol{\Theta}^*_{kk}\big| &\le \big|\widehat{\boldsymbol{\Theta}}_{jj} - \boldsymbol{\Theta}^*_{jj}\big| \cdot \big|\widehat{\Theta}_{kk}\big| + \big|\Theta^*_{jj}\big| \cdot \big|\widehat{\boldsymbol{\Theta}}_{kk} - \boldsymbol{\Theta}^*_{kk}\big| \\
&\le \|\widehat{\boldsymbol{\Theta}} - \boldsymbol{\Theta}^*\|_2\big(\|\widehat{\boldsymbol{\Theta}} - \boldsymbol{\Theta}^*\|_2 + \nu\big) + \nu\|\widehat{\boldsymbol{\Theta}} - \boldsymbol{\Theta}^*\|_2 \\
&= O_{\mathbb{P}}\bigg(s\sqrt{\frac{\log d}{n}}\bigg), \qquad \text{(C.13)}
\end{aligned}
$$

where the second inequality uses $\big|\Theta^*_{jj}\big| \le \lambda_{\max}(\boldsymbol{\Theta}^*) \le \nu$ and $\big|\widehat{\Theta}_{kk}\big| \le \big|\widehat{\Theta}_{kk} - \Theta^*_{kk}\big| + \big|\Theta^*_{kk}\big| \le \big\|\widehat{\boldsymbol{\Theta}} - \boldsymbol{\Theta}^*\big\|_2 + \nu$, and the equality follows from $\big\|\widehat{\boldsymbol{\Theta}} - \boldsymbol{\Theta}^*\big\|_2 = O\big(s\sqrt{\log d/n}\big)$ and $\nu = O(1)$. Substituting (C.13) into (C.12), we obtain

$$
\big|\big(\widehat{\boldsymbol{\Theta}} \otimes \widehat{\boldsymbol{\Theta}}\big)_{(jk),(jk)} - \big(\boldsymbol{\Theta}^* \otimes \boldsymbol{\Theta}^*\big)_{(jk),(jk)}\big| = O_{\mathbb{P}}\bigg(s\sqrt{\frac{\log d}{n}}\bigg). \qquad \text{(C.14)}
$$

Submitting (C.14) into (C.11), we can obtain that

$$
\big(\widehat{\boldsymbol{\Theta}} \otimes \widehat{\boldsymbol{\Theta}}\big)_{(jk),(jk)} \ge \frac{1}{\nu^2} - O_{\mathbb{P}}\bigg(s\sqrt{\frac{\log d}{n}}\bigg) \ge \frac{1}{2\nu^2},
$$

when $n$ is sufficiently large. Therefore, we have (i).a $= 1/\big(\widehat{\boldsymbol{\Theta}} \otimes \widehat{\boldsymbol{\Theta}}\big)_{(jk),(jk)} \le 2\nu^2$. Now we are going to bound term (i).b. We have

$$
\begin{aligned}
\text{(i).b} \le{}& \big\|\big((\widehat{\boldsymbol{\Theta}} - \boldsymbol{\Theta}^*) \otimes (\widehat{\boldsymbol{\Theta}} - \boldsymbol{\Theta}^*)\big)_{(jk)^c,(jk)}\big\|_1 + \big\|\big((\widehat{\boldsymbol{\Theta}} - \boldsymbol{\Theta}^*) \otimes \boldsymbol{\Theta}^*\big)_{(jk)^c,(jk)}\big\|_1 \\
&+ \big\|\big((\boldsymbol{\Theta}^* \otimes (\widehat{\boldsymbol{\Theta}} - \boldsymbol{\Theta}^*)\big)_{(jk)^c,(jk)}\big\|_1 \\
={}& \underbrace{\big\|(\widehat{\boldsymbol{\Theta}} - \boldsymbol{\Theta}^*)_{*j} \otimes (\widehat{\boldsymbol{\Theta}} - \boldsymbol{\Theta}^*)_{*k}\big\|_1}_{I_1} + \underbrace{\big\|(\widehat{\boldsymbol{\Theta}} - \boldsymbol{\Theta}^*)_{*j} \otimes \boldsymbol{\Theta}^*_{*k}\big\|_1}_{I_2} + \underbrace{\big\|\boldsymbol{\Theta}^*_{*j} \otimes (\widehat{\boldsymbol{\Theta}} - \boldsymbol{\Theta}^*)_{*k}\big\|_1}_{I_3}.
\end{aligned}
$$

For $I_1$, we have

$$
I_1 = \big\|(\widehat{\boldsymbol{\Theta}} - \boldsymbol{\Theta}^*)_{*j}\big\|_1 \cdot \big\|(\widehat{\boldsymbol{\Theta}} - \boldsymbol{\Theta}^*)_{*k}\big\|_1 \le \big\|\widehat{\boldsymbol{\Theta}} - \boldsymbol{\Theta}^*\big\|_1^2 = O_{\mathbb{P}}\big(s^2 \log d/n\big).
$$

For $I_2$, we have

$$
I_2 \le \big\|(\widehat{\boldsymbol{\Theta}} - \boldsymbol{\Theta}^*)_{*j}\big\|_1 \cdot \big\|\boldsymbol{\Theta}^*_{*k}\big\|_1 \le \big\|\widehat{\boldsymbol{\Theta}} - \boldsymbol{\Theta}^*\big\|_1 \cdot M = O_{\mathbb{P}}\big(s\sqrt{\log d/n}\big),
$$

where the second inequality follows from the fact that $\boldsymbol{\Theta}^* \in \mathcal{U}(s, M)$. Similarly, we have $I_3 = O_{\mathbb{P}}\big(s\sqrt{\log d/n}\big)$. Combining terms $I_1, I_2$ and $I_3$, we obtain that (i).b $= O_{\mathbb{P}}\big(s^2 \log d/n + s\sqrt{\log d/n}\big)$. Combining terms (i).a and (i).b and substituting it into (C.10), we obtain that

$$
\text{(i)} = O_{\mathbb{P}}\bigg(s^2\frac{\log d}{n} + s\sqrt{\frac{\log d}{n}}\bigg). \qquad \text{(C.15)}
$$

**Bounding term** (ii)**:** We have

$$
\begin{aligned}
\text{(ii)} = &\underbrace{\frac{1}{(\widehat{\boldsymbol{\Theta}} \otimes \widehat{\boldsymbol{\Theta}})_{(jk),(jk)} (\boldsymbol{\Theta}^* \otimes \boldsymbol{\Theta}^*)_{(jk),(jk)}}}_{\text{(ii).a}} \underbrace{\left|(\widehat{\boldsymbol{\Theta}} \otimes \widehat{\boldsymbol{\Theta}})_{(jk),(jk)} - (\boldsymbol{\Theta}^* \otimes \boldsymbol{\Theta}^*)_{(jk),(jk)}\right|}_{\text{(ii).b}} \\
&\cdot \underbrace{\left\|(\boldsymbol{\Theta}^* \otimes \boldsymbol{\Theta}^*)_{(jk)^c,(jk)}\right\|_1}_{\text{(ii).c}}. 
\end{aligned} \tag{C.16}
$$

For term (ii).a, by similar technique used for bounding (i).a, when $n$ is sufficiently large, we have

$$
\frac{1}{(\widehat{\boldsymbol{\Theta}} \otimes \widehat{\boldsymbol{\Theta}})_{(jk),(jk)} (\boldsymbol{\Theta}^* \otimes \boldsymbol{\Theta}^*)_{(jk),(jk)}} \le 2\nu^4. \tag{C.17}
$$

For term (ii).b, we have already bounded the same term in (C.14), i.e.,

$$
\left|(\widehat{\boldsymbol{\Theta}} \otimes \widehat{\boldsymbol{\Theta}})_{(jk),(jk)} - (\boldsymbol{\Theta}^* \otimes \boldsymbol{\Theta}^*)_{(jk),(jk)}\right| = O_{\mathbb{P}}\left(s\sqrt{\frac{\log d}{n}}\right). \tag{C.18}
$$

For term (ii).c, we have

$$
\text{(ii).c} = \left\|(\boldsymbol{\Theta}^* \otimes \boldsymbol{\Theta}^*)_{(jk)^c,(jk)}\right\|_1 = \|\boldsymbol{\Theta}^*_{*j}\|_1 \cdot \|\boldsymbol{\Theta}^*_{*k}\|_1 \le M^2, \tag{C.19}
$$

since $\boldsymbol{\Theta}^* \in \mathcal{U}(s, M)$. Substituting (C.17), (C.18) and (C.19) into (C.16), we obtain that

$$
\text{(ii)} = O_{\mathbb{P}}\left(s\sqrt{\frac{\log d}{n}}\right), \tag{C.20}
$$

since $\nu = O(1)$. Combining the above bounds in (C.9), (C.15) and (C.20), we obtain that

$$
\left\|\widehat{\mathbf{w}} - \mathbf{w}^*\right\|_1 \le O_{\mathbb{P}}\left(s^2\frac{\log d}{n} + s\sqrt{\frac{\log d}{n}}\right).
$$

□

## C.4 Proof of Lemma B.5

*Proof.* Denote $\widehat{R}_{(jk),(j'k')}$ by

$$
\widehat{R}_{(jk),(j'k')}(\boldsymbol{\Sigma}^{-1}) = \frac{1}{n(n-1)^2} \sum_{i=1}^{n} \sum_{i' \ne i, i'' \ne i} \sqrt{1 - \Sigma_{jk}^2} \cdot \sqrt{1 - \Sigma_{j'k'}^2} \cdot h_{(jk)(j'k')}^{i,i',i''}(\Sigma_{jk}, \Sigma_{j'k'}),
$$

where $h^{i,i',i''}_{(jk)(j'k')}(\Sigma_{jk},\Sigma_{j'k'}) = h^{ii'}_{jk}(\Sigma_{jk}) \cdot h^{ii''}_{j'k'}(\Sigma_{j'k'})$ is calculated as

$$h^{i,i',i''}_{(jk),(j'k')}(\Sigma_{jk},\Sigma_{j'k'}) = \Big\{ -\frac{\pi}{2}\,\mathrm{sign}\,\big[\big(X_{ij}-X_{i'j}\big)\big(X_{ik}-X_{i'k}\big)\big] + \arcsin\big(\Sigma_{jk}\big)\Big\} \cdot \Big\{ -\frac{\pi}{2}\,\mathrm{sign}\,\big[\big(X_{ij'}-X_{i''j'}\big)\big(X_{ik'}-X_{i''k'}\big)\big] + \arcsin\big(\Sigma_{j'k'}\big)\Big\}.$$

Recall that $\widehat{\mathbf{R}} = \widehat{\mathbf{R}}(\widehat{\mathbf{\Sigma}}^{-1})$. We have

$$\big\|\widehat{\mathbf{R}}(\widehat{\mathbf{\Sigma}}^{-1}) - \mathbf{R}\big\|_{\max} = \underbrace{\big\|\widehat{\mathbf{R}}(\widehat{\mathbf{\Sigma}}^{-1}) - \widehat{\mathbf{R}}(\mathbf{\Sigma}^{*-1})\big\|_{\max}}_{\text{(i)}} + \underbrace{\big\|\widehat{\mathbf{R}}(\mathbf{\Sigma}^{*-1}) - \mathbf{R}\big\|_{\max}}_{\text{(ii)}}.$$

In the following, we bound terms (i) and (ii) respectively.
**Bounding Term** (i)**:** Let $\widetilde{h}^{i,i',i''}_{(jk)(j'k')}(\Sigma_{jk},\Sigma_{j'k'}) = \sqrt{1-\Sigma_{jk}^2}\cdot\sqrt{1-\Sigma_{j'k'}^2}\cdot h^{i,i',i''}_{(jk)(j'k')}(\Sigma_{jk},\Sigma_{j'k'})$. Since

$$\frac{\partial \widetilde{h}^{i,i',i''}_{(jk),(j'k')}(\Sigma_{jk},\Sigma_{j'k'})}{\partial \Sigma_{jk}} = \frac{-\Sigma_{jk}}{\sqrt{1-\Sigma_{jk}^2}} \cdot \sqrt{1-\Sigma_{j'k'}^2} \cdot h^{i,i',i''}_{(jk),(j'k')}(\Sigma_{jk},\Sigma_{j'k'}) + \Big\{ -\frac{\pi}{2}\,\mathrm{sign}\,\big[\big(X_{ij}-X_{i'j}\big)\big(X_{ik}-X_{i'k}\big)\big] + \arcsin\big(\Sigma_{j'k'}\big)\Big\}\sqrt{1-\Sigma_{j'k'}^2},$$

By Assumption 4.2 and $\mathrm{diag}(\mathbf{\Sigma}^*) = \mathbf{I}$, we have $\Sigma^*_{jk} \le \xi < 1$ for $j \neq k$, we have for any $\widetilde{\Sigma}_{jk} = \Sigma^*_{jk} + t(\widehat{\Sigma}_{jk} - \Sigma^*_{jk})$ with $t \in [0,1]$ that

$$\bigg[\frac{\partial \widetilde{h}^{i,i',i''}_{(jk),(j'k')}(\Sigma_{jk},\Sigma_{j'k'})}{\partial \Sigma_{jk}}\bigg]\bigg|_{\Sigma_{jk}=\widetilde{\Sigma}_{jk}} \le \frac{C\xi}{\sqrt{1-\xi^2}}\pi^2 + \pi, \tag{C.21}$$

where $C$ is a constant. Therefore, by mean value theorem, we have

$$\begin{aligned}
&\big|\widehat{R}_{(jk),(j'k')}(\widehat{\mathbf{\Sigma}}^{-1}) - \widehat{R}_{(jk),(j'k')}(\mathbf{\Sigma}^{*-1})\big| \\
&= \bigg\langle \frac{1}{n(n-1)^2}\sum_{i=1}^{n}\sum_{i'\neq i, i''\neq i}\bigg[\frac{\partial \widetilde{h}^{i,i',i''}_{(jk),(j'k')}(\Sigma_{jk},\Sigma_{j'k'})}{\partial \Sigma_{jk}}\bigg]\bigg|_{\Sigma_{jk}=\widetilde{\Sigma}_{jk}}, \widehat{\Sigma}_{jk} - \Sigma^*_{jk}\bigg\rangle \\
&\quad + \bigg\langle \frac{1}{n(n-1)^2}\sum_{i=1}^{n}\sum_{i'\neq i, i''\neq i}\bigg[\frac{\partial \widetilde{h}^{i,i',i''}_{(jk),(j'k')}(\Sigma_{jk},\Sigma_{j'k'})}{\partial \Sigma_{j'k'}}\bigg]\bigg|_{\Sigma_{jk}=\widetilde{\Sigma}_{jk}}, \widehat{\Sigma}_{j'k'} - \Sigma^*_{j'k'}\bigg\rangle \\
&\le \bigg(\frac{C\xi}{\sqrt{1-\xi^2}}\pi^2 + \pi\bigg) \cdot \Big(\big|\widehat{\Sigma}_{jk} - \Sigma^*_{jk}\big| + \big|\widehat{\Sigma}_{j'k'} - \Sigma^*_{j'k'}\big|\Big),
\end{aligned} \tag{C.22}$$

where $\widetilde{\Sigma}_{jk} = \Sigma^*_{jk} + t(\widehat{\Sigma}_{jk} - \Sigma^*_{jk})$ for some $t \in [0,1]$, and the last inequality follows from (C.21).

Using (C.22), by union bound, we have

$$(\mathrm{i}) = \big\|\widehat{\mathbf{R}}(\widehat{\mathbf{\Sigma}}^{-1}) - \widehat{\mathbf{R}}(\mathbf{\Sigma}^{*-1})\big\|_{\max} \leq 2 \cdot \Big(\frac{\xi}{\sqrt{1-\xi^2}}\pi^2 + \pi\Big) \cdot \big\|\widehat{\mathbf{\Sigma}} - \mathbf{\Sigma}\big\|_{\max} = O_{\mathbb{P}}\bigg(\sqrt{\frac{\log d}{n}}\bigg). \tag{C.23}$$

**Bounding Term** (ii)**:** By the definition of $\mathbf{R}$ and $h_{jk}^{ii'|i}$, we have

$$\begin{aligned} R_{(jk),(j'k')} &= \frac{1}{n(n-1)^2}\sum_{i=1}^{n}\sum_{i'\neq i, i''\neq i, i'\neq i''} F_{jk}\cdot F_{j'k'}\cdot \mathbb{E}\big[h_{jk}^{ii'|i}h_{j'k'}^{ii''|i}\big] \\ &+ \frac{1}{n(n-1)^2}\sum_{i=1}^{n}\sum_{i'\neq i} F_{jk}\cdot F_{j'k'}\cdot \mathbb{E}\big[h_{jk}^{ii'|i}\cdot h_{j'k'}^{ii'|i}\big] \\ &= \frac{1}{n(n-1)^2}\sum_{i=1}^{n}\sum_{i'\neq i, i''\neq i} F_{jk}F_{j'k'}\mathbb{E}\big[h_{jk}^{ii'}h_{jk}^{ii''}\big] + \frac{1}{n(n-1)^2}\sum_{i=1}^{n}\sum_{i'\neq i} F_{jk}F_{j'k'}\mathbb{E}\big[h_{jk}^{ii'|i}\cdot h_{j'k'}^{ii'|i}\big], \end{aligned}$$

where the last step follows from the fact that $h_{jk}^{ii'}$ and $h_{jk}^{ii''}$ are independent given $\boldsymbol{X}_i$. Recall that $h_{(jk),(j'k')}^{i,i',i''} = h_{jk}^{ii'}h_{j'k'}^{ii''}$, we have

$$\begin{aligned} &[\widehat{R}(\mathbf{\Sigma}^{*-1})]_{(jk),(j'k')} - R_{(jk),(j'k')} \\ &= \underbrace{\bigg\{\frac{1}{n(n-1)(n-2)}\sum_{i'\neq i, i''\neq i, i'\neq i''} F_{jk}F_{j'k'}\Big[h_{(jk),(j'k')}^{i,i',i''}(\Sigma_{jk}^*, \Sigma_{j'k'}^*) - \mathbb{E}\big[h_{(jk),(j'k')}^{i,i',i''}(\Sigma_{jk}^*, \Sigma_{j'k'}^*)\big]\Big]\bigg\}}_{I_1} \\ &\cdot \frac{n-2}{n-1} + \frac{1}{n-1}\underbrace{\bigg\{\frac{1}{n(n-1)}\sum_{i'\neq i} F_{jk}F_{j'k'}\Big[h_{jk}^{ii'|i}\cdot h_{j'k'}^{ii'|i} - \mathbb{E}\big[h_{jk}^{ii'|i}\cdot h_{j'k'}^{ii'|i}\big]\Big]\bigg\}}_{I_2}. \end{aligned} \tag{C.24}$$

For term $I_1$, it can be seen that $I_1$ is zero mean third order U-statistic. Notice that $|F_{jk}F_{j'k'}h_{(jk),(j'k')}^{i,i',i''}| \leq 2\pi$, by Hoeffding's inequality for U-statistics (Hoeffding, 1963), we have

$$\mathbb{P}(|I_1| > t) \leq 2\exp\bigg(-\frac{Ct^2}{4\pi^2}\Big\lfloor\frac{n}{3}\Big\rfloor\bigg). \tag{C.25}$$

For term $I_2$, it can be seen that $I_2$ is a zero mean second-order U-statistic. Moreover, we have $|F_{jk}F_{j'k'}\cdot h_{jk}^{ii'|i}\cdot h_{j'k'}^{ii'|i}| \leq 4\pi^2$. Thus, by Hoeffding's inequality for U-statistics (Hoeffding, 1963), we have

$$\mathbb{P}(|I_2| > t) \leq 2\exp\bigg(-\frac{Ct^2}{16\pi^4}\Big\lfloor\frac{n}{2}\Big\rfloor\bigg). \tag{C.26}$$

Substituting (C.25) and (C.26) into (C.24), and invoking union bound, we have

$$\mathbb{P}\big(\big\|\widehat{\mathbf{R}}(\mathbf{\Sigma}^{*-1}) - \mathbf{R}\big\|_{\max} > t\big) \leq \frac{n-1}{n} 2d^2 \exp\left(-\frac{Ct^2}{4\pi^2}\left\lfloor\frac{n}{3}\right\rfloor\right) + \frac{1}{n-1} 2d^2 \exp\left(-\frac{Ct^2}{16\pi^4}\left\lfloor\frac{n}{2}\right\rfloor\right),$$

which immediately yields

$$\text{(ii)} = \big\|\widehat{\mathbf{R}}(\mathbf{\Sigma}^{*-1}) - \mathbf{R}\big\|_{\max} = O_{\mathbb{P}}\big(\sqrt{\log d/n}\big). \tag{C.27}$$

Finally, we combine (C.23) and (C.27) and complete the proof. □

## C.5 Proof of Lemma B.6

*Proof.* Recall that we have $S_n(\mathbf{\Theta}) = \mathbf{e}_k^\top \mathbf{\Theta}^\top \widehat{\mathbf{\Sigma}} \mathbf{\Theta} \mathbf{e}_k - \Theta_{jk}/(\Theta_{jj}\Theta_{kk})$, so we can obtain

$$\begin{aligned}
&\left| S_n(\Theta_{jk}, \mathbf{\Theta}_{(jk)^c}) - S_n(0, \mathbf{\Theta}_{(jk)^c}) - \frac{\Theta_{jk}\left[\Theta_{jj}\Theta_{kk} + \Theta_{jk}^2\right]}{\left(\Theta_{jj}\Theta_{kk}\right)^2} \right| \\
&= \left| \frac{\Theta_{jk}\big(\widehat{\mathbf{\Sigma}}_{k*}\mathbf{\Theta}_{*k} + \widehat{\mathbf{\Sigma}}_{j*}\mathbf{\Theta}_{*j} - 1\big)}{\Theta_{jj}\Theta_{kk}} - \frac{\Theta_{jk}\left[\Theta_{jj}\Theta_{kk} + \Theta_{jk}^2\right]}{\left(\Theta_{jj}\Theta_{kk}\right)^2} \right| \\
&= |\Theta_{jk}| \cdot \left| \frac{\big([\widehat{\mathbf{\Sigma}}\mathbf{\Theta}]_{kk} + [\widehat{\mathbf{\Sigma}}\mathbf{\Theta}]_{jj} - 2\big)\Theta_{jj}\Theta_{kk} - \Theta_{jk}^2}{\left(\Theta_{jj}\Theta_{kk}\right)^2} \right|.
\end{aligned} \tag{C.28}$$

The right hand side of (C.28) can be further bounded by

$$\underbrace{|\Theta_{jk}| \cdot \left| \frac{\big([\widehat{\mathbf{\Sigma}}\mathbf{\Theta}]_{kk} + [\widehat{\mathbf{\Sigma}}\mathbf{\Theta}]_{jj} - 2\big)}{\Theta_{jj}\Theta_{kk}} \right|}_{\text{(i)}} + \underbrace{|\Theta_{jk}| \cdot \left| \frac{\Theta_{jk}^2}{\left(\Theta_{jj}\Theta_{kk}\right)^2} \right|}_{\text{(ii)}}.$$

In particular, we have

$$[\widehat{\mathbf{\Sigma}}\mathbf{\Theta}]_{kk} = [(\widehat{\mathbf{\Sigma}} - \mathbf{\Sigma})\mathbf{\Theta}]_{kk} + 1 \leq \|\widehat{\mathbf{\Sigma}} - \mathbf{\Sigma}\|_{\max} \cdot \|\mathbf{\Theta}\|_1 + 1 \leq \sqrt{\frac{\log d}{n}} M + 1 \tag{C.29}$$

Substituting (C.29) into (i), we obtain (i) $\leq K \cdot n^{-\eta}\sqrt{\log d/n} M/(\nu^2)$. For term (ii), we have (ii) $\leq K^3 \cdot n^{-3\eta}/(\nu^4)$. Thus, when $\eta \geq 1/2$, we have

$$\left| S_n(\Theta_{jk}, \mathbf{\Theta}_{(jk)^c}) - S_n(0, \mathbf{\Theta}_{(jk)^c}) - \frac{\Theta_{jk}\left[\Theta_{jj}\Theta_{kk} + \Theta_{jk}^2\right]}{\left(\Theta_{jj}\Theta_{kk}\right)^2} \right| \leq C \cdot K \cdot n^{-\eta}\sqrt{\frac{\log d}{n}}.$$

Since the above argument is for any $\mathbf{\Theta} \in \mathcal{U}_1(K, \eta, s^*, M^*)$, this completes the proof. □

## C.6 Proof of Lemma B.7

*Proof.* The proof is similar to Lemma B.5. We define several events as follows:

$$\mathcal{E}_1 = \Big\{ \big\|\widehat{\boldsymbol{\Sigma}} - \boldsymbol{\Theta}^{-1}\big\|_{\max} \le C_4\sqrt{\log d/n} \Big\}.$$

We have

$$\big\|\widehat{\mathbf{R}}(\widehat{\boldsymbol{\Sigma}}^{-1}) - \mathbf{R}\big\|_{\max} = \underbrace{\big\|\widehat{\mathbf{R}}(\widehat{\boldsymbol{\Sigma}}^{-1}) - \widehat{\mathbf{R}}(\boldsymbol{\Sigma}^{-1})\big\|_{\max}}_{\text{(i)}} + \underbrace{\big\|\widehat{\mathbf{R}}(\boldsymbol{\Sigma}^{-1}) - \mathbf{R}\big\|_{\max}}_{\text{(ii)}}.$$

In the following, we will bound terms (i) and (ii) respectively.
**Bounding Term (i):** By the similar argument as in the proof of Lemma B.5, we have

$$\text{(i)} = \big\|\widehat{\mathbf{R}}(\widehat{\boldsymbol{\Sigma}}^{-1}) - \widehat{\mathbf{R}}(\boldsymbol{\Sigma}^{-1})\big\|_{\max} \le 2\cdot\Big(\frac{\xi}{\sqrt{1-\xi^2}}\pi^2 + \pi\Big)\cdot \|\widehat{\boldsymbol{\Sigma}} - \boldsymbol{\Sigma}\|_{\max}. \tag{C.30}$$

Under event $\mathcal{E}_1$, we obtain that

$$\big\|\widehat{\mathbf{R}}(\widehat{\boldsymbol{\Sigma}}^{-1}) - \widehat{\mathbf{R}}(\boldsymbol{\Sigma}^{-1})\big\|_{\max} \le C\cdot\sqrt{\log d/n}. \tag{C.31}$$

**Bounding Term (ii):** By the similar proof of Lemma B.5, we have

$$\begin{aligned}
&[\widehat{R}(\boldsymbol{\Sigma}^{-1})]_{(jk),(j'k')} - R_{(jk),(j'k')} \\
&= \underbrace{\Big\{\frac{1}{n(n-1)(n-2)}\sum_{i'\neq i, i''\neq i, i'\neq i''} F_{jk}F_{j'k'}\Big[h^{i,i',i''}_{(jk),(j'k')}(\Sigma_{jk},\Sigma_{j'k'}) - \mathbb{E}\big[h^{i,i',i''}_{(jk),(j'k')}(\Sigma_{jk},\Sigma_{j'k'})\big]\Big]\Big\}}_{I_1} \\
&\quad\cdot\frac{n-2}{n-1} + \frac{1}{n-1}\underbrace{\Big\{\frac{1}{n(n-1)}\sum_{i'\neq i} F_{jk}F_{j'k'}\Big[h^{ii'|i}_{jk}\cdot h^{ii'|i}_{j'k'} - \mathbb{E}\big[h^{ii'|i}_{jk}\cdot h^{ii'|i}_{j'k'}\big]\Big]\Big\}}_{I_2}.
\end{aligned} \tag{C.32}$$

For term $I_1$, it can be seen that $I_1$ is zero mean third order U-statistic. Notice that $|F_{jk}F_{j'k'}h^{i,i',i''}_{(jk),(j'k')}| \le 2\pi$ for all $\boldsymbol{\Theta} \in \mathcal{U}_1(K,\eta,s^*,M^*)$, by Hoeffding's inequality for U-statistics (Hoeffding, 1963), we have

$$\mathbb{P}(|I_1| > t) \le 2\exp\Big(-\frac{Ct^2}{4\pi^2}\Big\lfloor\frac{n}{3}\Big\rfloor\Big) \tag{C.33}$$

for all $\boldsymbol{\Theta} \in \mathcal{U}_1(K,\eta,s^*,M^*)$. For term $I_2$, it can be seen that $I_2$ is a zero mean second-order U-statistic. Moreover, we have $|F_{jk}F_{j'k'}\cdot h^{ii'|i}_{jk}\cdot h^{ii'|i}_{j'k'}| \le 4\pi^2$ for all $\boldsymbol{\Theta} \in \mathcal{U}_1(K,\eta,s^*,M^*)$. Thus, by Hoeffding's inequality for U-statistics (Hoeffding, 1963), we have

$$\mathbb{P}(|I_2| > t) \le 2\exp\Big(-\frac{Ct^2}{16\pi^4}\Big\lfloor\frac{n}{2}\Big\rfloor\Big) \tag{C.34}$$

for all $\boldsymbol{\Theta} \in \mathcal{U}_1(K, \eta, s^*, M^*)$ Substituting (C.33) and (C.34) into (C.32), and invoking union bound, we have

$$\mathbb{P}\big(\|\widehat{\mathbf{R}}(\boldsymbol{\Sigma}^{-1}) - \mathbf{R}\|_{\max} > t\big) \leq \frac{n-1}{n} 2d^2 \exp\Big(-\frac{Ct^2}{4\pi^2}\Big\lfloor\frac{n}{3}\Big\rfloor\Big) + \frac{1}{n-1} 2d^2 \exp\Big(-\frac{Ct^2}{16\pi^4}\Big\lfloor\frac{n}{2}\Big\rfloor\Big)$$

for all $\boldsymbol{\Theta} \in \mathcal{U}_1(K, \eta, s^*, M^*)$. It immediately yields that,

$$\big\|\widehat{\mathbf{R}}(\boldsymbol{\Sigma}^{-1}) - \mathbf{R}\big\|_{\max} \leq C' \cdot \sqrt{\log d/n}. \tag{C.35}$$

Finally, we combine (C.30) and (C.35), yielding

$$\inf_{\boldsymbol{\Theta} \in \mathcal{U}_1(K,\eta,s^*,M^*)} \mathbb{P}_{\boldsymbol{\Theta}}\left(\big\|\widehat{\mathbf{R}} - \mathbf{R}\big\|_{\max} \leq C''\sqrt{\frac{\log d}{n}}\right) \geq \inf_{\boldsymbol{\Theta} \in \mathcal{U}_1(K,\eta,s^*,M^*)} \mathbb{P}_{\boldsymbol{\Theta}}\big(\mathcal{E}_1\big) \to 1.$$

This completes the proof. □

## C.7 Proof of Lemma B.8

*Proof.* The proof is identical to the proof of Lemma C.1, followed by applying Berry-Essen Theorem (Van der Vaart, 1998). □

## C.8 Proof of Lemma B.9

*Proof.* It is similar to the proof of Lemma B.3. Hence we omit it. □

## C.9 Proof of Lemma B.10

*Proof.* First, we prove (B.16). By the similar proof of Lemma B.2, we obtain

$$\|\widehat{\boldsymbol{\Theta}}^{-1} - \boldsymbol{\Theta}^{-1}\|_{\max} = \|R(\widehat{\boldsymbol{\Delta}})\|_{\max} \leq 2K^2_{\boldsymbol{\Sigma}^*}\|\widehat{\boldsymbol{\Delta}}\|_{\max}.$$

By Assumption 4.9 and the fact that $\nu = O(1)$, we have

$$\lim_{n\to\infty} \inf_{\boldsymbol{\Theta} \in \mathcal{U}_1(K,\eta,s^*,M^*)} \mathbb{P}_{\boldsymbol{\Theta}}\left(\|\widehat{\boldsymbol{\Theta}}^{-1} - \boldsymbol{\Theta}^{-1}\|_{\max} \leq C_5\sqrt{\frac{\log d}{n}}\right) = 1, \tag{C.36}$$

where $C_5$ is a constant.

Now we prove (B.17). In particular, we have $\big\| - \widehat{\boldsymbol{\Sigma}} + \widehat{\boldsymbol{\Theta}}^{-1}\big\|_{\max} \leq \big\| - \widehat{\boldsymbol{\Sigma}} + \boldsymbol{\Sigma}\big\|_{\max} + \big\| - \boldsymbol{\Theta}^{-1} + \widehat{\boldsymbol{\Theta}}^{-1}\big\|_{\max}$. By Theorem 4.2 in (Liu et al., 2012) and (C.36), we have

$$\lim_{n\to\infty} \inf_{\boldsymbol{\Theta} \in \mathcal{U}_1(K,\eta,s^*,M^*)} \mathbb{P}_{\boldsymbol{\Theta}}\left(\big\| - \widehat{\boldsymbol{\Sigma}} + \widehat{\boldsymbol{\Theta}}^{-1}\big\|_{\max} \leq C_6\sqrt{\frac{\log d}{n}}\right) = 1.$$

This completes the proof. □

### C.10 Proof of Lemma B.11

*Proof.* The proof of this lemma is analogous to the proof of Theorem 4.7, but studies the local alternative hypothesis, i.e., $\mathcal{U}_1(K, \eta, s^*, M^*)$. In particular, we define several events:

$$\mathcal{E}_0 = \Big\{ \big\|\widehat{\mathbf{\Theta}} - \mathbf{\Theta}\big\|_{\max} \le C_0\sqrt{\log d/n} \Big\}, \mathcal{E}_6 = \Big\{ \|\widehat{\mathbf{\Theta}}^{-1} - \widehat{\mathbf{\Sigma}}\|_{\max} \le C_6\sqrt{\frac{\log d}{n}} \Big\},$$
$$\mathcal{E}_3 = \Big\{ \big\|\widehat{\mathbf{w}}(\widehat{\mathbf{\Theta}}) - \mathbf{w}\big\|_1 \le C_3\Big( s^2\frac{\log d}{n} + s\sqrt{\frac{\log d}{n}} \Big) \Big\}.$$

We first prove (B.18). By Taylor expansion, we obtain

$$\widehat{S}_n\big(0, \widehat{\mathbf{\Theta}}_{(jk)^c}\big) = S_n(\mathbf{\Theta}) - K\cdot n^{-\eta}\widetilde{\sigma}^2 + \underbrace{(\mathbf{w} - \widehat{\mathbf{w}})^\top \mathrm{vec}\big(-\widehat{\mathbf{\Sigma}}_{(jk)^c} + [\widehat{\mathbf{\Theta}}^{-1}]_{(jk)^c}\big)}_{\text{(i)}}$$
$$\underbrace{-[\mathbf{\Theta}^{-1}]_{jk} + [\widehat{\mathbf{\Theta}}^{-1}]_{jk} - \mathbf{w}^\top \mathrm{vec}\big([\widehat{\mathbf{\Theta}}^{-1}]_{(jk)^c} - [\mathbf{\Theta}^{-1}]_{(jk)^c}\big)}_{\text{(ii)}} + \underbrace{S_n\big(0, \mathbf{\Theta}_{(jk)^c}\big) - S_n(\mathbf{\Theta}) + K\cdot n^{-\eta}\widetilde{\sigma}^2}_{\text{(iii)}}. \tag{C.37}$$

In the following, we are going to bound terms (i), (ii) and (iii) respectively.
**Bounding term** (i)**:** We have that

$$\text{(i)} \le \big\|\widehat{\mathbf{w}} - \mathbf{w}\big\|_1 \cdot \big\|\mathrm{vec}\big(-\widehat{\mathbf{\Sigma}}_{(jk)^c} + [\widehat{\mathbf{\Theta}}^{-1}]_{(jk)^c}\big)\big\|_\infty \le \big\|\widehat{\mathbf{w}} - \mathbf{w}\big\|_1 \cdot \big\| - \widehat{\mathbf{\Sigma}} + \widehat{\mathbf{\Theta}}^{-1}\big\|_{\max}, \tag{C.38}$$

where the first inequality follows from Hölder's inequality, and the second inequality follows from $\|\mathrm{vec}(\mathbf{A})\|_\infty = \|\mathbf{A}\|_{\max}$. Given event $\mathcal{E}_3$ and $\mathcal{E}_6$, we have

$$\Big|(\mathbf{w} - \widehat{\mathbf{w}})^\top \mathrm{vec}\big(-\widehat{\mathbf{\Sigma}}_{(jk)^c} + [\widehat{\mathbf{\Theta}}^{-1}]_{(jk)^c}\big)\Big| \le C_3C_6\bigg[ s^2\Big(\frac{\log d}{n}\Big)^{3/2} + s\frac{\log d}{n} \bigg]. \tag{C.39}$$

**Bounding term** (ii)**:** By the similar proof of Theorem 4.7, we obtain

$$\text{(ii)} = \underbrace{-\mathbf{H}_{(jk),(jk)^c} - \big\langle \mathbf{w}, -\mathbf{H}_{(jk)^c,(jk)^c}\mathrm{vec}(\widehat{\mathbf{\Delta}}_{(jk)^c})\big\rangle}_{\text{(ii).a}} + \underbrace{[\mathbf{R}_1(\widehat{\mathbf{\Delta}})]_{jk}}_{\text{(ii).b}} + \underbrace{\big\langle \mathbf{w}, \mathrm{vec}\big([\mathbf{R}_1(\widehat{\mathbf{\Delta}})]_{(jk)^c}\big)\big\rangle}_{\text{(ii).c}}. \tag{C.40}$$

By the definition of $\mathbf{w} = [\mathbf{H}_{(jk)^c,(jk)^c}]^{-1}\mathbf{H}_{(jk),(jk)^c}$, we have (ii).a $= 0$. It remains to bound terms (ii).b and (ii).c. Recall that in Theorem 4.7, we have proved that $\big\|\mathbf{R}_1(\widehat{\mathbf{\Delta}})\big\|_{\max} = O_{\mathbb{P}}\big(s\log d/n\big)$. Then given event $\mathcal{E}_0$, we have

$$\big|[\mathbf{R}_1(\widehat{\mathbf{\Delta}})]_{(jk)^c}\big| \le \big\|\mathbf{R}_1(\widehat{\mathbf{\Delta}})\big\|_{\max} \le C_0^2 s\log d/n.$$

Similarly, we can show that

$$\text{(ii).c} \le \|\mathbf{w}\|_1 \cdot \big\|[\mathbf{R}_1(\widehat{\boldsymbol{\Delta}})]_{(jk)^c}\big\|_\infty \le \|\mathbf{w}\|_1 \cdot \|\mathbf{R}_1(\widehat{\boldsymbol{\Delta}})\|_{\max}.$$

Thus, given event $\mathcal{E}_0$, we have

$$\Big|\big\langle \mathbf{w}, \mathrm{vec}\big([\mathbf{R}_1(\widehat{\boldsymbol{\Delta}})]_{(jk)^c}\big)\big\rangle\Big| \le C' s \log d/n,$$

since $\|\mathbf{w}\|_1 \le \nu^2 M^2$. Combining terms (ii).a, (ii).b and (ii).c, under event $\mathcal{E}_0$, we can show that

$$\Big| - [\boldsymbol{\Theta}^{-1}]_{jk} + [\widehat{\boldsymbol{\Theta}}^{-1}]_{jk} - \mathbf{w}^\top \mathrm{vec}\big([\widehat{\boldsymbol{\Theta}}^{-1}]_{(jk)^c} - [\boldsymbol{\Theta}^{-1}]_{(jk)^c}\big)\Big| \le C'' \log d/n, \tag{C.41}$$

**Bounding term** (iii)**:** By Lemma B.6, we have

$$\Big|S_n\big(0, \boldsymbol{\Theta}_{(jk)^c}\big) - S_n(\boldsymbol{\Theta}) + K \cdot n^{-\eta}\widetilde{\sigma}^2\Big| \le C_0 \cdot K \cdot n^{-\eta}\sqrt{\log d/n}. \tag{C.42}$$

Let $\eta(n) = \sqrt{n}/\kappa\Big[C_3 C_6 s^2 (\log d/n)^{3/2} + C s \log d/n + C_0 \cdot K \cdot n^{-\eta}\sqrt{\log d/n}\Big]$, and substitute (C.39) and (C.41) and (C.42) into (C.37), together with Lemma B.9, we obtain

$$\begin{aligned}
&\inf_{\boldsymbol{\Theta}\in\mathcal{U}_1(K,\eta,s^*,M^*)} \mathbb{P}_{\boldsymbol{\Theta}}\Big[\Big|\widehat{S}\big(0, \widehat{\boldsymbol{\Theta}}_{(jk)^c}\big) - S\big(0, \boldsymbol{\Theta}_{(jk)^c}\big)\Big| \le \eta(n)\Big] \\
&\ge \inf_{\boldsymbol{\Theta}\in\mathcal{U}_1(K,\eta,s^*,M^*)} \mathbb{P}_{\boldsymbol{\Theta}}(\mathcal{E}_0 \cap \mathcal{E}_3 \cap \mathcal{E}_6) \\
&\ge 1 - \sup_{\boldsymbol{\Theta}\in\mathcal{U}_1(K,\eta,s^*,M^*)} \mathbb{P}_{\boldsymbol{\Theta}}(\mathcal{E}_0) - \sup_{\boldsymbol{\Theta}\in\mathcal{U}_1(K,\eta,s^*,M^*)} \mathbb{P}_{\boldsymbol{\Theta}}(\mathcal{E}_3) - \sup_{\boldsymbol{\Theta}\in\mathcal{U}_1(K,\eta,s^*,M^*)} \mathbb{P}_{\boldsymbol{\Theta}}(\mathcal{E}_6) \to 1,
\end{aligned}$$

as $n \to \infty$.

Now we are going to prove (B.19). Let $\mathcal{E} = \mathcal{E}_0 \cap \mathcal{E}_3 \cap \mathcal{E}_6$. Note that

$$\begin{aligned}
&\mathbb{P}_{\boldsymbol{\Theta}}\big(\sqrt{n}\widehat{S}(0, \widehat{\boldsymbol{\Theta}}_{(jk)^c})/(2\sigma) \le t\big) \le \mathbb{P}_{\boldsymbol{\Theta}}\big(\sqrt{n}\widehat{S}(0, \widehat{\boldsymbol{\Theta}}_{(jk)^c})/(2\sigma) \le t, \mathcal{E}\big) + \mathbb{P}_{\boldsymbol{\Theta}}(\mathcal{E}^c) \\
&\le \mathbb{P}_{\boldsymbol{\Theta}}\bigg(\sqrt{n}S(0, \boldsymbol{\Theta}_{(jk)^c})/(2\sigma) \le t + K n^{1/2-\eta}\frac{\widetilde{\sigma}^2}{2\sigma} + \eta(n)\bigg) + \mathbb{P}_{\boldsymbol{\Theta}}(\mathcal{E}^c).
\end{aligned}$$

Therefore, we have

$$\begin{aligned}
&\mathbb{P}_{\boldsymbol{\Theta}}\big(\sqrt{n}\widehat{S}(0, \widehat{\boldsymbol{\Theta}}_{(jk)^c})/(2\sigma) \le t\big) - \Phi(t) \\
&\le \mathbb{P}_{\boldsymbol{\Theta}}\bigg(\sqrt{n}S(0, \boldsymbol{\Theta}_{(jk)^c})/(2\sigma) \le t + K n^{1/2-\eta}\frac{\widetilde{\sigma}^2}{2\sigma} + \eta(n)\bigg) - \Phi(t) + \mathbb{P}_{\boldsymbol{\Theta}}(\mathcal{E}^c) \\
&= \mathbb{P}_{\boldsymbol{\Theta}}\bigg(\sqrt{n}S(0, \boldsymbol{\Theta}_{(jk)^c})/(2\sigma) \le t + K n^{1/2-\eta}\frac{\widetilde{\sigma}^2}{2\sigma} + \eta(n)\bigg) - \Phi\Big(t + K n^{1/2-\eta}\frac{\widetilde{\sigma}^2}{2\sigma} + \eta(n)\Big) \\
&\quad + \Phi\Big(t + K n^{1/2-\eta}\frac{\widetilde{\sigma}^2}{2\sigma} + \eta(n)\Big) - \Phi(t) + \mathbb{P}_{\boldsymbol{\Theta}}(\mathcal{E}^c).
\end{aligned}$$

By Lemma B.8 and the fact that $\Phi\big(t+Kn^{1/2-\eta}\widetilde{\sigma}^2/(2\sigma)+\eta(n)\big)-\Phi(t)\le\big(Kn^{1/2-\eta}\widetilde{\sigma}^2/(2\sigma)+\eta(n)\big)/\sqrt{2\pi}$ and $\eta>1/2$, we obtain that

$$\limsup_{n\to\infty}\sup_{\mathbf{\Theta}\in\mathcal{U}_1(K,\eta,s^*,M^*)}\sup_{t\in\mathbb{R}}\Big(\mathbb{P}_{\mathbf{\Theta}}\big(\sqrt{n}\widehat{S}(0,\widehat{\mathbf{\Theta}}_{(jk)^c})/(2\sigma)\le t\big)-\Phi(t)\Big)\le 0. \tag{C.43}$$

Following a similar argument, we can also show that

$$\liminf_{n\to\infty}\inf_{\mathbf{\Theta}\in\mathcal{U}_1(K,\eta,s^*,M^*)}\inf_{t\in\mathbb{R}}\Big(\mathbb{P}_{\mathbf{\Theta}}\big(\sqrt{n}\widehat{S}(0,\widehat{\mathbf{\Theta}}_{(jk)^c})/(2\sigma)\le t\big)-\Phi(t)\Big)\ge 0. \tag{C.44}$$

Combining (C.43) and (C.44), and using the fact that $\lim_{n\to\infty}\sup_{\mathbf{\Theta}\in\mathcal{U}_1(K,\eta,s^*,M^*)}\mathbb{P}_{\mathbf{\Theta}}(\mathcal{E}^c)=0$ completes the proof of (B.19). Note that (B.20) can be proved by the same argument, so we omit it. Finally, we are going to prove (B.21). Given event $\mathcal{E}$, it is easy to seen that $|\sqrt{n}\widehat{S}_n(0,\widehat{\mathbf{\Theta}}_{(jk)^c})/2\sigma|\le t$ implies that $\rho_{\min}\le\sqrt{n}\widehat{S}_n(0,\widehat{\mathbf{\Theta}}_{(jk)^c})/2\sigma\le\rho_{\max}$, where

$$\rho_{\min}=-t-\eta(n)+Kn^{1/2-\eta}\frac{\widetilde{\sigma}^2}{2\sigma},\quad \rho_{\max}=t+\eta(n)+Kn^{1/2-\eta}\frac{\widetilde{\sigma}^2}{2\sigma}.$$

We can show that

$$\begin{aligned}&\mathbb{P}_{\mathbf{\Theta}}\big(|\sqrt{n}\widehat{S}_n(0,\widehat{\mathbf{\Theta}}_{(jk)^c})/2\sigma|\le t\big)\le\mathbb{P}_{\mathbf{\Theta}}\big(\rho_{\min}\le\sqrt{n}\widehat{S}_n(0,\widehat{\mathbf{\Theta}}_{(jk)^c})/2\sigma\le\rho_{\max}\big)+\mathbb{P}_{\mathbf{\Theta}}(\mathcal{E}^c)\\&\le\mathbb{P}_{\mathbf{\Theta}}\big(\rho_{\min}\le\sqrt{n}\widehat{S}_n(0,\widehat{\mathbf{\Theta}}_{(jk)^c})/2\sigma\le\rho_{\max}\big)-\mathbb{P}(\rho_{\min}\le Z\le\rho_{\max})\\&\quad+\mathbb{P}(\rho_{\min}\le Z\le\rho_{\max})-\mathbb{P}_{\mathbf{\Theta}}(\mathcal{E}^c),\end{aligned}$$

where $Z\sim N(0,1)$. By Lemma B.8, we have

$$\limsup_{n\to\infty}\sup_{\mathbf{\Theta}\in\mathcal{U}_1}\big|\mathbb{P}_{\mathbf{\Theta}}\big(\rho_{\min}\le\sqrt{n}\widehat{S}_n(0,\widehat{\mathbf{\Theta}}_{(jk)^c})/2\sigma\le\rho_{\max}\big)-\mathbb{P}(\rho_{\min}\le Z\le\rho_{\max})\big|=0.$$

In addition, when $\eta<1/2$, if $K<0$, we have $\mathbb{P}(\rho_{\min}\le Z\le\rho_{\max})\le\Phi(\rho_{\max})\to 0$ uniformly over $\mathbf{\Theta}\in\mathcal{U}_1(K,\eta,s^*,M^*)$ as $n\to\infty$. Furthermore, we have that $\limsup_{n\to\infty}\sup_{\mathbf{\Theta}\in\mathcal{U}_1(K,eta,s^*,M^*)}\mathbb{P}_{\mathbf{\Theta}}(\mathcal{E}^c)=0$. The same arguments apply to the case that $K>0$, since $\mathbb{P}(\rho_{\min}\le Z\le\rho_{\max})\le 1-\Phi(\rho_{\min})\to 0$ uniformly over $\mathbf{\Theta}\in\mathcal{U}_1(K,\eta,s^*,M^*)$ as $n\to\infty$. This completes the proof of (B.21) □

## C.11 Proof of Lemma B.15

In order to prove Lemma B.15, we need the following auxiliary lemmas.

**Lemma C.2.** Let $A_n=\big\|1/n\sum_{i=1}^n\big(\mathbf{F}\odot\widehat{\mathbf{G}}^i\big)\cdot e_i\big\|_{\max}=\max_{1\le j,k\le d}\big|1/n\sum_{i=1}^n F_{jk}\cdot\widehat{G}^i_{jk}\cdot e_i\big|$. We have $A_n=O_{\mathbb{P}}\big(\log d/\sqrt{n}\big)$.

**Lemma C.3.** Let $B_n=\big|\mathbf{b}^\top 1/\sqrt{n}\sum_{i=1}^n\mathrm{vec}\big(\mathbf{F}\odot(\widehat{\mathbf{G}}^i-\overline{\mathbf{G}}^i)\big)\cdot e_i\big|$. We have $B_n=o_{\mathbb{P}}(1)$.

*Proof of Lemma B.15.* We prove this theorem by three steps. First, we will obtain a bound on $|T-T_0|$, then we will bound $|W-W_0|$.

**Bounds for $|T - T_0|$:** By triangle inequality, we have

$$|T - T_0| \le \underbrace{|T - T_1|}_{\text{(i)}} + \underbrace{|T_1 - T_2|}_{\text{(ii)}} + \underbrace{|T_2 - T_0|}_{\text{(iii)}}.$$

For term (i), we have

$$\begin{aligned}
&|T - T_1| \\
&= \left| \max_{(j,k)\in[d]\times[d]} \sqrt{n}\Big(\widehat{\Theta}_{jk} - \Theta^*_{jk}\Big) \cdot \frac{H_{(jk)|(jk)^c}}{2\sigma} - \max_{(j,k)\in[d]\times[d]} \sqrt{n}\Big(\widehat{\Theta}_{jk} - \Theta^*_{jk}\Big) \cdot \frac{\widehat{H}_{(jk)|(jk)^c}}{2\widehat{\sigma}} \right| \\
&\le \max_{(j,k)\in[d]\times[d]} \sqrt{n} \cdot \left| \Big(\widehat{\Theta}_{jk} - \Theta^*_{jk}\Big) \cdot \Big( H_{(jk)|(jk)^c}/(2\sigma) - \widehat{H}_{(jk)|(jk)^c}/(2\widehat{\sigma}) \Big) \right| \\
&\le \max_{(j,k)\in[d]\times[d]} \sqrt{n} \big|\widehat{\Theta}_{jk} - \Theta^*_{jk}\big| \cdot \big| H_{(jk)|(jk)^c}/(2\sigma) - \widehat{H}_{(jk)|(jk)^c}/(2\widehat{\sigma}) \big|. \qquad \text{(C.45)}
\end{aligned}$$

Since we have $\big|\widehat{\Theta}_{jk} - \Theta^*_{jk}\big| \le \|\widehat{\boldsymbol{\Theta}} - \boldsymbol{\Theta}^*\|_{\max} = O_{\mathbb{P}}(s\sqrt{\log d/n})$, and $\|H_{(jk)|(jk)^c}/(2\sigma) - \widehat{H}_{(jk)|(jk)^c}/(2\widehat{\sigma})\| = o_{\mathbb{P}}(1)$, we obtain $|T - T_1| = o_{\mathbb{P}}(1)$. For term (ii), we have

$$\begin{aligned}
|T_1 - T_2| &= \left| \max_{(j,k)\in[d]\times[d]} \sqrt{n}\Big(\widehat{\Theta}_{jk} - \Theta^*_{jk}\Big) \cdot H_{(jk)|(jk)^c}/(2\sigma) - \max_{(j,k)\in[d]\times[d]} \sqrt{n} S_n(\boldsymbol{\Theta}^*)/(2\sigma) \right| \\
&\le \max_{(j,k)\in[d]\times[d]} \sqrt{n} \cdot \left| \Big(\widehat{\Theta}_{jk} - \Theta^*_{jk}\Big) \cdot H_{(jk)|(jk)^c}/(2\sigma) - S_n(\boldsymbol{\Theta}^*)/(2\sigma) \right| = o_{\mathbb{P}}(1), \quad \text{(C.46)}
\end{aligned}$$

where the last equality follows from Theorem 4.11. For term (iii), we have

$$\begin{aligned}
|T_2 - T_0| &= \left| \max_{(j,k)\in[d]\times[d]} \sqrt{n} S_n(\boldsymbol{\Theta}^*)/(2\sigma) - \max_{(jk)\in[d]\times[d]} \frac{1}{\sqrt{n}} \sum_{i=1}^n Z_{ijk} \right| \\
&\le \max_{(j,k)\in[d]\times[d]} \left| \frac{1}{\sqrt{n}} \sum_{i=1}^n 1/(2\sigma) \cdot \mathbf{b}^\top \mathrm{vec}\Big(\mathbf{T} \odot (\mathbf{G}^i - \overline{\mathbf{G}}^i)\Big) \right|. \qquad \text{(C.47)}
\end{aligned}$$

In the proof of Lemma C.1, we have shown that $|T_2 - T_0| = o_{\mathbb{P}}(1)$. Furthermore, these bounds are independent of $(j,k)$, which means that after taking maximum over all $(j,k) \in [d] \times [d]$, the same bounds still hold. Therefore, we have $|T_0 - T_1| \xrightarrow{p} 0$. Combining terms (i), (ii) and (iii), we obtain $|T - T_0| \xrightarrow{p} 0$.

**Bounds for $|W - W_0|$:** We have

$$|W - W_0| \le \max_{(jk)\in[d]\times[d]} \underbrace{\left| \frac{1}{\sqrt{n}} \frac{\widehat{\sigma}}{\sigma} \sum_{i=1}^n z_{ijk} e_i - \frac{1}{\sqrt{n}} \sum_{i=1}^n Z_{ijk} e_i \right|}_{\text{(i)}} + \max_{(jk)\in[d]\times[d]} \underbrace{\frac{|\sigma - \widehat{\sigma}|}{\sigma} \cdot \left| \frac{1}{\sqrt{n}} \sum_{i=1}^n z_{ijk} e_i \right|}_{\text{(ii)}}.$$

By definition, for each $(j,k)\in[d]\times[d]$, we have

$$
\begin{aligned}
\text{(i)} \le\ & \underbrace{\frac{\sqrt{n}}{\sigma}\left|(\mathbf{b}-\widehat{\mathbf{b}})^\top \frac{1}{n}\sum_{i=1}^n \operatorname{vec}\big(\mathbf{F}\odot\widehat{\mathbf{G}}^i\big)e_i\right|}_{\text{(i).a}} + \underbrace{\frac{1}{\sigma}\left|\mathbf{b}^\top \frac{1}{\sqrt{n}}\sum_{i=1}^n \operatorname{vec}\big(\mathbf{F}\odot(\widehat{\mathbf{G}}^i-\overline{\mathbf{G}}^i)\big)e_i\right|}_{\text{(i).b}} \\
& + \underbrace{\frac{1}{\sigma}\left|\mathbf{b}^\top \frac{1}{\sqrt{n}}\sum_{i=1}^n \operatorname{vec}\big((\mathbf{F}-\mathbf{T})\odot\overline{\mathbf{G}}^i\big)e_i\right|}_{\text{(i).c}}.
\end{aligned}
$$

In the following, we are going to bound terms (i).a, (i).b and (i).c respectively. For term (i).a, we have that

$$
\begin{aligned}
\text{(i).a} &\le \big\|\widehat{\mathbf{b}}-\mathbf{b}^*\big\|_1\cdot\left\|\frac{1}{n}\sum_{i=1}^n \operatorname{vec}\big(\mathbf{F}\odot\widehat{\mathbf{G}}^i\big)\cdot e_i\right\|_\infty \le \big\|\widehat{\mathbf{b}}-\mathbf{b}^*\big\|_1\cdot\left\|\frac{1}{n}\sum_{i=1}^n \big(\mathbf{F}\odot\widehat{\mathbf{G}}^i\big)\cdot e_i\right\|_{\max} \\
&= \big\|\widehat{\mathbf{w}}-\mathbf{w}^*\big\|_1\cdot O_{\mathbb{P}}\bigg(\frac{\log d}{\sqrt{n}}\bigg), \qquad \text{(C.48)}
\end{aligned}
$$

where the last equality follows from $\widehat{\mathbf{b}}-\mathbf{b}^*=\widehat{\mathbf{w}}-\mathbf{w}^*$ and Lemma C.2. According to the proof of Lemma B.3, we have

$$
\big\|\widehat{\mathbf{w}}-\mathbf{w}^*\big\|_1 = O_{\mathbb{P}}\bigg(s^2\frac{\log d}{n}+s\sqrt{\frac{\log d}{n}}\bigg). \qquad \text{(C.49)}
$$

Substituting (C.49) into (C.48), we obtain

$$
\text{(i).a} = O_{\mathbb{P}}\bigg(s^2\bigg(\frac{\log d^2}{n^{3/2}}\bigg)+s\frac{(\log d)^{3/2}}{n}\bigg) = o_{\mathbb{P}}\big(n^{-1/2}\big). \qquad \text{(C.50)}
$$

Since the bound in (C.50) does not depend on specific $(j,k)\in[d]\times[d]$, taking maximum over all $(j,k)\in[d]\times[d]$, we have

$$
\max_{(j,k)\in[d]\times[d]}\left|(\mathbf{b}-\widehat{\mathbf{b}})^\top\frac{1}{n}\sum_{i=1}^n \operatorname{vec}\big(\mathbf{F}\odot\widehat{\mathbf{G}}^i\big)\cdot e_i\right| = o_{\mathbb{P}}\big(n^{-1/2}\big).
$$

For term (i).b, using Lemma C.3, we have

$$
\left|\mathbf{b}^\top\frac{1}{\sqrt{n}}\sum_{i=1}^n \operatorname{vec}\big(\mathbf{F}\odot(\widehat{\mathbf{G}}^i-\overline{\mathbf{G}}^i)\big)\cdot e_i\right| = o_{\mathbb{P}}(1). \qquad \text{(C.51)}
$$

We see that the bound of (i).b is uniform for all $(i,j)\in[d]\times[d]$. For term (i).c, we have

$$
\text{(i).c} \le \|\mathbf{b}_{\mathcal{B}}\|_1\cdot\sqrt{n}\big\|(\mathbf{F}-\mathbf{T})_{\mathcal{B}\times\mathcal{B}}\big\|_{\max}\cdot\left\|\bigg(\frac{1}{n}\sum_{i=1}^n\overline{\mathbf{G}}^i\bigg)_{\mathcal{B}\times\mathcal{B}}\right\|_{\max} \qquad \text{(C.52)}
$$

where $\mathcal{B} = \mathrm{supp}(\mathbf{b})$. In the proof of Lemma B.1, we have shown $\big\|(1/n\sum_{i=1}^n \overline{\mathbf{G}}^i)_{\mathcal{B}\times\mathcal{B}}\big\|_{\max} = O_{\mathbb{P}}(\sqrt{\log s/n})$ and $\big\|(\mathbf{T}-\mathbf{F})_{\mathcal{B}\times\mathcal{B}}\big\|_{\max} = O_{\mathbb{P}}(\sqrt{\log s/n})$. Therefore, we have (i).c $= O_{\mathbb{P}}\big((1+\nu^2M^2)\log d/n\big) = o_{\mathbb{P}}(1)$. For term (ii), it is easy to show that (ii) $= o_{\mathbb{P}}(1)$. Combining the results above, we get $|W - W_0| \xrightarrow{p} 0$. This completes the proof. □

### C.12 Proof of Lemma B.16

*Proof.* By Lemma B.15, we know that there exist $\xi_1$, $\xi_2$ depending on $n$ and $\xi_1 \to 0, \xi_2 \to 0$ as $n \to \infty$ such that

$$\mathbb{P}\big(|T - T_0| > \xi_1\big) \le \xi_2, \quad \mathbb{P}\big(|W - W_0| > \xi_1\big) \le \xi_2.$$

Without loss of generality, we can further assume that $\xi_2 \le 1$ because otherwise we can simply consider $n$ large enough to make this holds. Setting $\zeta_1 = \xi_1, \ \zeta_2 = \sqrt{\xi_2}$, we obtain

$$\mathbb{P}(|T - T_0| > \zeta_1) \le \zeta_2^2 \le \zeta_2, \quad \mathbb{P}(|W - W_0| > \zeta_1) \le \zeta_2^2.$$

By Markov's inequality, we have

$$\mathbb{P}\big(\mathbb{P}_e(|W - W_0| > \zeta_1) > \zeta_2\big) \le \frac{1}{\zeta_2}\mathbb{E}\big[\mathbb{P}_e(|W - W_0| > \zeta_1)\big] = \frac{1}{\zeta_2}\mathbb{P}(|W - W_0| > \zeta_1) \le \zeta_2.$$

This completes the proof. □

# D Proof of Auxiliary Lemmas in Appendix C

*Proof of Lemma C.1.* Recall that $\mathbf{T} = [T_{jk}] \in \mathbb{R}^{d\times d}$ such that $T_{jk} = \tau_{jk}(\boldsymbol{\Theta}^*)$, and $\mathbf{G} = [G_{jk}] \in \mathbb{R}^{d\times d}$ such that $G_{jk} = \frac{2}{n(n-1)}\sum_{1\le i<i'\le n} h_{jk}^{ii'}(\boldsymbol{\Theta})$. In addition, we have $\mathbf{G}^i = [G_{jk}^i]$ such that $G_{jk}^i = 1/(n-1)\sum_{i'\neq i} h_{jk}^{ii'}(\boldsymbol{\Theta}^*)$. In other words, we have $G_{jk} = 2/n\sum_{i=1}^n G_{jk}^i$. Since $\nabla \ell_n(\boldsymbol{\Theta}^*) = \mathbf{T}\odot\mathbf{G}$, we have

$$\begin{aligned}
&\frac{\sqrt{n}}{2}\big(\mathbf{b}^\top\mathbf{R}\mathbf{b}\big)^{-1/2}\mathbf{b}^\top\mathrm{vec}\big(\nabla\ell_n(\boldsymbol{\Theta}^*)\big)\\
&= \underbrace{\big(\mathbf{b}^\top\mathbf{R}\mathbf{b}\big)^{-1/2}\frac{1}{\sqrt{n}}\sum_{i=1}^n \mathbf{b}^\top\mathrm{vec}(\mathbf{F}\odot\overline{\mathbf{G}}^i)}_{\text{(i)}} + \underbrace{\frac{\sqrt{n}}{2}\big(\mathbf{b}^\top\mathbf{R}\mathbf{b}\big)^{-1/2}\mathbf{b}^\top\mathrm{vec}\big(\mathbf{F}\odot(\mathbf{G}-\overline{\mathbf{G}})\big)}_{\text{(ii)}}\\
&\quad + \underbrace{\big(\mathbf{b}^\top\mathbf{R}\mathbf{b}\big)^{-1/2}\frac{1}{2\sqrt{n}}\sum_{i=1}^n \mathbf{b}^\top\mathrm{vec}\big((\mathbf{T}-\mathbf{F})\odot\mathbf{G}^i\big)}_{\text{(iii)}}. \qquad \text{(D.1)}
\end{aligned}$$

In what follows, we will bound terms (i), (ii) and (iii) respectively.
**Bounding Term (i):** Term (i) is a sum of $n$ independent zero mean random variables. In addition, its variance is 1. We now verify the Lyapunov condition for term (i). By Assumption 4.2 and $\|\mathbf{b}\|_2 = \sqrt{1+\|\mathbf{w}^*\|_2^2} \le \sqrt{1+(s^2\|\mathbf{w}^*\|_1)^2} \le \sqrt{1+s^4\nu^4M^4}$, $\mathbf{b}^\top\mathbf{R}\mathbf{b}$ is lower bounded by $(1+s^4\nu^4M^4)\lambda_{\min}(\mathbf{R})$. In addition, we have $F_{jk} = \sqrt{1-\left(\Sigma_{jk}^*\right)^2} \le 1$, thus $F_{jk} = O_{\mathbb{P}}(1)$. Therefore,

$$n^{-3/2}\sum_{i=1}^n \mathbb{E}\Big[\Big|\big(\mathbf{b}^\top\mathbf{R}\mathbf{b}\big)^{-1/2}\frac{1}{\sqrt{n}}\sum_{i=1}^n \mathbf{b}^\top\mathrm{vec}(\mathbf{F}\odot\overline{\mathbf{G}}^i)\Big|^3\Big] = O_{\mathbb{P}}(1)n^{-3/2}\sum_{i=1}^n \mathbb{E}\Big[\big|\mathbf{b}^\top\mathrm{vec}(\mathbf{F}\odot\overline{\mathbf{G}}^i)\big|^3\Big]. \tag{D.2}$$

We have $\|\mathbf{b}\|_1 = 1+\|\mathbf{w}^*\|_1 = 1+\nu^2M^2$. Furthermore, we have

$$\big|h_{jk}^{ii'}(\mathbf{\Theta}^*)\big| \le \Big|-\frac{\pi}{2}\,\mathrm{sign}\left[\left(X_{ij}-X_{i'j}\right)\left(X_{ik}-X_{i'k}\right)\right]\Big| + \big|\arcsin\left([\mathbf{\Theta}^{*-1}]_{jk}\right)\big| \le \frac{\pi}{2}+\frac{\pi}{2} = \pi.$$

Similarly, we can show that $\big|h_{jk}^{ii'|i}(\mathbf{\Theta}^*)\big| \le \pi$ and $\big|g_{jk}(\boldsymbol{X}_i,\mathbf{\Theta}^*)\big| \le \pi$. Therefore, we have $\big\|\mathrm{vec}(\mathbf{F}\odot\overline{\mathbf{G}}^i)\big\|_\infty \le \|\mathbf{F}\|_{\max}\cdot\|\overline{\mathbf{G}}^i\|_{\max} \le \pi$, by Hölder's inequality, we obtain

$$|\mathbf{b}^\top\mathrm{vec}(\mathbf{F}\odot\overline{\mathbf{G}}^i)| \le \|\mathbf{b}\|_1\cdot\big\|\mathrm{vec}(\mathbf{F}\odot\overline{\mathbf{G}}^i)\big\|_\infty \le \pi(1+\nu^2M^2). \tag{D.3}$$

Substituting (D.3) into (D.2), we obtain

$$n^{-3/2}\sum_{i=1}^n \mathbb{E}\Big[\Big|\big(\mathbf{b}^\top\mathbf{R}\mathbf{b}\big)^{-1/2}\frac{1}{\sqrt{n}}\sum_{i=1}^n \mathbf{b}^\top\mathrm{vec}(\mathbf{F}\odot\overline{\mathbf{G}}^i)\Big|^3\Big] = O_{\mathbb{P}}\big(n^{-1/2}\pi^3(1+\nu^2M^2)^3\big) = o_{\mathbb{P}}(1).$$

Thus, by invoking the Lyapunov central limit theorem for term (i), we establish that

$$\big(\mathbf{b}^\top\mathbf{R}\mathbf{b}\big)^{-1/2}\frac{1}{\sqrt{n}}\sum_{i=1}^n \mathbf{b}^\top\mathrm{vec}(\mathbf{F}\odot\overline{\mathbf{G}}^i) \rightsquigarrow N(0,1). \tag{D.4}$$

**Bounding Term (ii):** We define a matrix $\mathbf{V}^{ii'} = [V_{jk}^{ii'}]$ such that $V_{jk}^{ii'} = h_{jk}^{ii'}(\mathbf{\Theta}^*) - h_{jk}^{ii'|i}(\mathbf{\Theta}^*) - h_{jk}^{ii'|i'}(\mathbf{\Theta}^*)$. Term (ii) can be rewritten as

$$\text{(ii)} = \frac{\sqrt{n}}{2}(\mathbf{b}^\top\mathbf{R}\mathbf{b})^{-1/2}\frac{1}{n(n-1)}\sum_{1\le i<i'\le n}\mathbf{b}^\top\mathrm{vec}(\mathbf{F}\odot\mathbf{V}^{ii'}). \tag{D.5}$$

In the following, we introduce $h_{jk}^{ii'}$ as the shorthand of $h_{jk}^{ii'}(\mathbf{\Theta}^*)$ and $h_{jk}^{ii'|i}$ as the shorthand of $h_{jk}^{ii'|i}(\mathbf{\Theta}^*)$. We consider two cases. For the first case, suppose $i \neq \ell,\ell'$, $i' \le \ell,\ell'$, we have

$$\begin{aligned}\mathbb{E}\big[V_{jk}^{ii'}V_{j'k'}^{\ell\ell'}\big] &= \mathbb{E}\big[h_{jk}^{ii'}h_{j'k'}^{\ell\ell'}\big] - \mathbb{E}\big[h_{jk}^{ii'}h_{j'k'}^{\ell\ell'|\ell}\big] - \mathbb{E}\big[h_{jk}^{ii'}h_{j'k'}^{\ell\ell'|\ell'}\big] - \mathbb{E}\big[h_{jk}^{ii'|i}h_{j'k'}^{\ell\ell'}\big] - \mathbb{E}\big[h_{jk}^{ii'|i'}h_{j'k'}^{\ell\ell'}\big]\\ &\quad + \mathbb{E}\big[h_{jk}^{ii'|i}h_{j'k'}^{\ell\ell'|\ell}\big] + \mathbb{E}\big[h_{jk}^{ii'|i}h_{j'k'}^{\ell\ell'|\ell'}\big] + \mathbb{E}\big[h_{jk}^{ii'|i'}h_{j'k'}^{\ell\ell'|\ell}\big] + \mathbb{E}\big[h_{jk}^{ii'|i'}h_{j'k'}^{\ell\ell'|\ell'}\big].\end{aligned}$$

Since $\mathbb{E}\big[h_{jk}^{ii'}\big] = \mathbb{E}\big[h_{jk}^{\ell\ell'}\big] = 0$, and $h_{jk}^{ii'}$ and $h_{jk}^{\ell\ell'}$ are independent, we can show that $\mathbb{E}\big[V_{jk}^{ii'}V_{j'k'}^{\ell\ell'}\big] = 0$ for the first case. For the second case, suppose one of $i$ and $i'$ is identical to one of $\ell$ and $\ell'$,

e.g., $i = \ell$, then

$$\mathbb{E}\big[V_{jk}^{ii'} V_{j'k'}^{i\ell'}\big] = \mathbb{E}\big[h_{jk}^{ii'} h_{j'k'}^{i\ell'}\big] - \mathbb{E}\big[h_{jk}^{ii'} h_{j'k'}^{i\ell'|i}\big] - \mathbb{E}\big[h_{jk}^{ii'|i} h_{j'k'}^{i\ell'}\big] + \mathbb{E}\big[h_{jk}^{ii'|i} h_{j'k'}^{i\ell'|i}\big]. \tag{D.6}$$

Note that we have

$$\begin{aligned}
\mathbb{E}\big[h_{jk}^{ii'} h_{j'k'}^{i\ell'}\big] &= \mathbb{E}\Big[\mathbb{E}\big[h_{jk}^{ii'} h_{j'k'}^{i\ell'} \big| \boldsymbol{X}_i\big]\Big] = \mathbb{E}\big[h_{jk}^{ii'|i} h_{j'k'}^{i\ell'|i}\big] = 0, \\
\mathbb{E}\big[h_{jk}^{ii'} h_{j'k'}^{i\ell'|i}\big] &= \mathbb{E}\Big[\mathbb{E}\big[h_{jk}^{ii'} \big| \boldsymbol{X}_i\big] h_{j'k'}^{i\ell'|i}\Big] = \mathbb{E}\big[h_{jk}^{ii'|i} h_{j'k'}^{i\ell'|i}\big] = 0, \\
\mathbb{E}\big[h_{jk}^{ii'|i} h_{j'k'}^{i\ell'}\big] &= \mathbb{E}\Big[h_{jk}^{ii'|i} \mathbb{E}\big[h_{j'k'}^{i\ell'} \big| \boldsymbol{X}_i\big]\Big] = \mathbb{E}\big[h_{jk}^{ii'|i} h_{j'k'}^{i\ell'|i}\big] = 0.
\end{aligned} \tag{D.7}$$

Substituting (D.7) into (D.6), we have $\mathbb{E}\big[V_{jk}^{ii'} V_{j'k'}^{i\ell'}\big] = 0$ for the second case. The variance of term (ii) can be bounded by

$$\begin{aligned}
&\frac{n}{4}(\mathbf{b}^\top \mathbf{R}\mathbf{b})^{-1} \frac{1}{n^2(n-1)^2} \sum_{1 \le i < i' \le n} \sum_{1 \le \ell < \ell' \le n} \mathbf{b}^\top \mathbb{E}\big[\mathrm{vec}\big(\mathbf{F} \odot \mathbf{V}^{ii'}\big) \mathrm{vec}\big(\mathbf{F} \odot \mathbf{V}^{\ell\ell'}\big)\big] \mathbf{b} \\
&= \frac{n}{4}(\mathbf{b}^\top \mathbf{R}\mathbf{b})^{-1} \frac{1}{n^2(n-1)^2} \sum_{1 \le i < i' \le n} \mathbf{b}^\top \mathbb{E}\big[\mathrm{vec}\big(\mathbf{F} \odot \mathbf{V}^{ii'}\big) \mathrm{vec}\big(\mathbf{F} \odot \mathbf{V}^{ii'}\big)\big] \mathbf{b}.
\end{aligned}$$

Recall that we have $\mathbf{b}^\top \mathbf{R}\mathbf{b}$ is lower bounded by $(1+s^4\nu^4M^4)\lambda_{\min}(\mathbf{R})$ and therefore $(\mathbf{b}^\top \mathbf{R}\mathbf{b})^{-1} = O(1)$, We have $|V_{jk}^{ii'}| \le 3\pi$ and $|V_{jk}^{ii'} V_{j'k'}^{ii'}| \le 9\pi^2$. Thus, the variance of term (ii) can be further bounded by

$$\begin{aligned}
&O(1) O_{\mathbb{P}}(1) \frac{1}{n(n-1)^2} \sum_{1 \le i < i' \le n} \mathbf{b}^\top \mathbb{E}\big[\mathrm{vec}\big(\mathbf{V}^{ii'}\big) \mathrm{vec}\big(\mathbf{V}^{ii'}\big)\big] \mathbf{b} \\
&= O_{\mathbb{P}}(1) \frac{1}{n(n-1)^2} \sum_{1 \le i < i' \le n} \Big\| \mathbb{E}\big[\mathrm{vec}\big(\mathbf{V}^{ii'}\big) \mathrm{vec}\big(\mathbf{V}^{ii'}\big)\big] \Big\|_{\max} \cdot \|\mathbf{b}\|_1^2 \\
&= O_{\mathbb{P}}\big(\pi^2(1+\nu^2M^2)^2/(n-1)\big) = o_{\mathbb{P}}(1).
\end{aligned} \tag{D.8}$$

By Chebyshev's inequality and (D.8), we have (ii) $= o_{\mathbb{P}}(1)$.

**Bounding Term** (iii)**:** Let $\mathcal{B} = \mathrm{supp}(\mathbf{b})$. We have

$$\begin{aligned}
\text{(iii)} &\le (\mathbf{b}^\top \mathbf{R}\mathbf{b})^{-1/2} \|\mathbf{b}_{\mathcal{B}}\|_1 \cdot \frac{\sqrt{n}}{2} \cdot \left\| \left( \frac{1}{n} \sum_{i=1}^n (\mathbf{T} - \mathbf{F}) \odot \mathbf{G}^i \right)_{\mathcal{B} \times \mathcal{B}} \right\|_{\max} \\
&\le (\mathbf{b}^\top \mathbf{R}\mathbf{b})^{-1/2} \|\mathbf{b}_{\mathcal{B}}\|_1 \cdot \frac{\sqrt{n}}{2} \cdot \big\| (\mathbf{T} - \mathbf{F})_{\mathcal{B} \times \mathcal{B}} \big\|_{\max} \cdot \left\| \left( \frac{1}{n} \sum_{i=1}^n \mathbf{G}^i \right)_{\mathcal{B} \times \mathcal{B}} \right\|_{\max}.
\end{aligned} \tag{D.9}$$

We have already showed that $(\mathbf{b}^\top \mathbf{R}\mathbf{b})^{-1} = O(1)$, $\|\mathbf{b}\|_1 = 1 + \nu^2 M^2$.

By mean value theorem, we have

$$|G_{jk}^i| = \left| \frac{1}{n-1} \sum_{i' \neq i} h_{jk}^{ii'}(\boldsymbol{\Theta}^*) \right| \leq \frac{C}{\sqrt{1-\xi^2}} \left| \widehat{\Sigma}_{jk} - \Sigma_{jk}^* \right|, \ \forall (j,k) \in [n] \times [n], \tag{D.10}$$

where the inequality follows from Assumption 4.2 and $\mathrm{diag}(\boldsymbol{\Sigma}^*) = \mathbf{I}$, and $C$ is a constant. By (D.10) and Theorem 4.2 in (Liu et al., 2012), we obtain $\left\|(1/n \sum_{i=1}^n \mathbf{G}^i)_{\mathcal{B}\times\mathcal{B}}\right\|_{\max} = O_{\mathbb{P}}(\sqrt{\log s/n})$. Similarly, by mean value theorem, we have

$$\begin{aligned} |T_{jk} - F_{jk}| &= \left| \cos[t \arcsin\left(\Sigma_{jk}^*\right) + (1-t)\arcsin(\widehat{\Sigma}_{jk})] - \cos[\arcsin(\Sigma_{jk}^*)] \right| \\ &\leq \left| (t-1)\arcsin(\Sigma_{jk}^*) + (1-t)\arcsin(\widehat{\Sigma}_{jk}) \right| \leq \frac{C}{\sqrt{1-\xi^2}} |\Sigma_{jk}^* - \widehat{\Sigma}_{jk}|, \end{aligned} \tag{D.11}$$

where the last inequality follows from Assumption 4.2 and $\mathrm{diag}(\boldsymbol{\Sigma}^*) = \mathbf{I}$, and $C$ is a constant. Thus, by (D.11) and Theorem 4.2 in (Liu et al., 2012), we obtain $\left\|(\mathbf{T} - \mathbf{F})_{\mathcal{B}\times\mathcal{B}}\right\|_{\max} = O_{\mathbb{P}}(\sqrt{\log s/n})$. Therefore, we have (iii) $= \big((1+\nu^2 M^2)\log s\big)/\sqrt{n} = o_{\mathbb{P}}(1)$.

Combining terms (i), (ii) and (iii), and invoking Slutsky's theorem, we completes the proof. □

*Proof of Lemma C.2.* We have

$$\begin{aligned} \mathbb{E}\big[A_n\big] &\leq \mathbb{E}_X \max_{1\leq j,k\leq d} \frac{1}{n} \sqrt{\mathbb{E}_e \left[\sum_{i=1}^n F_{jk} \cdot \widehat{G}_{jk}^i \cdot e_i\right]^2} \sqrt{4\log d} \\ &\leq \mathbb{E}_X \max_{1\leq j,k\leq d} \frac{1}{n} \sqrt{\sum_{i=1}^n \big(F_{jk} \cdot \widehat{G}_{jk}^i\big)^2} \sqrt{4\log d}. \end{aligned} \tag{D.12}$$

The right hand side of (D.12) can be further bounded by

$$\mathbb{E}\big[A_n\big] \leq \max_{1\leq j,k\leq d} \frac{4\log d}{n} \sqrt{\mathbb{E}_X \sum_{i=1}^n \big(F_{jk} \cdot \widehat{G}_{jk}^i\big)^2},$$

which follows from maximal inequality for sub-Gaussian random variables. Since $|F_{jk} \cdot \widehat{G}_{jk}^i| \leq \|\mathbf{F}\|_{\max} \cdot \|\widehat{\mathbf{G}}^i\|_{\max} \leq \pi$, we directly get the upper bound as $\mathbb{E}\big[A_n\big] \leq 4\log d/n\sqrt{n\pi^2} \leq 4\pi \log d/\sqrt{n}$. We complete the proof by applying Markov's inequality. □

*Proof of Lemma C.3.* We have

$$\mathbb{E}\big[B_n\big] \le \mathbb{E}_X \sqrt{\mathbb{E}_e\left[\frac{1}{\sqrt{n}}\sum_{i=1}^n \mathbf{b}^\top \mathrm{vec}\big(\mathbf{F}\odot(\widehat{\mathbf{G}}^i - \overline{\mathbf{G}}^i)\big)\cdot e_i\right]^2}$$
$$\le \mathbb{E}_X \sqrt{\frac{1}{n}\sum_{i=1}^n \big[\mathbf{b}^\top \mathrm{vec}\big(\mathbf{F}\odot(\widehat{\mathbf{G}}^i - \overline{\mathbf{G}}^i)\big)\big]^2}. \tag{D.13}$$

The right hand side of (D.13) can be further bounded by

$$\mathbb{E}_X \sqrt{\frac{1}{n}\sum_{i=1}^n \big[\mathbf{b}^\top \mathrm{vec}\big(\mathbf{F}\odot(\widehat{\mathbf{G}}^i - \overline{\mathbf{G}}^i)\big)\big]^2} \le \sqrt{\mathbb{E}_X\left\{\frac{1}{n}\sum_{i=1}^n \big[\mathbf{b}^\top \mathrm{vec}\big(\mathbf{F}\odot(\widehat{\mathbf{G}}^i - \overline{\mathbf{G}}^i)\big)\big]^2\right\}}.$$

By the similar proof of Lemma C.1, we can show that $\mathbb{E}_x\Big\{1/n\sum_{i=1}^n \big[\mathbf{b}^\top \mathrm{vec}\big(\mathbf{F}\odot(\widehat{\mathbf{G}}^i - \overline{\mathbf{G}}^i)\big)\big]^2\Big\} = O_{\mathbb{P}}(1/n^2)$. Therefore, we have $\mathbb{E}\big[B_n\big] = O_{\mathbb{P}}\big(1/n\big) = o_{\mathbb{P}}(1)$. We complete the proof by applying Markov's inequality. □